%% file: main_iclr.tex
\documentclass{article} % For LaTeX2e
\usepackage{iclr2024_conference,times}

\input{math_commands.tex}

\usepackage{url}
\usepackage[pagebackref=true,breaklinks=true,letterpaper=true,colorlinks,citecolor=darkerblue,bookmarks=false]{hyperref}

\def\submissionarXiv{1}  % arXiv long version
\def\submissionICLR{2}  % ICLR review version
\def\submissionFinal{3}  % ICLR final version
\def\submission{\submissionarXiv}  % preprint
\usepackage{graphicx}
\usepackage{subfig}
\usepackage{subfloat}
\usepackage{booktabs} % for professional tables
\usepackage{listings}
\usepackage{multirow}
\usepackage{float}
\usepackage{amsmath}
\usepackage{amssymb}
\usepackage{mathtools}
\usepackage{amsthm}

\usepackage{xcolor}  % colors
\usepackage{wrapfig}
\usepackage{colortbl}
\usepackage[makeroom]{cancel}
\usepackage{cases}

\definecolor{gray}{rgb}{0.5,0.5,0.5}
\definecolor{darkergreen}{RGB}{21, 152, 56}
\definecolor{darkerblue}{rgb}{0,0.08,0.45}
\definecolor{darkerred}{RGB}{220, 35, 120}
\definecolor{RoyalBlue}{RGB}{65,105,225}
\definecolor{YellowOrange}{RGB}{255,165,0}
\definecolor{gray94}{gray}{.92}
\definecolor{gray90}{gray}{.90}
\definecolor{gray85}{gray}{.85}

\usepackage{color,soul} % add background colors for text
\sethlcolor{gray90} % set the background color

\if\submission\submissionICLR  % ICLR rebuttal version
    \newcommand{\pl}[1]{\textcolor{darkerred}{#1}}  % rebuttal
\else
    \newcommand{\pl}[1]{\textcolor{black}{#1}}  % final version
\fi
\usepackage{pifont}% 
\usepackage{xspace}
\usepackage{placeins}
\newcommand{\cmarkg}{\textcolor{gray}{\ding{51}}\xspace}%
\newcommand{\xmarkg}{\textcolor{gray}{\ding{55}}\xspace}%
\title{MogaNet: Multi-order Gated Aggregation Network}  % iclr

\author{  %%% update arXiv 2025
    \hspace{-0.5em}
    Siyuan Li$^{1,2}$\thanks{First two authors contribute equally.\ \ \ $^\dag$Corrsponding author (\texttt{stan.zq.li@westlake.edu.cn}).}~~~
    Zedong Wang$^{2*}$~~%\textsuperscript{\rm 1}
    Zicheng Liu$^{1,2}$~~%\textsuperscript{\rm 1}
    Cheng Tan$^{1,2}$~~%\textsuperscript{\rm 1}
    Haitao Lin$^{1,2}$~~%\textsuperscript{\rm 1}
    Di Wu$^{1,2}$~\\%\textsuperscript{\rm 1}
    \textbf{Zhiyuan Chen}$^{1,2}$~~%\textsuperscript{\rm 1}
    \textbf{Jiangbin Zheng}$^{1,2}$~~%\textsuperscript{\rm 1}
    \textbf{Stan Z. Li}$^{2\dag}$\\%\thanks{Corrsponding Author.}\\
    $^{1}$Zhejiang University, College of Computer Science and Technology, Hangzhou, China\\
    $^{2}$AI Lab, Research Center for Industries of the Future, Westlake University, Hangzhou, China\\
}

\iclrfinalcopy % Uncomment for camera-ready version, but NOT for submission.
\begin{document}

\maketitle

%%%%%%%%% ABSTRACT
\input{0_abstract}

%%%%%%%%% BODY TEXT
\input{1_introduction}
\input{2_related_works}
\input{3_method}
\input{4_experiments}
\input{5_conclusion}

%%%%%%%%% REFERENCES
{
\bibliography{reference}
\bibliographystyle{iclr2024_conference}
}

%%%%%%%%% APPENDIX
\clearpage
\input{6_appendix}

\end{document}

%% file: math_commands.tex
\usepackage{amsmath,amsfonts,bm}

\def\eqref#1{equation~\ref{#1}}
\def\1{\bm{1}}

\DeclareMathAlphabet{\mathsfit}{\encodingdefault}{\sfdefault}{m}{sl}
\SetMathAlphabet{\mathsfit}{bold}{\encodingdefault}{\sfdefault}{bx}{n}

%% file: 0_abstract.tex
\begin{abstract}

By contextualizing the kernel as global as possible, Modern ConvNets have shown great potential in computer vision tasks.
However, recent progress on \textit{multi-order game-theoretic interaction} within deep neural networks (DNNs) reveals the representation bottleneck of modern ConvNets, where the expressive interactions have not been effectively encoded with the increased kernel size. 
%\pl{Attempts have been made to address this challenge through loss function design and data distribution alteration.}
To tackle this challenge, we propose a new family of modern ConvNets, dubbed MogaNet, for discriminative visual representation learning in pure ConvNet-based models with favorable complexity-performance trade-offs. 
%In this paper, we explore the representation ability of ConvNet through the prism of \textit{multi-order game-theoretic interaction}, which portrays inter-variable interaction effects w.r.t.~varying scales of context via game theory.
%
MogaNet encapsulates conceptually simple yet effective convolutions and gated aggregation into a compact module, where discriminative features are efficiently gathered and contextualized adaptively.
%Extensive experiments show that MogaNet exhibits great scalability, impressive efficiency of model parameters, and competitive performance compared to state-of-the-art ViTs and ConvNets on ImageNet and various downstream vision benchmarks, including COCO object detection, ADE20K semantic segmentation, 2D\&3D human pose estimation, and video prediction.
MogaNet exhibits great scalability, impressive efficiency of parameters, and competitive performance compared to state-of-the-art ViTs and ConvNets on ImageNet and various downstream vision benchmarks, including COCO object detection, ADE20K semantic segmentation, 2D\&3D human pose estimation, and video prediction.
Notably, MogaNet hits 80.0\% and 87.8\% accuracy with 5.2M and 181M parameters on ImageNet-1K, outperforming ParC-Net and ConvNeXt-L, while saving 59\% FLOPs and 17M parameters, respectively.
% 
% code (arxiv & final version)
The source code is available at \url{https://github.com/Westlake-AI/MogaNet}.
% \vspace{-0.5em}

\end{abstract}

%% file: 1_introduction.tex
\section{Introduction}
\label{sec:intro}
% figure: Acc vs Param vs GFLOPs
\begin{wrapfigure}{r}{0.5\linewidth}
    \vspace{-4.5em}
    \begin{center}
    \includegraphics[width=1.0\linewidth]{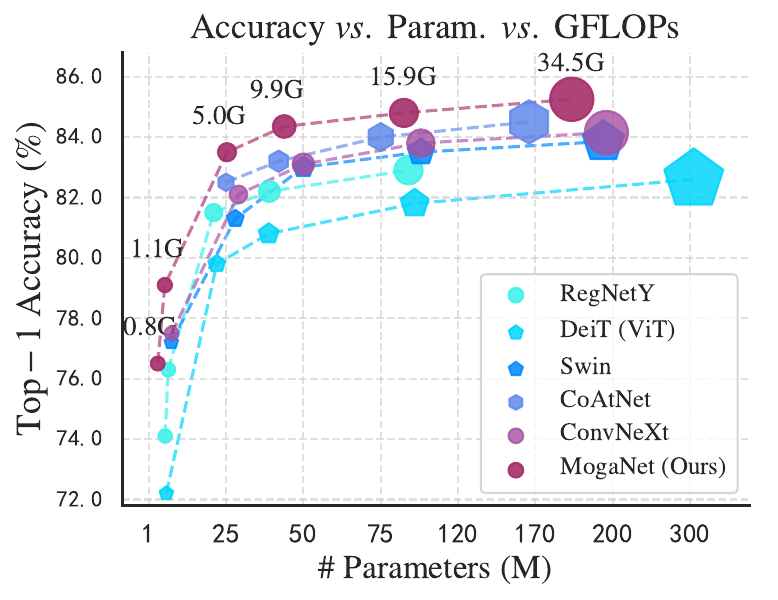}
    \end{center}
    \vspace{-1.5em}
    \caption{
    \textbf{Performance on ImageNet-1K validation set at $224^2$ resolutions.} MogaNet %with pure Convolutions 
    outperforms Transformers (DeiT\citep{icml2021deit} and Swin~\citep{liu2021swin}), ConvNets (RegNetY~\citep{cvpr2020regnet} and ConvNeXt~\citep{cvpr2022convnext}), and hybrid models (CoAtNet~\citep{nips2021coatnet}) across all %parameter
    scales.
    }
    \label{fig:in1k_acc_param}
    \vspace{-1.5em}
\end{wrapfigure}

By relaxing local inductive bias, Vision Transformers (ViTs)~\citep{iclr2021vit, liu2021swin} have rapidly challenged the long dominance of Convolutional Neural Networks (ConvNets)~\citep{tpami2015faster, he2016deep, cvpr2019semanticFPN} for visual recognition.
It is %a unanimous consensus 
commonly conjectured that such superiority of ViT stems from its self-attention operation \citep{bahdanau2014neural, vaswani2017attention}, which facilitates the global-range feature interaction.
From a practical standpoint, however, the quadratic complexity within self-attention prohibitively restricts its computational efficiency~\citep{nips2020linformer, icml2022FLASH} and applications to high-resolution fine-grained scenarios~\citep{zhu2020deformable, jiang2021transgan, cvpr2022VideoSwin}. Additionally, the dearth of local bias induces the detriment of neighborhood correlations~\citep{pinto2022impartial}. 

% Local ViT & Modern ConvNet Architectures
To resolve this problem, endeavors have been made by reintroducing %shift-invariant priors
locality priors~\citep{wu2021cvt, nips2021coatnet, han2021transformer, iclr2022uniformer, cvpr2022MobileFormer} and pyramid-like hierarchical layouts~\citep{liu2021swin, fan2021multiscale, iccv2021PVT} to ViTs, albeit at the expense of model generalizability and expressivity. 
% Recent work has shown that the representation capacity of ViT should mainly be credited to its macro-level framework rather than the commonly-conjectured self-attention mechanism~\citep{nips2021MLPMixer, raghu2021vision, yu2022metaformer}.
Meanwhile, further explorations toward ViTs~\citep{nips2021MLPMixer, raghu2021vision, yu2022metaformer} have triggered the resurgence of modern ConvNets~\citep{cvpr2022convnext, cvpr2022replknet}. 
With advanced training setup and ViT-style framework design, ConvNets can readily deliver competitive performance \emph{w.r.t.}~well-tuned ViTs across a wide range of vision benchmarks~\citep{wightman2021rsb, pinto2022impartial}.
%and further alters the roadmap for deep network architecture design.
% Nevertheless, there remains a representation bottleneck for existing approaches~\citep{hermann2020origins, iclr2022how, deng2021discovering, wu2022bottleneck}: naive implementation of self-attention or large kernels hampers the modeling of discriminative contextual information and global interactions, leading to a cognition gap between DNNs and human visual system.
% As in feature integration theory~\citep{treisman1980feature}, human brains not only extract local features but simultaneously aggregate these features for global perception, which is more compact and efficient than DNNs~\citep{liu2021swin, cvpr2022convnext}.
Essentially, most of the modern ConvNets aim to perform feature extraction in a \textit{local-global blended fashion} by contextualizing the convolutional kernel or the perception module as \textit{global} as possible.
%Despite their superior performance, previous progress of  \textit{local-global blending} operation is usually derived from intuitive insights, lacking a hierarchical theoretic reference and guidance.
%or by injecting the computationally efficient \textit{locality} into ViTs (hierarchical ViTs). 
%Despite their superior performance, existing \textit{local-global blending} operation is usually derived from intuitive insights, lacking a hierarchical theoretic reference and guidance.
%Moreover, the essential \textit{adaptive} nature of attention in ViTs has not been well leveraged and grafted into ConvNets, which unveils the pitfalls but has great potential for modern ConvNet architecture.

Despite their superior performance, recent progress on \textit{multi-order game-theoretic interaction} within DNNs~\citep{icml2019explaining, zhang2020interpreting, cheng2021game} unravels that the representation capacity of modern ConvNets has not been exploited well. 
Holistically, low-order interactions tend to model relatively simple and common local visual concepts, which are of poor expressivity and \pl{are incapable of capturing high-level semantic patterns.}
In comparison, the high-order ones represent the complex concepts of absolute global scope yet are vulnerable to attacks and with poor generalizability.
%More details about multi-order interaction theory have been discussed in Sec.3.
\cite{deng2021discovering} \pl{first shows that modern networks are implicitly prone to encoding extremely low- or high-order interactions rather than the empirically proved more discriminative middle ones.} 
%As shown in Fig.~\ref{fig:spatial_interaction}, existing modern ConvNets are implicitly prone to encoding extremely low- or high-order interactions with poor expressivity or generalizability rather than the \pl{empirically proved} discriminative middle ones~\citep{deng2021discovering, li2022A2MIM}. 
%
\pl{Attempts have been made to tackle this issue from the perspective of loss function~\citep{deng2021discovering} and modeling contextual relations~\citep{Wu2022DiscoveringTR, li2022A2MIM}.}
This unveils the serious challenge but also the great potential for modern ConvNet architecture design.
%

%\textbf{Multi-order interaction in vision.}
%In this work, we cast our sights on \textit{multi-order game-theoretic interaction}~\citep{icml2019explaining, zhang2020interpreting, cheng2021game}, which portrays marginal contribution brought by interactions among two selected pixels and varying scales of involved contexts, where the \textit{order} indicates the scale of involved contextual pixels within such process.
%chould be an overlooked hierarchical analysis tool for DNNs~\citep{deng2021discovering}.
%
%limits their representation abilities and robustness to complex samples.

To this end, we present a new ConvNet architecture named \textbf{{M}}ulti-\textbf{{o}}rder \textbf{{g}}ated \textbf{{a}}ggregation Network (MogaNet) to achieve \textit{adaptive} context extraction and further pursue more discriminative and efficient visual representation learning \pl{first under the guidance of interaction} within modern ConvNets.
In MogaNet, we encapsulate both locality perception and gated context aggregation into a compact spatial aggregation block, where features encoded \pl{by the inherent overlooked interactions are forced to }congregated and contextualized efficiently in parallel.
From the channel perspective, as existing methods are prone to huge channel-wise information redundancy~\citep{raghu2021vision, icml2022FLASH}, we design a conceptually simple yet
effective channel aggregation block to adaptively \pl{force the network to encode expressive interactions that would have originally been ignored.} Intuitively, it performs channel-wise reallocation to the input, which outperforms prevalent counterparts (\textit{e.g.}, SE~\citep{hu2018squeeze}, \pl{RepMLP~\citep{ding2022repmlpnet}}) with more favorable computational overhead. 

% Experimental results
Extensive experiments demonstrate the consistent efficiency of model parameters and competitive performance of MogaNet at different model scales on various vision tasks, including image classification, object detection, semantic segmentation, instance segmentation, pose estimation, \textit{etc.}
% We empirically show that interaction complexity can serve as an essential indication, like the receptive field, for high-quality visual architecture design.
As shown in Fig.~\ref{fig:in1k_acc_param}, MogaNet achieves 83.4\% and 87.8\% top-1 accuracy with 25M and 181M parameters, which exhibits favorable computational overhead compared with existing lightweight models. 
MogaNet-T attains 80.0\% accuracy on ImageNet-1K, outperforming the state-of-the-art ParC-Net-S~\citep{eccv2022edgeformer} by 1.0\% with 2.04G lower FLOPs. 
MogaNet also shows great performance gain on various downstream tasks, \textit{e.g.,} surpassing Swin-L~\citep{liu2021swin} by 2.3\% AP$^b$ on COCO detection with fewer parameters and computational budget.
%Hence, the observed performance gain cannot just be attributed to the expansion of the model scale but the more efficient use of parameters.
It is surprising that the parameter efficiency of MogaNet exceeds our expectations. This is probably owing to the network encodes more discriminative middle-order interactions, which maximizes the usage of model parameters.

%% file: 2_related_works.tex
% \section{Related Work}
% \label{sec:related_work}
% \paragraph{Post-ViT Modern ConvNets}

\section{Related Work}
\label{sec:preliminaries}
\vspace{-0.25em}
\subsection{Vision Transformers}
Since the success of Transformer~\citep{vaswani2017attention} in natural language processing~\citep{devlin2018bert},
ViT has been proposed~\citep{iclr2021vit} and attained impressive results on ImageNet~\citep{cvpr2009imagenet}. %It splits raw images into non-overlapping fixed-size patches as visual tokens to capture long-range interactions among these tokens by self-attention. 
% Hybrid ViTs: Swin, Uniformer, Next-ViT
Yet, compared to ConvNets, ViTs are over-parameterized and rely on large-scale pre-training \citep{ bao2021beit, cvpr2022mae, li2022A2MIM}. Targeting this problem, one branch of researchers presents lightweight ViTs~\citep{nips2021vitc, iclr2022mobilevit, nips2022EfficientFormer, chen2022CFViT} with efficient attentions~\citep{nips2020linformer}.
Meanwhile, the incorporation of self-attention and convolution as a hybrid backbone has been studied~\citep{guo2021cmt, wu2021cvt, nips2021coatnet, d2021convit, iclr2022uniformer, aaai2022LIT, nips2022iformer} for imparting locality priors to ViTs.
By introducing local inductive bias~\citep{zhu2020deformable, chen2021pre, jiang2021transgan, arnab2021vivit}, advanced training strategies~\citep{icml2021deit, yuan2021tokens, eccv2022deit3} or extra knowledge~\citep{nips2021TL, Lin2022SuperViT, eccv2022tinyvit}, ViTs can achieve superior performance and have been extended to various vision areas.
MetaFormer~\citep{yu2022metaformer} %as shown in Fig. \ref{fig:framework} 
considerably influenced the roadmap of deep architecture design, where all ViTs~\citep{2022convmixer, aaai2022shiftvit} can be classified by the token-mixing strategy, such as relative position encoding~\citep{wu2021rethinking}, local window shifting~\citep{liu2021swin} and MLP layer~\citep{nips2021MLPMixer}, \textit{etc.}

% fig: framework
\begin{figure*}[t]
    \vspace{-2.5em}
    \centering
    \includegraphics[width=0.85\textwidth]{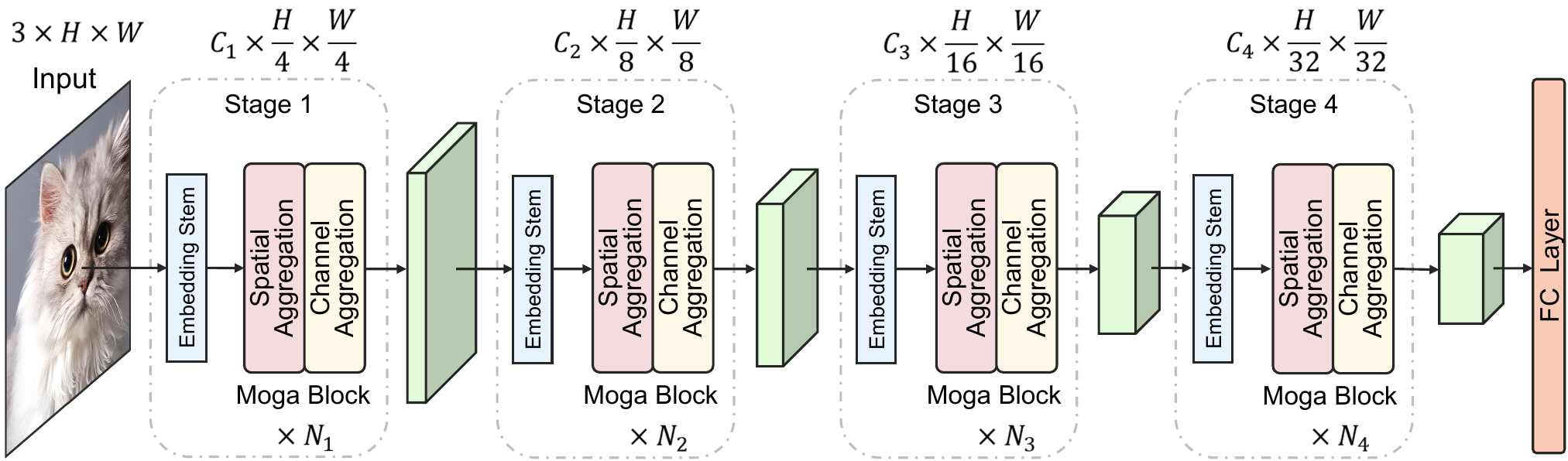}
    \vspace{-0.5em}
    \caption{\textbf{MogaNet architecture with four stages.} Similar to~\citep{liu2021swin, cvpr2022convnext}, MogaNet uses hierarchical architecture of 4 stages. Each stage $i$ consists of an embedding stem and $N_{i}$ Moga Blocks, which contain spatial aggregation blocks and channel aggregation blocks.
    }
    \label{fig:moga_framework}
    \vspace{-1.0em}
\end{figure*}

\vspace{-0.25em}
\subsection{Post-ViT Modern ConvNets}
%Classical ConvNets~\citep{simonyan2014very, xie2017aggregated} fails to capture long-range interactions constrained by their receptive fields.
Taking the merits of ViT-style framework design~\citep{yu2022metaformer}, modern ConvNets~\citep{cvpr2022convnext, Liu2022SLak, nips2022hornet, nips2022focalnet} show superior performance with large kernel depth-wise convolutions~\citep{han2021demystifying} for global perception (\pl{view Appendix~\ref{sec:related_work} for detail backgrounds}).
%Images require operations with local and geometrical inductive biases such as convolutions~\citep{cvpr2016inceptionv3, iclr2016dilated, chollet2017xception}, spatial MLP~\citep{nips2021MLPMixer}, or even non-parametric operations like pooling~\citep{yu2022metaformer} and spatial shifting~\citep{aaai2022shiftvit}.  
%
It primarily comprises three components: (\romannumeral1) embedding stem, (\romannumeral2) spatial mixing block, and (\romannumeral3) channel mixing block.
Embedding stem downsamples the input to reduce redundancies and computational overload. We assume the input feature $X$ %and output $Z$ are in the same shape $\mathbb{R}^{C\times H\times W}$, we have:
\pl{is in the shape} $\mathbb{R}^{C\times H\times W}$, we have:
% \vspace{-0.30em}
\begin{equation}
    Z = \mathrm{Stem}(X),
    \vspace{-0.40em}
\end{equation}
where $Z$ is downsampled features, \textit{e.g.,}. 
Then, the feature flows to a stack of residual blocks. In each stage, the network modules can be decoupled into two separate functional components, $\mathrm{SMixer}(\cdot)$ and $\mathrm{CMixer}(\cdot)$ for spatial-wise and channel-wise information propagation,
\vspace{-0.50em}
\begin{equation}
    \label{eq:smixer}
    Y = X + \mathrm{SMixer}\big(\mathrm{Norm}(X)\big),
    \vspace{-0.50em}
\end{equation}
\begin{equation}
    \label{eq:cmixer}
    Z = Y + \mathrm{CMixer}\big(\mathrm{Norm}(Y)\big),
    \vspace{-0.20em}
\end{equation}
where $\mathrm{Norm}(\cdot)$ denotes a normalization layer, \textit{e.g.,} BatchNorm~\citep{ioffe2015batch} (BN). $\mathrm{SMixer}(\cdot)$ can be various spatial operations (\textit{e.g.,} self-attention, convolution), while $\mathrm{CMixer}(\cdot)$ is usually achieved by channel MLP with inverted bottleneck~\citep{cvpr2018mobilenetv2} and expand ratio $r$.
Notably, we abstract \textit{context aggregation} in modern ConvNets as a series of operations that can \textit{adaptively} aggregate contextual information while suppressing trivial redundancies in spatial mixing block $\mathrm{SMixer}(\cdot)$ between two embedded features:
\vspace{-0.40em}
\begin{equation}\label{eq:aggregate}
    O = \mathcal{S}\big(\mathcal{F}_{\phi}(X), \mathcal{G}_{\psi}(X)\big),
    \vspace{-0.40em}
\end{equation}
where $\mathcal{F}_{\phi}(\cdot)$ and $\mathcal{G}_{\psi}(\cdot)$ are the aggregation and context branches with parameters $\phi$ and $\psi$. Context aggregation models the importance of each position on $X$ by the aggregation branch $\mathcal{F}_{\phi}(X)$ and reweights the embedded feature from the context branch $\mathcal{G}_{\psi}(X)$ by operation $\mathcal{S}(\cdot,\cdot)$.

%\input{Tabs/tab_attention.tex}

%As shown in Table~\ref{tab:attention}, there are mainly two types of context aggregations for modern ConvNets: self-attention mechanism~\citep{vaswani2017attention, wang2018non, iclr2021vit} and gating attention~\citep{dauphin2017language, hu2018squeeze}.
%The importance of each position on $X$ is calculated by global interactions of all other positions in $\mathcal{F}_{\phi}(\cdot)$ with a dot-product, which results in quadratic computational complexity. To overcome this limitation, attention variants in linear complexity~\citep{nips2021SOFT, iclr2022cosFormer} were proposed to substitute vanilla self-attention, \textit{e.g.,} linear attention~\citep{nips2020linformer, nips2022hilo} in the second line of Table~\ref{tab:attention}, but they might degenerate to trivial attentions~\citep{icml2022Flowformer}. Unlike self-attention, gating unit employs an element-wise product $\odot$ as $\mathcal{S}(\cdot,\cdot)$ in linear complexity, \textit{e.g.,} gated linear unit (GLU) variants~\citep{Shazeer2020GLU} and squeeze-and-excitation (SE) modules~\citep{hu2018squeeze} in the last two lines of Table~\ref{tab:attention}. % However, they only aggregate the information of each position or the overall context with global average pooling (GAP), which lacks spatial interactions.

%% file: 3_method.tex
\section{Multi-order Game-theoretic Interaction for Deep Architecture Design}
\begin{wrapfigure}{r}{0.423\linewidth}
    \vspace{-3.0em}
    \hspace{-1.5em}
    \begin{center}
    \includegraphics[width=1.03\linewidth]{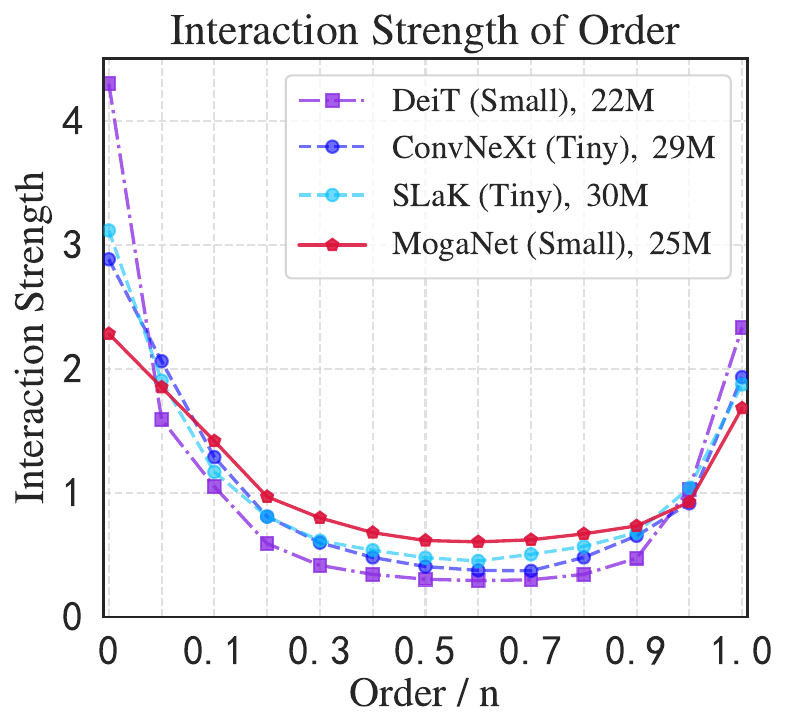}
    \end{center}
    \vspace{-1.5em}
    \caption{
    \textbf{Distributions of the interaction strength $J^{(m)}$} for Transformers and ConvNets on ImageNet-1K with $224^2$ resolutions and $n$ = $14\times 14$. %Middle-order strengths mean the middle-complex interaction, where a medium number of patches (\textit{e.g.,} 0.2$\sim$0.8n) participate.
    }
    \label{fig:spatial_interaction}
    \vspace{-1.5em}
\end{wrapfigure}

\vspace{-0.25em}
\label{sec:rep_bottleneck}
\paragraph{Representation Bottleneck of DNNs}
Recent studies toward the generalizability~\citep{iclr2019shapebias, ancona2019explaining, 2021shapebias, nips2021partial} and robustness~\citep{naseer2021intriguing, icml2022FAN, iclr2022how} of DNNs deliver a new perspective to improve deep architectures.
Apart from them, the investigation of multi-order game-theoretic interaction unveils the representation bottleneck of DNNs.
% As shown in Fig.~\ref{fig:mask}, 
Methodologically, multi-order interactions between two input variables represent the marginal contribution brought by collaborations among these two and other involved contextual variables, where the order indicates the number of contextual variables within the collaboration.
%DNNs can still recognize the target object under extreme occlusion ratios (\textit{e.g.,} only 10$\sim$20\% visible patches) but produce less information gain with intermediate occlusions~\citep{deng2021discovering, naseer2021intriguing}.
%Interestingly, our human brains attain the sharpest knowledge upsurge from images with around 50\% patches, which indicates an intriguing cognition gap between human vision and deep models.
%
Formally, it can be explained by $m$-th order game-theoretic interaction $I^{(m)}(i,j)$ and $m$-order interaction strength $J^{(m)}$, as defined in \citep{zhang2020interpreting, deng2021discovering}.
Considering the image with $n$ patches in total, $I^{(m)}(i,j)$ measures the average interaction complexity between the patch pair $i,j$ over all contexts consisting of $m$ patches, where $0\le m \le n-2$ and the order $m$ reflects the scale of the context involved in the game-theoretic interactions between pixels $i$ and $j$. Normalized by the average of interaction strength, the relative interaction strength $J^{(m)}$ with $m\in (0,1)$ measures the complexity of interactions encoded in DNNs. 
Notably, low-order interactions tend to encode \textbf{common} or \textbf{widely-shared local texture}, and the high-order ones are inclined to forcibly memorize the pattern of \textbf{rare outliers}~\citep{deng2021discovering, cheng2021game}. 
As shown in Fig.~\ref{fig:spatial_interaction}, existing DNNs are implicitly prone to excessively low- or high-order interactions while suppressing the most expressive and versatile middle-order ones~\citep{deng2021discovering, cheng2021game}.
Refer to Appendix~\ref{app:interaction} for definitions and more details.

\vspace{-1.0em}
\paragraph{Multi-order Interaction for Architecture Design.}
Existing deep architecture design is usually derived from intuitive insights, lacking hierarchical theoretical guidance. Multi-order interaction can serve as a reference that fits well with the already gained insights on computer vision and further guides the ongoing quest.
% For instance, the extremely high-order interactions encoded in ViTs (\textit{e.g.}, DeiT in Fig.~\ref{fig:spatial_interaction}) indicate its global-range feature interactions from the self-attention mechanism.
For instance, the extremely high-order interactions encoded in ViTs (\textit{e.g.}, DeiT in Fig.~\ref{fig:spatial_interaction}) \pl{may stem from its adaptive global-range self-attention mechanism.}
Its superior robustness can be attributed to its excessive low-order interactions, representing common and widely shared local patterns.
However, \pl{the absence of locality priors still leaves ViTs lacking middle-order interactions, which cannot be replaced by the low-order ones.}
As for modern ConvNets (\textit{e.g.}, SLaK in Fig.~\ref{fig:spatial_interaction}), despite the $51\times51$ kernel size, it still fails to encode enough expressive interactions (\pl{view more results in Appendix~\ref{app:interaction}}).
Likewise, we argue that such a dilemma may be attributed to the inappropriate composition of convolutional locality priors and global context injections \citep{treisman1980feature, 2021shapebias, li2022A2MIM}. 
A naive combination of self-attention or convolutions can be intrinsically prone to the strong bias of global shape~\citep{nips2021partial, cvpr2022replknet} or local texture~\citep{hermann2020origins}, infusing extreme-order interaction preference to models.
\pl{In MogaNet, we aim to provide an architecture that can \textit{adaptively force the network to encode expressive interactions that would have otherwise been ignored inherently}}.

%%% fix bugs of `Not in outer par mode`.
\begin{figure*}[b!]
    \vspace{-1.5em}
    \centering
    \subfloat[$\mathrm{Moga}(\cdot)$ Block]{\label{fig:moga_multiorder}
    \hspace{-0.5em}
    \includegraphics[width=0.375\linewidth,trim= 0 0 0 0,clip]{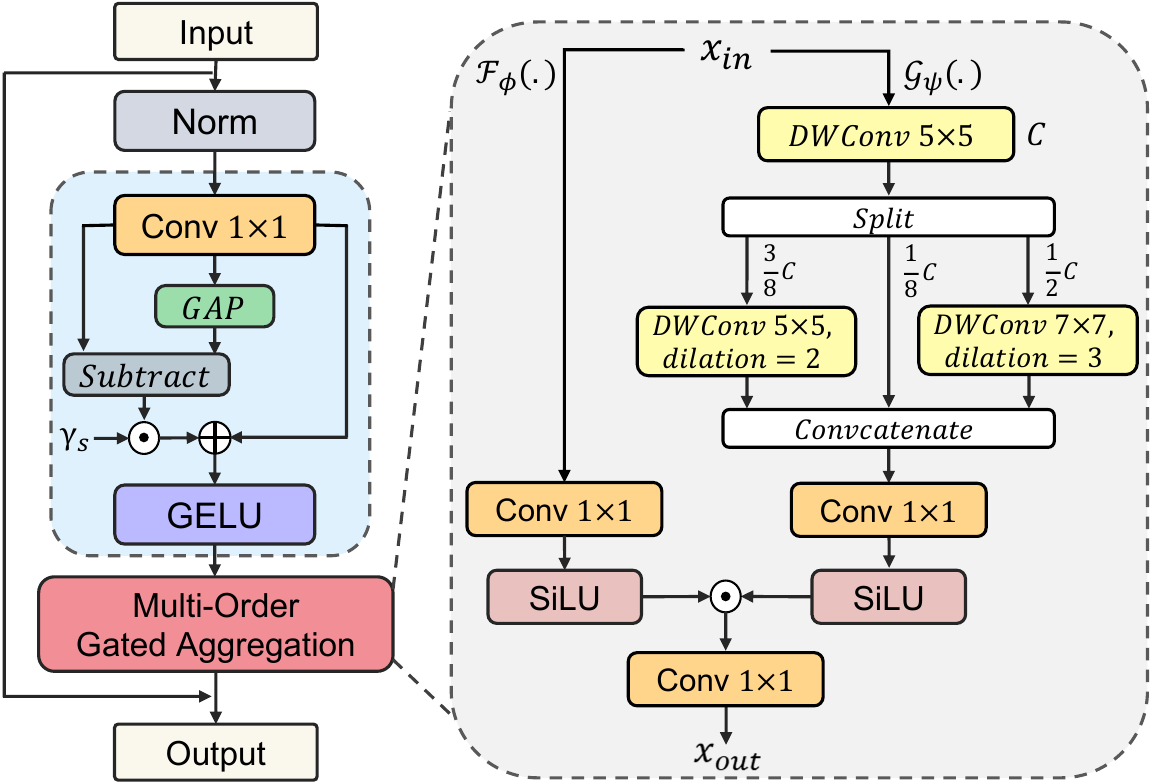}
    }
    \subfloat[$\mathrm{CA}(\cdot)$ Block]{\label{fig:channal_moga}
    \includegraphics[width=0.315\linewidth,trim= 0 0 0 0,clip]{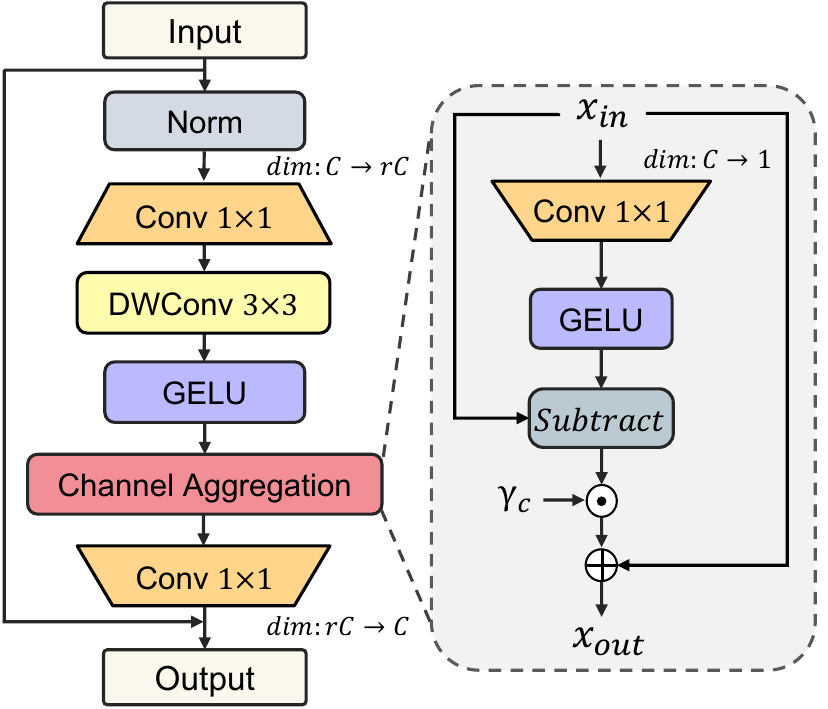}
    }
    \subfloat[]{\label{fig:channal_analysis}
    \hspace{-0.5em}
    \includegraphics[width=0.310\linewidth,trim= 1 8 0 1,clip]{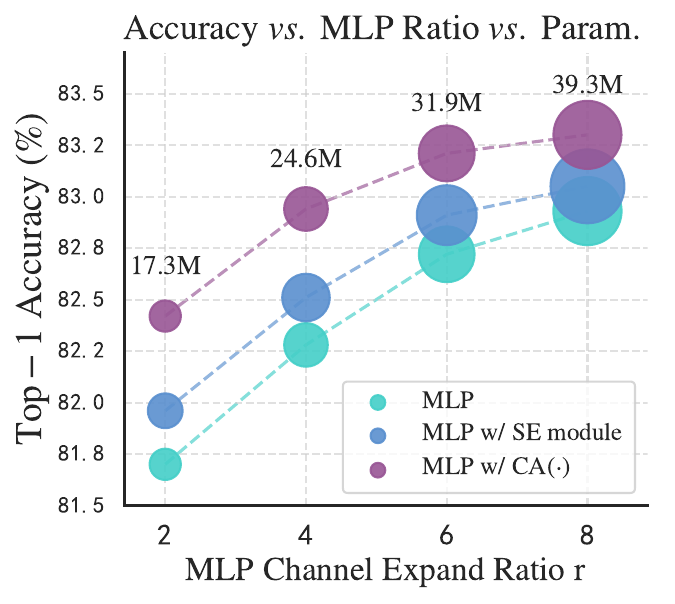}
    }
    \vspace{-0.75em}
    \caption{
    \textbf{(a) Structure of spatial aggregation block $\mathrm{Moga}(\cdot)$.}
    \textbf{(b) Structure of channel aggregation block.}
    \textbf{(c) Analysis of channel MLP and the channel aggregation module.} Based on MogaNet-S, the performances and model sizes of the raw channel MLP, MLP with SE block, and the channel aggregation are compared with the MLP ratio of $\{2,4,6,8\}$ on ImageNet-1K.
    }
    \vspace{-0.5em}
\end{figure*}

\section{Methodology}
\label{sec:method}
% In this section, we instantiate the overall framework as MogaNet equipped with spatial and channel aggregation blocks for multi-order context aggregation and channel-wise multi-order features reallocation.
% In this section, we instantiate the overall framework as MogaNet with convolution-based modules that efficiently combine regionality perception and context aggregation.

\vspace{-0.5em}
\subsection{Overview of MogaNet}
\label{sec:overview}
% Fig.~\ref{fig:app_moga_framework} comprehensively illustrates the four-stage MogaNet architecture.
Built upon modern ConvNets, we design a four-stage MogaNet architecture as illustrated in 
Fig.~\ref{fig:moga_framework}.
For stage $i$, the input image or feature is first fed into an embedding stem to regulate the resolutions and embed into $C_{i}$ dimensions. Assuming the input image in $H\times W$ resolutions, features of the four stages are in $\frac{H}{4}\times\frac{W}{4}$, $\frac{H}{8}\times\frac{W}{8}$, $\frac{H}{16}\times\frac{W}{16}$, and $\frac{H}{32}\times\frac{W}{32}$ resolutions respectively.
Then, the embedded feature flows into $N_{i}$ Moga Blocks, consisting of spatial and channel aggregation blocks (in Sec.~\ref{sec:moga} and \ref{sec:channel}), for further context aggregation. 
After the final output, a GAP and a linear layer are added for classification tasks. As for dense prediction tasks~\citep{2017iccvmaskrcnn, eccv2018upernet}, the output from four stages can be used through neck modules~\citep{cvpr2017fpn, cvpr2019semanticFPN}.

\vspace{-0.5em}
%\subsection{Multi-order Gated Aggregation}
\subsection{Multi-order Spatial Gated Aggregation}
\label{sec:moga}
\begin{figure*}[t!]
\vspace{-3.0em}
\centering
\begin{minipage}{0.61\linewidth}
    % \vspace{-0.5em}
    \includegraphics[width=1.0\linewidth]{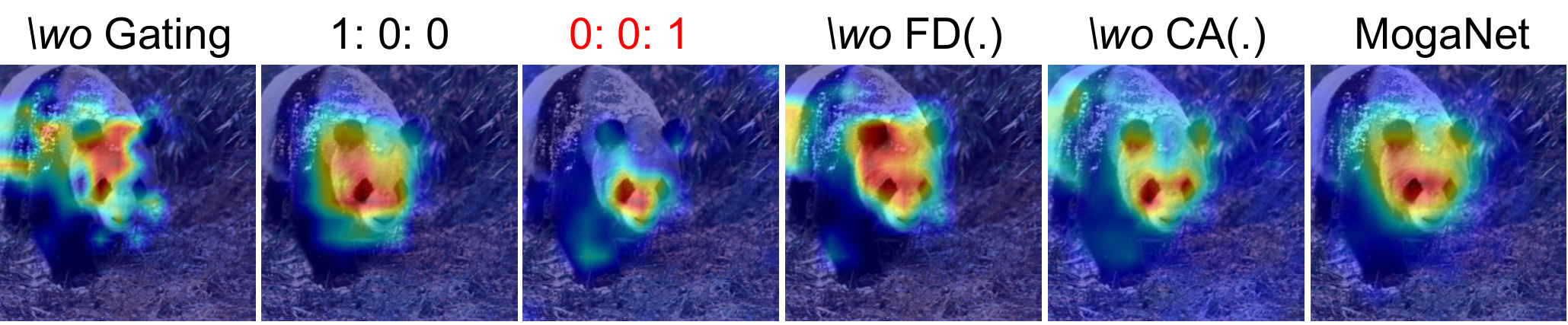}
    \vspace{-1.75em}
    \caption{\textbf{Grad-CAM visualization of ablations}. 1:\ 0:\ 0 and 0:\ 0:\ 1 denote only using $C_l$ or $C_h$ for Multi-order DWConv Layers in SA block. The models encoded extremely low- ($C_l$) or high- ($C_h$) order interactions are sensitive to similar regional textures (1:\ 0:\ 0) or excessive discriminative parts (0:\ 0:\ 1), not localizing precise semantic parts. Gating effectively eliminates the disturbing contextual noise (\textit{$\backslash$wo} Gating).}
    \label{fig:ablation_cam}
\end{minipage}
\begin{minipage}{0.38\linewidth}
    \vspace{-0.5em}
    \input{Tabs/tab_ablation.tex}
    \vspace{-1.0em}
\end{minipage}
\vspace{-1.5em}
\end{figure*}

%However, as discussed in Sec.~\ref{sec:rep_bottleneck}, the sole presence of regionality perception or context aggregation exhibits the inability to learn comprehensive contextual features with multi-order interactions~\citep{iclr2022uniformer, pinto2022impartial, deng2021discovering}.
As discussed in Sec.~\ref{sec:rep_bottleneck}, DNNs with the incompatible composition of locality perception and context aggregation can be implicitly prone to extreme-order game-theoretic interaction strengths while suppressing the more robust and expressive middle-order ones~\citep{iclr2022uniformer, pinto2022impartial, deng2021discovering}.
As shown in Fig.~\ref{fig:ablation_cam}, the primary obstacle pertains to \pl{how to \textbf{force} the network to encode the originally ignored expressive interactions and informative features.}
%Fig.~\ref{fig:interaction} shows conventional DNNs focus on extremely low or high-order interactions. They are missing the most informative middle-order interactions. Thus, the primary challenge is capturing contextual multi-order interactions effectively and efficiently.
%
We first suppose that the essential \textit{adaptive} nature of attention in ViTs has not been well leveraged and grafted into ConvNets. 
Thus, we propose a spatial aggregation (SA) block as an instantiation of $\mathrm{SMixer}(\cdot)$ to learn representations of multi-order interactions in a unified design, as shown in Fig.~\ref{fig:moga_multiorder}, consisting of two cascaded components. We instantiate Eq.~(\ref{eq:smixer}) as:
\vspace{-0.40em}
\begin{equation}
    \label{eq:moga_block}
    Z = X + \mathrm{Moga}\Big( \mathrm{FD}\big(\mathrm{Norm}(X)\big) \Big),
    \vspace{-0.50em}
\end{equation}
where $\mathrm{FD}(\cdot)$ indicates a feature decomposition module (FD) and $\mathrm{Moga}(\cdot)$ denotes a multi-order gated aggregation module comprising the gating $\mathcal{F}_{\phi}(\cdot)$ and context branch $\mathcal{G}_{\psi}(\cdot)$.

\paragraph{Context Extraction.}
\vspace{-1.0em}
As a pure ConvNet structure, we extract multi-order features with both \textit{static} and \textit{adaptive} locality perceptions. 
% Since convolutions are inherently high-pass filters~\citep{iclr2022how, wang2022anti}, we design $\mathrm{FD}(\cdot)$ to adaptively decompose and enhance the high-frequency details to learn more informative features of full-frequency bands, which is formulated as: 
There are two complementary counterparts, fine-grained local texture (low-order) and complex global shape (middle-order), which are instantiated by $\mathrm{Conv}_{1\times 1}(\cdot)$ and $\mathrm{GAP}(\cdot)$ respectively. To \textbf{force} the network against its \pl{implicitly inclined interaction strengths}, we design $\mathrm{FD}(\cdot)$ to adaptively exclude the trivial (overlooked) interactions, defined as:
\vspace{-0.25em}
\begin{align}
    \label{eq:FD_proj}
    Y &= \mathrm{Conv}_{1\times 1}(X),\\
    \label{eq:FD}
    Z &= \mathrm{GELU}\Big(Y + \gamma_{s}\odot\big( Y-\mathrm{GAP}(Y)\big) \Big),
    \vspace{-1.5em}
\end{align}
where ${\gamma}_{s} \in \mathbb{R}^{C\times 1}$ denotes a scaling factor initialized as zeros. By re-weighting the complementary interaction component $Y - \mathrm{GAP}(Y)$, $\mathrm{FD}(\cdot)$ also increases spatial feature diversities~\citep{iclr2022how, wang2022anti}.
Then, we ensemble depth-wise convolutions (DWConv) to encode multi-order features in the context branch of $\mathrm{Moga}(\cdot)$. Unlike previous works that simply combine DWConv with self-attentions to model local and global interactions~\citep{eccv2022edgeformer, nips2022hilo, nips2022iformer, nips2022hornet} , we employ three different DWConv layers with dilation ratios $d\in \{1,2,3\}$ in parallel to capture low, middle, and high-order interactions: given the input feature $X\in \mathbb{R}^{C\times HW}$, $\mathrm{DW}_{5\times 5, d=1}$ is first applied for low-order features; then, the output is factorized into ${X}_l \in \mathbb{R}^{C_l \times HW}$, ${X}_m \in \mathbb{R}^{C_m \times HW}$, and ${X}_h \in \mathbb{R}^{C_h \times HW}$ along the channel dimension, where $C_l + C_m + C_h =C$; afterward, ${X}_m$ and ${X}_h$ are assigned to $\mathrm{DW}_{5\times 5, d=2}$ and $\mathrm{DW}_{7\times 7, d=3}$, respectively, while ${X}_l$ serves as identical mapping; finally, the output of ${X}_l$, ${X}_m$, and ${X}_h$ are concatenated to form multi-order contexts, $Y_{C} = \mathrm{Concat}(Y_{l, 1:C_{l}}, Y_{m}, Y_{h})$. 
Notice that the proposed $\mathrm{FD}(\cdot)$ and multi-order DWConv layers only require a little extra computational overhead and parameters in comparison to $\mathrm{DW}_{7\times 7}$ used in ConvNeXt~\citep{cvpr2022convnext}, \textit{e.g.,} +multi-order and +$\mathrm{FD}(\cdot)$ increase 0.04M parameters and 0.01G FLOPS over $\mathrm{DW}_{7\times 7}$ as shown in Table~\ref{tab:ablation}.

\vspace{-1.0em}
\paragraph{Gated Aggregation.}
To \textit{adaptively} aggregate the extracted feature from the context branch, we employ SiLU~\citep{elfwing2018sigmoid} activation in the gating branch, \textit{i.e.,} $x\cdot \mathrm{Sigmoid}(x)$, which has been well-acknowledged as an advanced version of Sigmoid activation. 
As illustrated in Appendix~\ref{app:ablation_gating}, we empirically show that SiLU in MogaNet exhibits both the gating effects as Sigmoid and the stable training property.
Taking the output from $\mathrm{FD}(\cdot)$ as the input, we instantiate Eq.~(\ref{eq:aggregate}):
\vspace{-0.25em}
\begin{align}
    \label{eq:moga}
    Z &= \underbrace{\mathrm{SiLU}\big( \mathrm{Conv}_{1\times 1}(X) \big)}_{\mathcal{F}_{\phi}} \odot \underbrace{\mathrm{SiLU}\big( \mathrm{Conv}_{1\times 1}(Y_{C}) \big)}_{\mathcal{G}_{\psi}},
    \vspace{-1.5em}
\end{align}
% where $\mathcal{G}_{\psi}(\cdot)$ and $\mathcal{F}_{\phi}(\cdot)$ are defined as $\mathrm{SiLU}(\mathrm{Conv}_{1\times 1}(\cdot))$.
With the proposed SA blocks, MogaNet captures more middle-order interactions, as validated in Fig.~\ref{fig:spatial_interaction}. 
The SA block produces discriminative multi-order representations with similar parameters and FLOPs as $\mathrm{DW}_{7\times 7}$ in ConvNeXt, which is well beyond the reach of existing methods without the cost-consuming self-attentions.

\begin{wrapfigure}{r}{0.49\linewidth}
    \vspace{-1.5em}
    \begin{center}
    \includegraphics[width=1.0\linewidth]{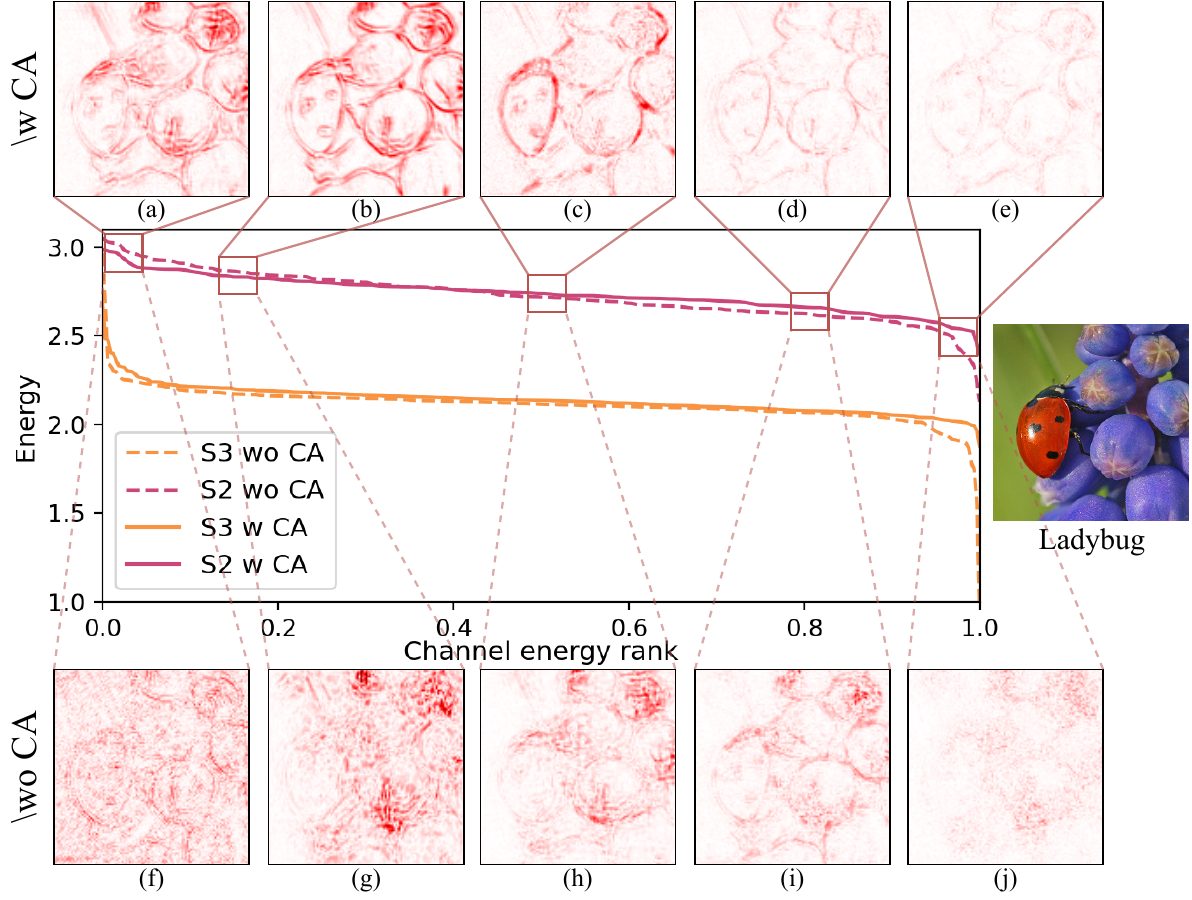}
    \end{center}
    \vspace{-1.5em}
    \caption{
    \textbf{Channel energy ranks and channel saliency maps (CSM)}~\citep{cvpr2022ReflashDIS} with or without our CA block based on MogaNet-S. The energy reflects the importance of the channel, while the highlighted regions of CSMs are the activated spatial features of each channel.
    }
    \label{fig:channel_en_saliency}
    \vspace{-1.5em}
\end{wrapfigure}

%\subsection{Multi-order Feature Reallocation by Channel Aggregation}
\subsection{Multi-order Channel Reallocation}
\vspace{-0.25em}
\label{sec:channel}
Prevalent architectures, as illustrated in Sec.~\ref{sec:preliminaries}, perform channel-mixing $\mathrm{CMixer}(\cdot)$ mainly by two linear projections, \textit{e.g.,} 2-layer channel-wise MLP~\citep{iclr2021vit, liu2021swin, nips2021MLPMixer} with a expand ratio $r$ or the MLP with a $3\times 3$ DWConv in between~\citep{cvmj2022PVTv2, aaai2022LIT, nips2022hilo}.
Due to the information redundancy cross channels~\citep{eccv2018CBAM, iccv2019GCNet, icml2019efficientnet, cvpr2020Orthogonal}, vanilla MLP requires a number of parameters ($r$ default to 4 or 8) to achieve expected performance, showing low computational efficiency as plotted in Fig.~\ref{fig:channal_analysis}. 
To address this issue, most current methods directly insert a channel enhancement module, \textit{e.g.,} SE module~\citep{hu2018squeeze}, into the MLP.
%This problem might arise from the inherent redundancy cross channels~\citep{eccv2018CBAM, iccv2019GCNet, icml2019efficientnet, cvpr2020Orthogonal}, and most current methods try to address this issue by starchedly inserting a channel enhancement module, %\textit{e.g.,} SE module~\citep{hu2018squeeze}, into MLP.
Unlike these designs requiring additional MLP bottleneck, \pl{motivated by $\mathrm{FD}(\cdot)$,} we introduce a lightweight channel aggregation module $\mathrm{CA}(\cdot)$ to adaptive reallocate channel-wise features in high-dimensional hidden spaces and further extend it to a channel aggregation (CA) block. 
% Unlike these attempts, we introduce a lightweight channel aggregation module $\mathrm{CA}(\cdot)$ to conduct adaptive channel-wise reallocation in high-dimensional hidden spaces and further extend it to a channel aggregation (CA) block. 
As shown in Fig.~\ref{fig:channal_moga}, we rewrite Eq.~(\ref{eq:cmixer}) for our CA block as:
\vspace{-0.5em}
\begin{equation}
\begin{aligned}
    Y &= \mathrm{GELU}\Big(\mathrm{DW_{3\times 3}}\big(\mathrm{Conv_{1\times 1}}(\mathrm{Norm}(X))\big)\Big),\\
    Z &= \mathrm{Conv_{1 \times 1}}\big(\mathrm{CA}(Y)\big) + X.
\end{aligned}
\vspace{-0.4em}
\end{equation}
Concretely, $\mathrm{CA}(\cdot)$ is implemented by a channel-reducing projection $W_{r}: \mathbb{R}^{C\times HW}\rightarrow \mathbb{R}^{1\times HW}$ and GELU to gather and reallocate channel-wise information:
\vspace{-0.35em}
\begin{equation}
    \mathrm{CA}(X) = X + \gamma_{c}\odot\big(X - \mathrm{GELU}(XW_{r})\big),
\vspace{-0.25em}
\end{equation}
where $\gamma_{c}$ is the channel-wise scaling factor initialized as zeros. It reallocates the channel-wise feature \pl{with} the complementary interactions $(X - \mathrm{GELU}(XW_{r}))$. As shown in Fig.~\ref{fig:ablation_interaction}, $\mathrm{CA}(\cdot)$ enhances \pl{originally overlooked} game-theoretic interactions.
Fig.~\ref{fig:channal_analysis} and Fig.~\ref{fig:channel_en_saliency} verify the effectiveness of $\mathrm{CA}(\cdot)$ compared with vanilla MLP and MLP with SE module in channel-wise efficiency and representation ability. Despite some improvements to the baseline, the MLP $w/$ SE module still requires large MLP ratios (\textit{e.g.,} $r$ = 6) to achieve expected performance while bringing extra parameters and overhead. 
Yet, our $\mathrm{CA}(\cdot)$ with $r$ = 4 brings 0.6\% gain over the baseline at a small extra cost (0.04M extra parameters \& 0.01G FLOPs) while achieving the same performance as the baseline with $r$ = 8.
% It is implemented by various attention mechanisms in recent vision Transformer models or spatial Multi-layer Perceptron (MLP) in MLP-like models. Note that the main function of the token mixer is to propagate token information, although some token mixers can also mix channels, like attention. At Stage II, the second sub-block primarily consists of a two-layered MLP with non-linear activation, looking at cross-channel correlations via a set of $1\times 1$ convolutions. 

\subsection{Implementation Details}
\label{sec:details}
Following the network design style of ConvNets~\citep{cvpr2022convnext}, we scale up MogaNet for six model sizes (X-Tiny, Tiny, Small, Base, Large, and X-Large) via stacking the different number of spatial and channel aggregation blocks at each stage, which has similar numbers of parameters as RegNet~\citep{cvpr2020regnet} variants. Network configurations and hyper-parameters are detailed in Table~\ref{tab:app_architecture}. FLOPs and throughputs are analyzed in Appendix \ref{app:flops_throughput}.
We set the channels of the multi-order DWConv layers to $C_l:$ $C_m:$ $C_h$ = 1:3:4 (see 
 Appendix~\ref{app:ablation_multiorder}).
% The embedding stem in canonical ResNet contains a 7$\times$7 convolution layer with stride 2, followed by a max-pooling layer, which results in a 4$\times$ downsampling of the input images.
Similar to \citep{2021patchconvnet, iclr2022uniformer, nips2022EfficientFormer}, the first embedding stem in MogaNet is designed as two stacked 3$\times$3 convolution layers with a stride of 2 while adopting the single-layer version for embedding stems in the other three stages.
We select GELU~\citep{hendrycks2016bridging} as the common activation function and only use SiLU in the Moga module as Eq.~(\ref{eq:moga}).

%% file: Tabs/tab_ablation.tex
\begin{table}[H]
    \setlength{\tabcolsep}{0.4mm}
    \centering
\resizebox{\linewidth}{!}{
    \begin{tabular}{ll|ccc}
    \toprule
\multicolumn{2}{l|}{Modules}                                                     & Top-1     & Params. & FLOPs \\
\multicolumn{2}{l|}{}                                                            & Acc (\%)  & (M)     & (G)   \\ \hline
\multicolumn{2}{l|}{Baseline}                                                    & 76.6      & 4.75    & 1.01  \\ \hline
\multirow{4}{*}{SMixer} & \cellcolor{gray94}+Gating branch                       & 77.3      & 5.09    & 1.07  \\
                        & +$\mathrm{DW}_{7\times 7}$                             & 77.5      & 5.14    & 1.09  \\
                        & \cellcolor{gray94}+Multi-order $\mathrm{DW}(\cdot)$    & 78.0      & 5.17    & 1.10  \\
                        & \cellcolor{gray94}+$\mathrm{FD(\cdot)}$                & 78.3      & 5.18    & 1.10  \\
\multirow{2}{*}{CMixer} & +SE module                                             & 78.6      & 5.29    & 1.14  \\
                        & \cellcolor{gray94}+$\mathrm{CA}(\cdot)$                & \bf{79.0} & 5.20    & 1.10  \\
    \bottomrule
    \end{tabular}
    }
    \vspace{-0.75em}
    \caption{\textbf{Ablation of designed modules on ImageNet-1K}. The baseline uses the non-linear projection and $\mathrm{DW}_{5\times 5}$ as $\mathrm{SMixer}(\cdot)$ and the vanilla MLP as $\mathrm{CMixer}(\cdot)$.
    % \hl{Gray} denotes the designed modules. Compared to $\mathrm{DW}_{7\times 7}$ and SE modules, Multi-order $\mathrm{DW}(\cdot)$+$\mathrm{FD(\cdot)}$ as multi-order regionality perceptions and $\mathrm{CA}(\cdot)$ bring significant improvements with a little extra cost.
    }
    \label{tab:ablation}
    % \vspace{-1.0em}
\end{table}

%% file: 4_experiments.tex
% \begin{figure*}[t!]
% \vspace{-3.0em}
% \centering
% \begin{minipage}{0.49\linewidth}
%     \centering
%     \input{Tabs/tab_in1k_tiny.tex}
%     \vspace{-2.25em}
%     \input{Tabs/tab_in1k}
% \end{minipage}
% ~\begin{minipage}{0.49\linewidth}
%     \centering
%     \input{Tabs/tab_coco.tex}
%     \vspace{-2.25em}
%     \input{Tabs/tab_ade20k.tex}
%     \vspace{-2.25em}
%     \input{Tabs/tab_coco_pose}
%     \vspace{-1.25em}
%     % \input{Tabs/tab_3d_vp}
% \end{minipage}
% \vspace{-3.0em}
% \end{figure*}

\section{Experiments}
\label{sec:expriments}
To impartially evaluate and compare MogaNet with the leading network architectures, we conduct extensive experiments across various popular vision tasks, including image classification, object detection, instance and semantic segmentation, 2D and 3D pose estimation, and video prediction. 
The experiments are implemented with PyTorch and run on NVIDIA A100 GPUs.
% To verify the effectiveness of our method, we conduct extensive experiments on ImageNet-1K \citep{cvpr2009imagenet} for image classification, COCO~\citep{2014MicrosoftCOCO} for object detection, instance segmentation, and pose estimation, and ADE20K~\citep{Zhou2018ADE20k} for semantic segmentation. All experiments are implemented with PyTorch on Ubuntu workstations with NVIDIA A100 GPUs. \textbf{Bold} and \hl{gray} indicate the best performance and our models.

\subsection{ImageNet Classification}
\label{sec:exp_in1k}
\vspace{-0.25em}
\paragraph{Settings.} 
For classification experiments on ImageNet \citep{cvpr2009imagenet}, we train our MogaNet following the standard procedure \citep{icml2021deit, liu2021swin} on ImageNet-1K (IN-1K) for a fair comparison, training 300 epochs with AdamW~\citep{iclr2019AdamW} optimizer, a basic learning rate of $1\times 10^{-3}$, and a cosine scheduler~\citep{loshchilov2016sgdr}. To explore the large model capacities, we pre-trained MogaNet-XL on ImageNet-21K (IN-21K) for 90 epochs and then fine-tuned 30 epochs on IN-1K following \citep{cvpr2022convnext}. Appendix~\ref{app:in1k_settings} and \ref{app:exp_in1k} provide implementation details and more results.
We compare three classical architectures: \textbf{Pure ConvNets} (C), \textbf{Transformers} (T), and \textbf{Hybrid model} (H) with both self-attention and convolution operations.
% We compare four typical architectures: (\romannumeral1) \textbf{Pure ConvNets} (C) include ResNet, ShuffleNetV2, EfficientNet, MobileNetV3, RegNet, ConvNeXt, RepLKNet, FocalNet, SLak, and HorNet. (\romannumeral2) \textbf{Transformers} (T) include DeiT, Swin, T2T-ViT, PVT, PVTV2 Focal, ViT-C, CSWin, SReT, and LiTV2. (\romannumeral3) \textbf{Hybrid architectures} (H) of attention and convolution include PiT, LeViT, CoaT, BoTNet, ViTAE, Twins, CoAtNet, MobileViT, Uniformer, Mobile-Former, ParC-Net, EfficientFormer, and MaxViT.

\vspace{-0.5em}
\paragraph{Results.}
With regard to the lightweight models, Table~\ref{tab:in1k_cls_tiny} shows that MogaNet-XT/T significantly outperforms existing lightweight architectures with a more efficient usage of parameters and FLOPs. 
MogaNet-T achieves 79.0\% top-1 accuracy, which improves models with $\sim$5M parameters by at least 1.1 at $224^2$ resolutions. Using $256^2$ resolutions, MogaNet-T outperforms the current SOTA ParC-Net-S by 1.0 while achieving 80.0\% top-1 accuracy with the refined settings. Even with only 3M parameters, MogaNet-XT still surpasses models with around 4M parameters, \textit{e.g.,} +4.6 over T2T-ViT-7. Particularly, MogaNet-T$^{\S}$ achieves 80.0\% top-1 accuracy using $256^2$ resolutions and the refined training settings (detailed in Appendix~\ref{app:advanced_tiny}).
% which adjusts $lr$ and replaces RandAugment~\citep{cubuk2020randaugment} with 3-Augment~\citep{eccv2022deit3}
As for scaling up models in Table~\ref{tab:in1k_cls_scaling}, MogaNet shows superior or comparable performances to SOTA architectures with similar parameters and computational costs.
For example, MogaNet-S achieves 83.4\% top-1 accuracy, outperforming Swin-T and ConvNeXt-T with a clear margin of 2.1 and 1.2. MogaNet-B/L also improves recently proposed ConvNets with fewer parameters, \textit{e.g.,} +0.3/0.4 and +0.5/0.7 points over HorNet-S/B and SLaK-S/B.
% As for 45M and 80M models, we summarize their performances in Table~\ref{tab:in1k_cls_scaling} and MogaNet-B/L still surpass the current state-of-the-art architectures, especially improving Swin-S/B and ConvNeXt-S/B by 1.2\%/ 1.1\% and 1.1\%/ 0.8\%.
When pre-trained on IN-21K, MogaNet-XL is boosted to 87.8\% top-1 accuracy with 181M parameters, saving 169M compared to ConvNeXt-XL. Noticeably, MogaNet-XL can achieve 85.1\% at $224^2$ resolutions without pre-training and improves ConvNeXt-L by 0.8, indicating MogaNets are easier to converge than existing models (also verified in Appendix~\ref{app:exp_in1k}).
% Additionally, we provide fast training results based on RSB A3 100-epoch scheme~\citep{wightman2021rsb} in Appendix~\ref{app:comparison}.

\subsection{Dense Prediction Tasks}
\label{sec:exp_det_seg}
\vspace{-0.25em}
\paragraph{Object detection and segmentation on COCO.}
We evaluate MogaNet for object detection and instance segmentation tasks on COCO~\citep{2014MicrosoftCOCO} with RetinaNet~\citep{iccv2017retinanet}, Mask-RCNN~\citep{2017iccvmaskrcnn}, and Cascade Mask R-CNN~\citep{tpami2019cascade} as detectors. Following the training and evaluation settings in \citep{liu2021swin, cvpr2022convnext}, we fine-tune the models by the AdamW optimizer for $1\times$ and $3\times$ training schedule on COCO~\textit{train2017} and evaluate on COCO~\textit{val2017}, implemented on 
MMDetection~\citep{mmdetection} codebase. The box mAP (AP$^{b}$) and mask mAP (AP$^{m}$) are adopted as metrics. Refer to Appendix~\ref{app:coco_det_settings} and \ref{app:exp_det_coco} for detailed settings and full results. Table~\ref{tab:coco} shows that detectors with MogaNet variants significantly outperform previous backbones. It is worth noticing that Mask R-CNN with MogaNet-T achieves 42.6 AP$^{b}$, outperforming Swin-T by 0.4 with 48\% and 27\% fewer parameters and FLOPs. Using advanced training settings and IN-21K pre-trained weights, Cascade Mask R-CNN with MogaNet-XL achieves 56.2 AP$^{b}$, +1.4 and +2.3 over ConvNeXt-L and RepLKNet-31L. 
% MogaNet-T gains 3.6\% AP$^{bb}$ and 4.6\% AP$^{mk}$ over ResNet-18; MogaNet-S outperforms Swin-T (Transformers) by 3.9\% AP$^{bb}$ and 2.7\% AP$^{mk}$, and surpasses UniFormer-S (hybrid) by 0.5\% AP$^{bb}$; MogaNet-B outperforms Swin-T and LITV2-M (Transformer) by 2.9\% AP$^{bb}$ and 1.2\% AP$^{mk}$ respectively.

\begin{figure*}[t!]
\vspace{-3.0em}
\centering
\begin{minipage}{0.49\linewidth}
    \centering
    \input{Tabs/tab_in1k_tiny.tex}

    \vspace{-2.25em}
    \input{Tabs/tab_in1k}
\end{minipage}
~\begin{minipage}{0.49\linewidth}
    \centering
    \input{Tabs/tab_coco.tex}
    \vspace{-2.25em}
    \input{Tabs/tab_ade20k.tex}
    \vspace{-2.25em}
    \input{Tabs/tab_coco_pose}
    \vspace{-1.25em}
\end{minipage}
\vspace{-3.0em}
\end{figure*}

\vspace{-0.75em}
\paragraph{Semantic segmentation on ADE20K.}
We also evaluate MogaNet for semantic segmentation tasks on ADE20K \citep{Zhou2018ADE20k} with Semantic FPN \citep{cvpr2019semanticFPN} and UperNet \citep{eccv2018upernet} following \citep{liu2021swin, yu2022metaformer}, implemented on MMSegmentation~\citep{mmseg2020} codebase. The performance is measured by single-scale mIoU. Initialized by IN-1K or IN-21K pre-trained weights, Semantic FPN and UperNet are fine-tuned for 80K and 160K iterations by the AdamW optimizer. See Appendix \ref{app:ade20k_seg_settings} and \ref{app:exp_seg_ade20k} for detailed settings and full results.
In Table~\ref{tab:ade20k}, Semantic FPN with MogaNet-S consistently outperforms Swin-T and Uniformer-S by 6.2 and 1.1 points; UperNet with MogaNet-S/B/L improves ConvNeXt-T/S/B by 2.5/1.4/1.8 points. Using higher resolutions and IN-21K pre-training, MogaNet-XL achieves 54.0 SS mIoU, surpassing ConvNeXt-L and RepLKNet-31L by 0.3 and 1.6.
% UperNet with MogaNet-S improves backbones of Transformers (+3.1\% over Swin-T), hybrid architectures (+1.6\% over UniFormer-S), and modern ConvNets (+1.1\% over HorNet-T$_{7\times 7}$. Refer to Appendix~\ref{app:ade20k_seg_settings} for more details.

\vspace{-0.75em}
\paragraph{2D and 3D Human Pose Estimation.}
We evaluate MogaNet on 2D and 3D human pose estimation tasks. As for 2D key points estimation on COCO, we conduct evaluations with SimpleBaseline~\citep{eccv2018simple} following \citep{iccv2021PVT, iclr2022uniformer}, which fine-tunes the model for 210 epochs by Adam optimizer \citep{iclr2014adam}. Table \ref{tab:coco_pose} shows that MogaNet variants yield at least 0.9 AP improvements for $256\times 192$ input, \textit{e.g.,} +2.5 and +1.2 over Swin-T and PVTV2-B2 by MogaNet-S. Using $384\times 288$ input, MogaNet-B outperforms Swin-L and Uniformer-B by 1.0 and 0.6 AP with fewer parameters.
As for 3D face/hand surface reconstruction tasks on Stirling/ESRC 3D \citep{feng2018evaluation} and FreiHAND \citep{iccv2019freihand} datasets, we benchmark backbones with ExPose \citep{eccv2020ExPose}, which fine-tunes the model for 100 epochs by Adam optimizer. 3DRMSE and Mean Per-Joint Position Error (PA-MPJPE) are the metrics. In Table \ref{tab:3d_vp}, MogaNet-S shows the lowest errors compared to Transformers and ConvNets.
We provide detailed implementations and results for 2D and 3D pose estimation tasks in Appendix \ref{app:exp_2d_pose} and \ref{app:exp_3d_pose}.

\input{Tabs/tab_3d_vp}
\vspace{-0.75em}
\paragraph{Video Prediction.}
We further objectively evaluate MogaNet for unsupervised video prediction tasks with SimVP~\citep{cvpr2022simvp} on MMNIST \citep{icml2015mmnist}, where the model predicts the successive 10 frames with the given 10 frames as the input. We train the model for 200 epochs from scratch with the Adam optimizer and evaluate it by MSE and Structural Similarity Index (SSIM). Table \ref{tab:3d_vp} shows that SimVP with MogaNet blocks improves the baseline by 6.58 MSE and outperforms ConvNeXt and HorNet by 1.37 and 4.07 MSE. Appendix \ref{app:mmnist_vp_settings} and \ref{app:exp_vp_mmnist} show more experiment settings and results.

\begin{figure*}[t!]
    \vspace{-3.0em}
\begin{minipage}{0.49\linewidth}
    \vspace{-0.5em}
    \begin{figure}[H]
    \centering
    \begin{minipage}{0.39\linewidth}
        \vspace{-1.5em}
        \centering
        \input{Tabs/tab_ablation_small.tex}
    \end{minipage}
    \hspace{-0.5em}
    \begin{minipage}{0.60\linewidth}
        \centering
        \includegraphics[width=1.0\linewidth,trim= 4 0 0 0,clip]{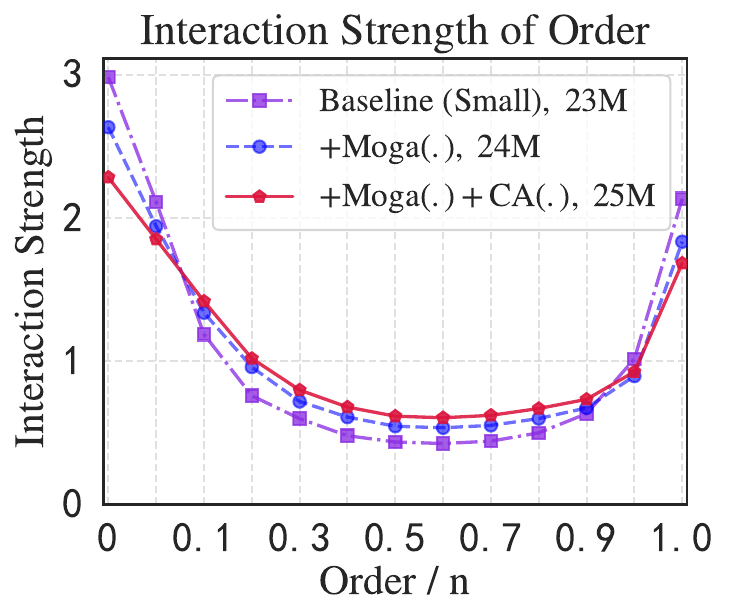}
        \vspace{-2.25em}
    \end{minipage}
        \caption{
        \textbf{Ablation of proposed modules} on IN-1K \textbf{Left}: the table ablates MogaNet modules by removing each of them based on the baseline of MogaNet-S. \textbf{Right}: the figure plots distributions of interaction strength $J^{(m)}$, which verifies that $\mathrm{Moga}(\cdot)$ and $\mathrm{CA}(\cdot)$ both contributes to learning multi-order interactions and better performance.
        }
        \label{fig:ablation_interaction}
    \vspace{-0.5em}
    \end{figure}
\end{minipage}
~\begin{minipage}{0.50\linewidth}
    \begin{figure}[H]
        % \vspace{-0.5em}
        \centering
        \includegraphics[width=1.0\linewidth,trim= 4 0 0 0,clip]{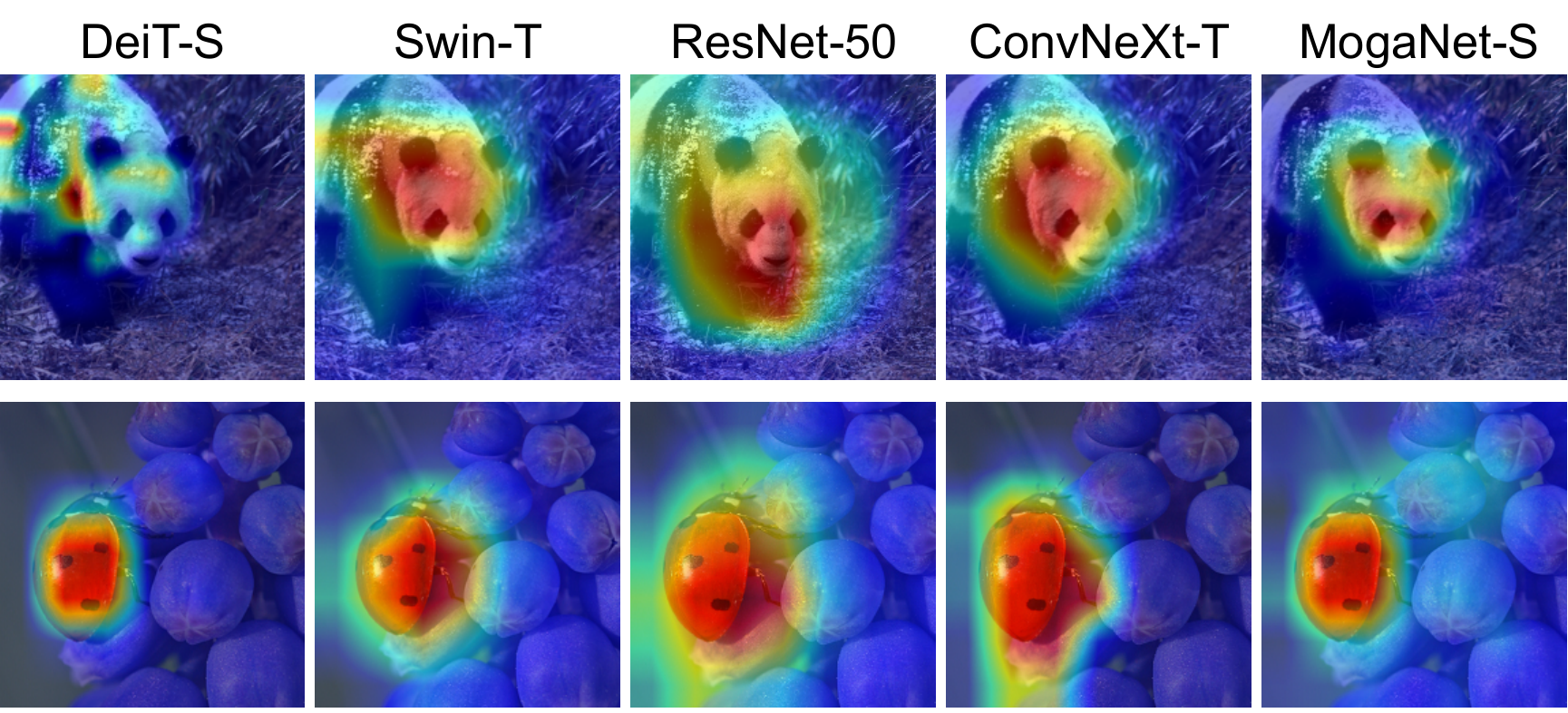}
        \vspace{-1.5em}
        \caption{
        \textbf{Grad-CAM activation maps} on IN-1K. MogaNet exhibits similar activation maps to attention architectures (Swin), which are located on the semantic targets. Unlike previous ConvNets that might activate some irrelevant regions, activation maps of MogaNet are more semantically gathered. See more results in Appendix~\ref{app:gradcam}.
        }
        \label{fig:analysis_gradcam}
    \end{figure}
\end{minipage}
\vspace{-1.25em}
\end{figure*}

\vspace{-0.25em}
\subsection{Ablation and Analysis}
\label{sec:exp_ablation}
\vspace{-0.25em}
We first ablate the spatial aggregation module and the channel aggregation module \textbf{CA}$(\cdot)$ in Table~\ref{tab:ablation} and Fig.~\ref{fig:ablation_interaction} (left). Spatial modules include \textbf{FD$(\cdot)$} and \textbf{Moga$(\cdot)$}, containing the \textbf{gating branch} and the context branch with multi-order DWConv layers \textbf{Multi-DW$(\cdot)$}. We found that all proposed modules yield improvements with favorable costs. Appendix~\ref{app:ablation} provides more ablation studies.
Furthermore, Fig.~\ref{fig:ablation_interaction} (right) empirically shows design modules can learn more middle-order interactions, and Fig.~\ref{fig:analysis_gradcam} visualizes class activation maps by Grad-CAM~\citep{cvpr2017grad} compared to existing models.

%% file: Tabs/tab_in1k_tiny.tex
\begin{table}[H]
    % \vspace{-0.25em}
    \setlength{\tabcolsep}{0.3mm}
    \centering
\resizebox{\linewidth}{!}{
\begin{tabular}{llccccc}
    \toprule
    Architecture                            & Date         & Type & Image   & Param. & FLOPs & Top-1     \\
                                            &              &      & Size    & (M)    & (G)   & Acc (\%)  \\ \hline
    ResNet-18                               & CVPR'2016    & C    & $224^2$ & 11.7   & 1.80  & 71.5      \\
    ShuffleNetV2~$2\times$                  & ECCV'2018    & C    & $224^2$ & 5.5    & 0.60  & 75.4      \\
    EfficientNet-B0                         & ICML'2019    & C    & $224^2$ & 5.3    & 0.39  & 77.1      \\
    % MobileNetV3~$1\times$                   & ICCV'2019    & C    & $224^2$ & 5.4    & 0.23  & 75.2      \\
    % RegNetY-400MF                          & CVPR'2020    & C    & $224^2$ & 5.3    & 0.40  & 74.1      \\
    RegNetY-800MF                           & CVPR'2020    & C    & $224^2$ & 6.3    & 0.80  & 76.3      \\
    DeiT-T$^\dag$                           & ICML'2021    & T    & $224^2$ & 5.7    & 1.08  & 74.1      \\
    % DeiT-2G                                & ICML'2021    & T    & $224^2$ & 13.2   & 1.90  & 75.1      \\
    PVT-T                                   & ICCV'2021    & T    & $224^2$ & 13.2   & 1.60  & 75.1      \\
    T2T-ViT-7                               & ICCV'2021    & T    & $224^2$ & 4.3    & 1.20  & 71.7      \\
    % T2T-ViT-12                             & ICCV'2021    & T    & $224^2$ & 6.9    & 1.80  & 76.5      \\
    ViT-C                                   & NIPS'2021    & T    & $224^2$ & 4.6    & 1.10  & 75.3      \\
    SReT-T$_{Distill}$                      & ECCV'2022    & T    & $224^2$ & 4.8    & 1.10  & 77.6      \\
    PiT-Ti                                  & ICCV'2021    & H    & $224^2$ & 4.9    & 0.70  & 74.6      \\
    LeViT-S                                 & ICCV'2021    & H    & $224^2$ & 7.8    & 0.31  & 76.6      \\
    CoaT-Lite-T                             & ICCV'2021    & H    & $224^2$ & 5.7    & 1.60  & 77.5      \\
    Swin-1G                                 & ICCV'2021    & H    & $224^2$ & 7.3    & 1.00  & 77.3      \\
    % Swin-2G                                 & ICCV'2021    & H    & $224^2$ & 12.8   & 2.00  & 79.3      \\
    % MobileViT-XS                            & ICLR'2022    & H    & $256^2$ & 2.3    & 1.73  & 74.8      \\
    MobileViT-S                             & ICLR'2022    & H    & $256^2$ & 5.6    & 4.02  & 78.4      \\
    % MobileFormer-151M                      & CVPR'2022    & H    & $224^2$ & 7.6    & 0.29  & 75.2      \\
    MobileFormer-294M                       & CVPR'2022    & H    & $224^2$ & 11.4   & 0.59  & 77.9      \\
    ConvNext-XT                             & CVPR'2022    & C    & $224^2$ & 7.4    & 0.60  & 77.5      \\
    VAN-B0                                  & CVMJ'2023   & C    & $224^2$ & 4.1    & 0.88  & 75.4      \\
    ParC-Net-S                              & ECCV'2022    & C    & $256^2$ & 5.0    & 3.48  & 78.6      \\
    \rowcolor{gray94}\bf{MogaNet-XT}        & Ours         & C    & $256^2$ & 3.0    & 1.04  & 77.2      \\
    \rowcolor{gray94}\bf{MogaNet-T}         & Ours         & C    & $224^2$ & 5.2    & 1.10  & 79.0      \\
    \rowcolor{gray94}\bf{MogaNet-T}$^\S$    & Ours         & C    & $256^2$ & 5.2    & 1.44  & \bf{80.0} \\
    \bottomrule
    \end{tabular}
    }
    \vspace{-1.0em}
    \caption{\textbf{IN-1K classification} with lightweight models. \small{$\S$} denotes the refined training scheme.
    % \small{$\dag$} and \small{$\S$} are RSB A2 and refined training schemes.
    }
    \label{tab:in1k_cls_tiny}
    % \vspace{-0.5em}
\end{table}

%% file: Tabs/tab_in1k.tex
\begin{table}[H]
    % \vspace{-0.25em}
    \setlength{\tabcolsep}{0.5mm}
    \centering
\resizebox{\linewidth}{!}{
\begin{tabular}{llccccc}
    \toprule
Architecture                             & Date      & Type & Image   & Param. & FLOPs & Top-1     \\
                                         &           &      & Size    & (M)    & (G)   & Acc (\%)  \\ \hline
Deit-S                                   & ICML'2021 & T    & $224^2$ & 22     & 4.6   & 79.8      \\
Swin-T                                   & ICCV'2021 & T    & $224^2$ & 28     & 4.5   & 81.3      \\
% T2T-ViT$_t$-14                           & ICCV'2021 & T    & $224^2$ & 22     & 6.1   & 81.7      \\
CSWin-T                                  & CVPR'2022 & T    & $224^2$ & 23     & 4.3   & 82.8      \\
LITV2-S                                  & NIPS'2022 & T    & $224^2$ & 28     & 3.7   & 82.0      \\
CoaT-S                                   & ICCV'2021 & H    & $224^2$ & 22     & 12.6  & 82.1      \\
CoAtNet-0                                & NIPS'2021 & H    & $224^2$ & 25     & 4.2   & 82.7      \\
UniFormer-S                              & ICLR'2022 & H    & $224^2$ & 22     & 3.6   & 82.9      \\
RegNetY-4GF$^\dag$                       & CVPR'2020 & C    & $224^2$ & 21     & 4.0   & 81.5      \\
ConvNeXt-T                               & CVPR'2022 & C    & $224^2$ & 29     & 4.5   & 82.1      \\
SLaK-T                                   & ICLR'2023 & C    & $224^2$ & 30     & 5.0   & 82.5      \\
HorNet-T$_{7\times 7}$                   & NIPS'2022 & C    & $224^2$ & 22     & 4.0   & 82.8      \\
\rowcolor{gray94}\bf{MogaNet-S}          & Ours      & C    & $224^2$ & 25     & 5.0   & \bf{83.4} \\ \hline
Swin-S                                   & ICCV'2021 & T    & $224^2$ & 50     & 8.7   & 83.0      \\
Focal-S                                  & NIPS'2021 & T    & $224^2$ & 51     & 9.1   & 83.6      \\
CSWin-S                                  & CVPR'2022 & T    & $224^2$ & 35     & 6.9   & 83.6      \\
LITV2-M                                  & NIPS'2022 & T    & $224^2$ & 49     & 7.5   & 83.3      \\
CoaT-M                                   & ICCV'2021 & H    & $224^2$ & 45     & 9.8   & 83.6      \\
% Twins-SVT-B                              & NIPS'2021 & H    & $224^2$ & 56     & 8.6   & 83.2      \\
CoAtNet-1                                & NIPS'2021 & H    & $224^2$ & 42     & 8.4   & 83.3      \\
UniFormer-B                              & ICLR'2022 & H    & $224^2$ & 50     & 8.3   & 83.9      \\
FAN-B-Hybrid                             & ICML'2022 & H    & $224^2$ & 50     & 11.3  & 83.9      \\
EfficientNet-B6                          & ICML'2019 & C    & $528^2$ & 43     & 19.0  & 84.0      \\
RegNetY-8GF$^\dag$                       & CVPR'2020 & C    & $224^2$ & 39     & 8.1   & 82.2      \\
ConvNeXt-S                               & CVPR'2022 & C    & $224^2$ & 50     & 8.7   & 83.1      \\
FocalNet-S (LRF)                         & NIPS'2022 & C    & $224^2$ & 50     & 8.7   & 83.5      \\
HorNet-S$_{7\times 7}$                   & NIPS'2022 & C    & $224^2$ & 50     & 8.8   & 84.0      \\
SLaK-S                                   & ICLR'2023 & C    & $224^2$ & 55     & 9.8   & 83.8      \\
\rowcolor{gray94}\bf{MogaNet-B}          & Ours      & C    & $224^2$ & 44     & 9.9   & \bf{84.3} \\ \hline
DeiT-B                                   & ICML'2021 & T    & $224^2$ & 86     & 17.5  & 81.8      \\
Swin-B                                   & ICCV'2021 & T    & $224^2$ & 89     & 15.4  & 83.5      \\
Focal-B                                  & NIPS'2021 & T    & $224^2$ & 90     & 16.4  & 84.0      \\
CSWin-B                                  & CVPR'2022 & T    & $224^2$ & 78     & 15.0  & 84.2      \\
DeiT III-B                               & ECCV'2022 & T    & $224^2$ & 87     & 18.0  & 83.8      \\
BoTNet-T7                                & CVPR'2021 & H    & $256^2$ & 79     & 19.3  & 84.2      \\
CoAtNet-2                                & NIPS'2021 & H    & $224^2$ & 75     & 15.7  & 84.1      \\
FAN-B-Hybrid                             & ICML'2022 & H    & $224^2$ & 77     & 16.9  & 84.3      \\
RegNetY-16GF                             & CVPR'2020 & C    & $224^2$ & 84     & 16.0  & 82.9      \\
ConvNeXt-B                               & CVPR'2022 & C    & $224^2$ & 89     & 15.4  & 83.8      \\
RepLKNet-31B                             & CVPR'2022 & C    & $224^2$ & 79     & 15.3  & 83.5      \\
FocalNet-B (LRF)                         & NIPS'2022 & C    & $224^2$ & 89     & 15.4  & 83.9      \\
HorNet-B$_{7\times 7}$                   & NIPS'2022 & C    & $224^2$ & 87     & 15.6  & 84.3      \\
SLaK-B                                   & ICLR'2023 & C    & $224^2$ & 95     & 17.1  & 84.0      \\
\rowcolor{gray94}\bf{MogaNet-L}          & Ours      & C    & $224^2$ & 83     & 15.9  & \bf{84.7} \\ \hline
Swin-L$^\ddag$                           & ICCV'2021 & T    & $384^2$ & 197    & 104   & 87.3      \\
DeiT III-L$^\ddag$                       & ECCV'2022 & T    & $384^2$ & 304    & 191   & 87.7      \\
CoAtNet-3$^\ddag$                        & NIPS'2021 & H    & $384^2$ & 168    & 107   & 87.6      \\
RepLKNet-31L$^\ddag$                     & CVPR'2022 & C    & $384^2$ & 172    & 96    & 86.6      \\
ConvNeXt-L                               & CVPR'2022 & C    & $224^2$ & 198    & 34.4  & 84.3      \\
ConvNeXt-L$^\ddag$                       & CVPR'2022 & C    & $384^2$ & 198    & 101   & 87.5      \\
ConvNeXt-XL$^\ddag$                      & CVPR'2022 & C    & $384^2$ & 350    & 179   & 87.8      \\
HorNet-L$^\ddag$                         & NIPS'2022 & C    & $384^2$ & 202    & 102   & 87.7      \\
\rowcolor{gray94}\bf{MogaNet-XL}         & Ours      & C    & $224^2$ & 181    & 34.5  & 85.1      \\
\rowcolor{gray94}\bf{MogaNet-XL}$^\ddag$ & Ours      & C    & $384^2$ & 181    & 102   & \bf{87.8} \\
    \bottomrule
    \end{tabular}
    }
    \vspace{-1.0em}
    \caption{\textbf{IN-1K classification} performance with scaling-up models. $^\ddag$ denotes the model is pre-trained on IN-21K and fine-tuned on IN-1K.}
    % \vspace{-2.0em}
    \label{tab:in1k_cls_scaling}
\end{table}

%% file: Tabs/tab_coco.tex
\begin{table}[H]
    % \vspace{-0.25em}
    \setlength{\tabcolsep}{0.4mm}
    \centering
\resizebox{\linewidth}{!}{
\begin{tabular}{lllcccc}
    \toprule
Architecture                             & Data      & Method       & Param. & FLOPs & AP$^{b}$  & AP$^{m}$  \\
                                         &           &              & (M)    & (G)   & (\%)      & (\%)      \\ \hline
ResNet-101                               & CVPR'2016 & RetinaNet    & 57     & 315   & 38.5      & -         \\
PVT-S                                    & ICCV'2021 & RetinaNet    & 34     & 226   & 40.4      & -         \\
CMT-S                                    & CVPR'2022 & RetinaNet    & 45     & 231   & 44.3      & -         \\
\rowcolor{gray94}\bf{MogaNet-S}          & Ours      & RetinaNet    & 35     & 253   & \bf{45.8} & -         \\ \hline
RegNet-1.6G                              & CVPR'2020 & Mask R-CNN   & 29     & 204   & 38.9      & 35.7      \\
PVT-T                                    & ICCV'2021 & Mask R-CNN   & 33     & 208   & 36.7      & 35.1      \\
\rowcolor{gray94}\bf{MogaNet-T}          & Ours      & Mask R-CNN   & 25     & 192   & \bf{42.6} & \bf{39.1} \\ \hline
Swin-T                                   & ICCV'2021 & Mask R-CNN   & 48     & 264   & 42.2      & 39.1      \\
Uniformer-S                              & ICLR'2022 & Mask R-CNN   & 41     & 269   & 45.6      & 41.6      \\
ConvNeXt-T                               & CVPR'2022 & Mask R-CNN   & 48     & 262   & 44.2      & 40.1      \\
PVTV2-B2                                 & CVMJ'2022 & Mask R-CNN   & 45     & 309   & 45.3      & 41.2      \\
LITV2-S                                  & NIPS'2022 & Mask R-CNN   & 47     & 261   & 44.9      & 40.8      \\
FocalNet-T                               & NIPS'2022 & Mask R-CNN   & 49     & 267   & 45.9      & 41.3      \\
\rowcolor{gray94}\bf{MogaNet-S}          & Ours      & Mask R-CNN   & 45     & 272   & \bf{46.7} & \bf{42.2} \\ \hline
Swin-S                                   & ICCV'2021 & Mask R-CNN   & 69     & 354   & 44.8      & 40.9      \\
Focal-S                                  & NIPS'2021 & Mask R-CNN   & 71     & 401   & 47.4      & 42.8      \\
ConvNeXt-S                               & CVPR'2022 & Mask R-CNN   & 70     & 348   & 45.4      & 41.8      \\
HorNet-B$_{7\times 7}$                   & NIPS'2022 & Mask R-CNN   & 68     & 322   & 47.4      & 42.3      \\
\rowcolor{gray94}\bf{MogaNet-B}          & Ours      & Mask R-CNN   & 63     & 373   & \bf{47.9} & \bf{43.2} \\ \hline
Swin-L$^\ddag$                           & ICCV'2021 & Cascade Mask & 253    & 1382  & 53.9      & 46.7      \\
ConvNeXt-L$^\ddag$                       & CVPR'2022 & Cascade Mask & 255    & 1354  & 54.8      & 47.6      \\
RepLKNet-31L$^\ddag$                     & CVPR'2022 & Cascade Mask & 229    & 1321  & 53.9      & 46.5      \\
HorNet-L$^\ddag$                         & NIPS'2022 & Cascade Mask & 259    & 1399  & 56.0      & 48.6      \\
\rowcolor{gray94}\bf{MogaNet-XL}$^\ddag$ & Ours      & Cascade Mask & 238    & 1355  & \bf{56.2} & \bf{48.8} \\
    \bottomrule
    \end{tabular}
    }
    \vspace{-1.0em}
    \caption{\textbf{COCO object detection and instance segmentation} with RetinaNet ($1\times$), Mask R-CNN ($1\times$), and Cascade Mask R-CNN (multi-scale $3\times$). $^\ddag$ indicates IN-21K pre-trained models. The FLOPs are measured at $800\times 1280$.}
    % \vspace{-1.0em}
    \label{tab:coco}
\end{table}

%% file: Tabs/tab_ade20k.tex
\begin{table}[H]
    % \vspace{-0.25em}
    \setlength{\tabcolsep}{0.9mm}
    \centering
\resizebox{\linewidth}{!}{
\begin{tabular}{c|llcccc}
    \toprule
Method           & Architecture                                 & Date                   & Crop                      & Param.                & FLOPs                  & mIoU$^{ss}$                 \\
                 &                                              &                        & size                      & (M)                   & (G)                    & (\%)                        \\ \hline
                 % & ResNet50                                     & CVPR'2016              & 512$^2$                   & 29                    & 183                    & 36.7                        \\
                 & PVT-S                                        & ICCV'2021              & 512$^2$                   & 28                    & 161                    & 39.8                        \\
\small{Semantic} & Twins-S                                      & NIPS'2021              & 512$^2$                   & 28                    & 162                    & 44.3                        \\
FPN              & Swin-T                                       & ICCV'2021              & 512$^2$                   & 32                    & 182                    & 41.5                        \\
(80K)            & Uniformer-S                                  & ICLR'2022              & 512$^2$                   & 25                    & 247                    & 46.6                        \\
                 & LITV2-S                                      & NIPS'2022              & 512$^2$                   & 31                    & 179                    & 44.3                        \\
                 & VAN-B2                                       & CVMJ'2023              & 512$^2$                   & 30                    & 164                    & 46.7                        \\
                 & \cellcolor{gray94}\bf{MogaNet-S}             & \cellcolor{gray94}Ours & \cellcolor{gray94}512$^2$ & \cellcolor{gray94}29  & \cellcolor{gray94}189  & \cellcolor{gray94}\bf{47.7} \\ \hline
                 & DeiT-S                                       & ICML'2021              & 512$^2$                   & 52                    & 1099                   & 44.0                        \\
                 & Swin-T                                       & ICCV'2021              & 512$^2$                   & 60                    & 945                    & 46.1                        \\
                 & ConvNeXt-T                                   & CVPR'2022              & 512$^2$                   & 60                    & 939                    & 46.7                        \\
                 % & Twins-S                                      & NIPS'2021              & 512$^2$                   & 54                    & 901                    & 46.2                        \\
                 & UniFormer-S                                  & ICLR'2022              & 512$^2$                   & 52                    & 1008                   & 47.6                        \\
                 & HorNet-T$_{7\times 7}$                       & NIPS'2022              & 512$^2$                   & 52                    & 926                    & 48.1                        \\
                 & \cellcolor{gray94}\bf{MogaNet-S}             & \cellcolor{gray94}Ours & \cellcolor{gray94}512$^2$ & \cellcolor{gray94}55  & \cellcolor{gray94}946  & \cellcolor{gray94}\bf{49.2} \\ \cline{2-7} 
                 & Swin-S                                       & ICCV'2021              & 512$^2$                   & 81                    & 1038                   & 48.1                        \\
                 & ConvNeXt-S                                   & CVPR'2022              & 512$^2$                   & 82                    & 1027                   & 48.7                        \\
UperNet          & SLaK-S                                       & ICLR'2023              & 512$^2$                   & 91                    & 1028                   & 49.4                        \\
(160K)           & \cellcolor{gray94}\bf{MogaNet-B}             & \cellcolor{gray94}Ours & \cellcolor{gray94}512$^2$ & \cellcolor{gray94}74  & \cellcolor{gray94}1050 & \cellcolor{gray94}\bf{50.1} \\ \cline{2-7} 
                 & Swin-B                                       & ICCV'2021              & 512$^2$                   & 121                   & 1188                   & 49.7                        \\
                 & ConvNeXt-B                                   & CVPR'2022              & 512$^2$                   & 122                   & 1170                   & 49.1                        \\
                 & RepLKNet-31B                                 & CVPR'2022              & 512$^2$                   & 112                   & 1170                   & 49.9                        \\
                 & SLaK-B                                       & ICLR'2023              & 512$^2$                   & 135                   & 1185                   & 50.2                        \\
                 & \cellcolor{gray94}\bf{MogaNet-L}             & \cellcolor{gray94}Ours & \cellcolor{gray94}512$^2$ & \cellcolor{gray94}113 & \cellcolor{gray94}1176 & \cellcolor{gray94}\bf{50.9} \\ \cline{2-7} 
                 & Swin-L$^\ddag$                               & ICCV'2021              & 640$^2$                   & 234                   & 2468                   & 52.1                        \\
                 & ConvNeXt-L$^\ddag$                           & CVPR'2022              & 640$^2$                   & 245                   & 2458                   & 53.7                        \\
                 & RepLKNet-31L$^\ddag$                         & CVPR'2022              & 640$^2$                   & 207                   & 2404                   & 52.4                        \\
                 & \cellcolor{gray94}\bf{MogaNet-XL}$^\ddag$    & \cellcolor{gray94}Ours & \cellcolor{gray94}640$^2$ & \cellcolor{gray94}214 & \cellcolor{gray94}2451 & \cellcolor{gray94}\bf{54.0} \\
    \bottomrule
    \end{tabular}
    }
    \vspace{-1.0em}
    \caption{\textbf{ADE20K semantic segmentation} with semantic FPN (80K) and UperNet (160K). $^\ddag$ indicates using IN-21K pre-trained models. The FLOPs are measured at $512\times 2048$ or $640\times 2560$.}
    % \vspace{-1.0em}
    \label{tab:ade20k}
\end{table}

% Semantic FPN
% ResNet-50 80k
% PVT-S 40k
% PVT.V2-S 40k
% Swin-T 80k
% Twins-S 80k
% Poolformer-M36 40k
% Uniformer-S 80k
% LIT.V2 80k
% VAN-B2 40k
% MogaNet-S 80k

%% file: Tabs/tab_coco_pose.tex
\begin{table}[H]
    % \vspace{-0.25em}
    \setlength{\tabcolsep}{0.5mm}
    \centering
\resizebox{\linewidth}{!}{
\begin{tabular}{llccccccc}
    \toprule
Architecture                    & Date      & Crop            & Param. & FLOPs & AP        & AP$^{50}$ & AP$^{75}$ & AR        \\
                                &           & size            & (M)    & (G)   & (\%)      & (\%)      & (\%)      & (\%)      \\ \hline
RSN-18                          & ECCV'2020 & $256\times 192$ & 9.1    & 2.3   & 70.4      & 88.7      & 77.9      & 77.1      \\
\rowcolor{gray94}\bf{MogaNet-T} & Ours      & $256\times 192$ & 8.1    & 2.2   & \bf{73.2} & \bf{90.1} & \bf{81.0} & \bf{78.8} \\ \hline
HRNet-W32                       & CVPR'2019 & $256\times 192$ & 28.5   & 7.1   & 74.4      & 90.5      & 81.9      & 78.9      \\
Swin-T                          & ICCV'2021 & $256\times 192$ & 32.8   & 6.1   & 72.4      & 90.1      & 80.6      & 78.2      \\
PVTV2-B2                        & CVML'2022 & $256\times 192$ & 29.1   & 4.3   & 73.7      & 90.5      & 81.2      & 79.1      \\
Uniformer-S                     & ICLR'2022 & $256\times 192$ & 25.2   & 4.7   & 74.0      & 90.3      & 82.2      & 79.5      \\
ConvNeXt-T                      & CVPR'2022 & $256\times 192$ & 33.1   & 5.5   & 73.2      & 90.0      & 80.9      & 78.8      \\
\rowcolor{gray94}\bf{MogaNet-S} & Ours      & $256\times 192$ & 29.0   & 6.0   & \bf{74.9} & \bf{90.7} & \bf{82.8} & \bf{80.1} \\ \hline
Uniformer-S                     & ICLR'2022 & $384\times 288$ & 25.2   & 11.1  & 75.9      & 90.6      & 83.4      & 81.4      \\
ConvNeXt-T                      & CVPR'2022 & $384\times 288$ & 33.1   & 33.1  & 75.3      & 90.4      & 82.1      & 80.5      \\
\rowcolor{gray94}\bf{MogaNet-S} & Ours      & $384\times 288$ & 29.0   & 13.5  & \bf{76.4} & \bf{91.0} & \bf{83.3} & \bf{81.4} \\ \hline
HRNet-W48                       & CVPR'2019 & $384\times 288$ & 63.6   & 32.9  & 76.3      & 90.8      & 82.0      & 81.2      \\
Swin-L                          & ICCV'2021 & $384\times 288$ & 203.4  & 86.9  & 76.3      & 91.2      & 83.0      & 814       \\
Uniformer-B                     & ICLR'2022 & $384\times 288$ & 53.5   & 14.8  & 76.7      & 90.8      & 84.0      & 81.4      \\
\rowcolor{gray94}\bf{MogaNet-B} & Ours      & $384\times 288$ & 47.4   & 24.4  & \bf{77.3} & \bf{91.4} & \bf{84.0} & \bf{82.2} \\
    \bottomrule
    \end{tabular}
    }
    \vspace{-1.0em}
    \caption{\textbf{COCO 2D human pose estimation} with Top-Down SimpleBaseline. The FLOPs are measured at $256\times 192$ or $384\times 288$.}
    \label{tab:coco_pose}
    % \vspace{-0.5em}
\end{table}

%% file: Tabs/tab_3d_vp.tex
% \begin{table}[H]
\begin{wraptable}{r}{0.525\textwidth}
    \vspace{-1.25em}
    \setlength{\tabcolsep}{0.3mm}
    \centering
\resizebox{\linewidth}{!}{
\begin{tabular}{l|ccc|ccc|cccc}
    \toprule
    Architecture                       & \multicolumn{3}{c|}{3D Face} & \multicolumn{3}{c|}{3D Hand}    & \multicolumn{4}{c}{Video Prediction}         \\
                                       & \#P.  & FLOPs & 3DRMSE       & \#P. & FLOPs & PA-MPJPE         & \#P. & FLOPs & MSE          & SSIM           \\
                                       & (M)   & (G)   & $\downarrow$ & (M)  & (G)   & (mm)$\downarrow$ & (M)  & (G)   & $\downarrow$ & (\%)$\uparrow$ \\ \hline
    DeiT-S                             & 25    & 6.6   & 2.52         & 25   & 4.8   & 7.86             & 46   & 16.9  & 35.2         & 91.4           \\
    Swin-T                             & 30    & 6.1   & 2.45         & 30   & 4.6   & 6.97             & 46   & 16.4  & 29.7         & 93.3           \\
    ConvNeXt-T                         & 30    & 5.8   & 2.34         & 30   & 4.5   & 6.46             & 37   & 14.1  & 26.9         & 94.0           \\
    HorNet-T                           & 25    & 5.6   & 2.39         & 25   & 4.3   & 6.23             & 46   & 16.3  & 29.6         & 93.3           \\
    \rowcolor{gray94}MogaNet-S         & 27    & 6.5   & \bf{2.24}    & 27   & 5.0   & \bf{6.08}        & 47   & 16.5  & \bf{25.6}    & \bf{94.3}      \\
    \bottomrule
    \end{tabular}
    }
    \vspace{-0.5em}
    \caption{\textbf{3D human pose estimation} and \textbf{video prediction} with ExPose and SimVP on Stirling/ESRC 3D, FreiHAND, and MMNIST datasets. FLOPs of the face and hand tasks are measured at $3\times 256^2$ and $3\times 224^2$ while using 10 frames at $1\times 64^2$ resolutions for video prediction.}
    \vspace{-1.25em}
    \label{tab:3d_vp}
\end{wraptable}
% \end{table}

%% file: Tabs/tab_ablation_small.tex
\begin{table}[H]
    \vspace{-0.5em}
    \setlength{\tabcolsep}{0.7mm}
    \centering
\resizebox{\linewidth}{!}{
    \begin{tabular}{l|c}
    \toprule
Modules                       & Top-1     \\
                              & Acc (\%)  \\ \hline
ConvNeXt-T                    & 82.1      \\
Baseline                      & 82.2      \\ \hline
\rowcolor{gray94}Moga Block   & \bf{83.4} \\
$- \mathrm{FD}(\cdot)$        & 83.2      \\
$-$Multi-$\mathrm{DW}(\cdot)$ & 83.1      \\
$- \mathrm{Moga}(\cdot)$      & 82.7      \\
$- \mathrm{CA}(\cdot)$        & 82.9      \\
    \bottomrule
    \end{tabular}
    }
    \vspace{-0.5em}
    % \caption{\textbf{Ablation of the designed modules on ImageNet-1K}.
    % }
    \label{tab:ablation_small}
    \vspace{-1.0em}
\end{table}

%% file: 5_conclusion.tex
% for ICLR submit
\if\submission\submissionFinal
    \vspace{-0.5em}
    \section{Conclusion}
    \label{sec:conclusion}
    \vspace{-0.25em}
    This paper introduces a new modern ConvNet architecture, named MogaNet, through the lens of multi-order game-theoretic interaction.
    Built upon the modern ConvNet framework, we present a compact Moga Block and channel aggregation module to force the network to emphasize the expressive but inherently overlooked interactions across spatial and channel perspectives.
    Extensive experiments verify the consistent superiority of MogaNet in terms of both performance and efficiency compared to popular ConvNets, ViTs, and hybrid architectures on various vision benchmarks.
\else
% for arXiv
% \if\submission\submissionarXiv
    \section{Conclusion}
    \label{sec:conclusion}
    In this paper, we introduce a new modern ConvNet architecture named MogaNet through the lens of multi-order game-theoretic interaction.
    % We demonstrate that using a spatial aggregation block and a channel aggregation block results in stronger feature interactions of intermediate complexities efficiently, boosting the performance of ConvNet architecture substantially on diverse vision scenarios.
    Built upon the modern ConvNet framework, we present a compact Moga Block and channel aggregation module to \pl{force the network to emphasize the expressive but inherently overlooked interactions} across spatial and channel spaces.
    Extensive experiments demonstrate the consistent superiority of MogaNet in terms of both accuracy and computational efficiency compared to popular ConvNets, ViTs, and hybrid architectures on various vision benchmarks.
    We hope our work can prompt people to perceive the importance of multi-order interaction in representation learning and to facilitate the development of efficient deep architecture design.
\fi

\section*{Acknowledgement}
This work was supported by the National Key R\&D Program of China (No. 2022ZD0115100), the National Natural Science Foundation of China Project (No. U21A20427), and Project (No. WU2022A009) from the Center of Synthetic Biology and Integrated Bioengineering of Westlake University.
This work was done when Zedong Wang and Zhiyuan Chen interned at Westlake University. We thank the AI Station of Westlake University for the support of GPUs. We also thank Mengzhao Chen, Zhangyang Gao, Jianzhu Guo, Fang Wu, and all anonymous reviewers for polishing the writing of the manuscript.

%% file: 6_appendix.tex
\renewcommand\thefigure{A\arabic{figure}}
\renewcommand\thetable{A\arabic{table}}
\setcounter{table}{0}
\setcounter{figure}{0}

\newpage
\appendix

\section{Implementation Details}
\label{app:implement}
\subsection{Architecture Details}
\label{app:architecture}
% fig: framework
\begin{wrapfigure}{r}{0.5\linewidth}
    \vspace{-4.5em}
    \begin{center}
    \includegraphics[width=1.0\linewidth]{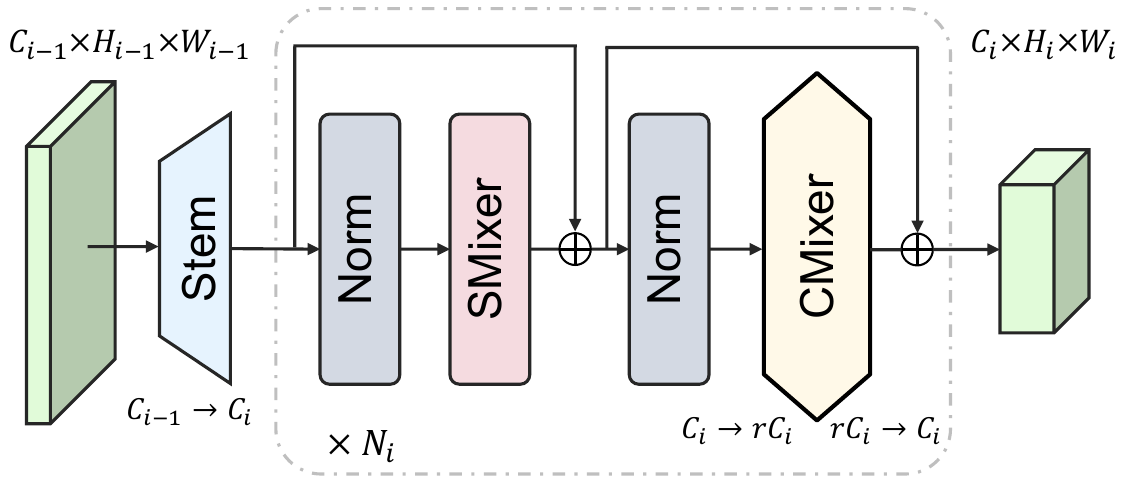}
    \end{center}
    \vspace{-1.0em}
    \caption{
\textbf{Modern ConvNet architecture.} It has 4 stages in hierarchical, and $i$-th stage contains an embedding stem and $N_{i}$ blocks of $\mathrm{SMixer}(\cdot)$ and $\mathrm{CMixer}(\cdot)$ with PreNorm~\citep{acl2019Learning} and identical connection~\citep{he2016deep}. The features within the $i$-th stage are in the same shape, except that $\mathrm{CMixer}(\cdot)$ will increase the dimension to $rC_{i}$ with an expand ratio $r$ as an inverted bottleneck \citep{cvpr2018mobilenetv2}.
    }
    \label{fig:framework}
    \vspace{-1.0em}
\end{wrapfigure}
The detailed architecture specifications of MogaNet are shown in Table~\ref{tab:app_architecture} and Fig.~\ref{fig:moga_framework}, where an input image of $224^2$ resolutions is assumed for all architectures. We rescale the groups of embedding dimensions the number of Moga Blocks for each stage corresponding to different models of varying magnitudes:
\romannumeral1) MogaNet-X-Tiny and MogaNet-Tiny with embedding dimensions of $\{32, 64, 96, 192\}$ and $\{32, 64, 128, 256\}$ exhibit competitive parameter numbers and computational overload as recently proposed light-weight architectures~\citep{iclr2022mobilevit, cvpr2022MobileFormer, eccv2022edgeformer};
\romannumeral2) MogaNet-Small adopts embedding dimensions of $\{64, 128, 320, 512\}$ in comparison to other prevailing small-scale architectures~\citep{liu2021swin, cvpr2022convnext};
\romannumeral3) MogaNet-Base with embedding dimensions of $\{64, 160, 320, 512\}$ in comparison to medium size architectures;
\romannumeral4) MogaNet-Large with embedding dimensions of $\{64, 160, 320, 640\}$ is designed for large-scale computer vision tasks.
\romannumeral5) MogaNet-X-Large with embedding dimensions of $\{96, 192, 480, 960\}$ is a scaling-up version (around 200M parameters) for large-scale tasks.
The FLOPs are measured for image classification on ImageNet~\citep{cvpr2009imagenet} at resolution $224^2$, where a global average pooling (GAP) layer is applied to the output feature map of the last stage, followed by a linear classifier.

\input{Tabs/tab_architecture.tex}

\begin{figure*}[t]
\vspace{-1.5em}
\begin{minipage}{0.543\linewidth}
\centering
    \input{Tabs/tab_train_conf.tex}
\end{minipage}
\begin{minipage}{0.462\linewidth}
\centering
    \input{Tabs/tab_train_conf_in21k}
\end{minipage}
\end{figure*}

\subsection{Experimental Settings for ImageNet}
\label{app:in1k_settings}
We conduct image classification experiments on ImageNet~\citep{cvpr2009imagenet} datasets. All experiments are implemented on \texttt{OpenMixup}~\citep{li2022openmixup} and \texttt{timm}~\citep{wightman2021rsb} codebases running on 8 NVIDIA A100 GPUs. View more results in Appendix \ref{app:exp_in1k}.

\paragraph{ImageNet-1K.}
We perform regular ImageNet-1K training mostly following the training settings of DeiT~\citep{icml2021deit} and RSB A2~\citep{wightman2021rsb} in Table~\ref{tab:in1k_config}, which are widely adopted for Transformer and ConvNet architectures. For all models, the default input image resolution is $224^2$ for training from scratch. We adopt $256^2$ resolutions for lightweight experiments according to MobileViT~\citep{iclr2022mobilevit}. Taking training settings for the model with 25M or more parameters as the default, we train all MogaNet models for 300 epochs by AdamW \citep{iclr2019AdamW} optimizer using a batch size of 1024, a basic learning rate of $1\times 10^{-3}$, a weight decay of 0.05, and a Cosine learning rate scheduler \citep{loshchilov2016sgdr} with 5 epochs of linear warmup~\citep{devlin2018bert}.
As for augmentation and regularization techniques, we adopt most of the data augmentation and regularization strategies applied in DeiT training settings, including Random Resized Crop (RRC) and Horizontal flip \citep{szegedy2015going}, RandAugment \citep{cubuk2020randaugment}, Mixup~\citep{zhang2017mixup}, CutMix~\citep{yun2019cutmix}, random erasing~\citep{zhong2020random}, ColorJitter \citep{he2016deep}, stochastic depth~\citep{eccv2016droppath}, and label smoothing \citep{cvpr2016inceptionv3}. Similar to ConvNeXt~\citep{cvpr2022convnext}, we do not apply Repeated augmentation \citep{cvpr2020repeat} and gradient clipping, which are designed for Transformers but do not enhance the performances of ConvNets while using Exponential Moving Average (EMA)~\citep{siam1992ema} with the decay rate of 0.9999 by default. We also remove additional augmentation strategies~\citep{cvpr2019AutoAugment, eccv2022AutoMix, Li2021SAMix, 2022decouplemix}, \textit{e.g.,} PCA lighting \citep{Krizhevsky2012ImageNetCW} and AutoAugment~\citep{cvpr2019AutoAugment}.
Since lightweight architectures (3$\sim$10M parameters) tend to get under-fitted with strong augmentations and regularization, we adjust the training configurations for MogaNet-XT/T following \citep{iclr2022mobilevit, cvpr2022MobileFormer, eccv2022edgeformer}, including employing the weight decay of 0.03 and 0.04, Mixup with $\alpha$ of 0.1, and RandAugment of $7/0.5$ for MogaNet-XT/T. Since EMA is proposed to stabilize the training process of large models, we also remove it for MogaNet-XT/T as a fair comparison. An increasing degree of stochastic depth path augmentation is employed for larger models. In evaluation, the top-1 accuracy using a single crop with a test crop ratio of 0.9 is reported as \citep{iccv2021t2t, yu2022metaformer, guo2022van}.

% \vspace{-0.5em}
\paragraph{ImageNet-21K.}
Following ConvNeXt, we further provide the training recipe for ImageNet-21K~\citep{cvpr2009imagenet} pre-training and ImageNet-1K fine-tuning with high resolutions in Table~\ref{tab:in21k_config}. EMA is removed in pre-training, while CutMix and Mixup are removed for fine-tuning.

\subsection{Object Detection and Segmentation on COCO}
\label{app:coco_det_settings}
Following Swin~\citep{liu2021swin} and PoolFormer~\citep{yu2022metaformer}, we evaluate objection detection and instance segmentation tasks on COCO~\citep{2014MicrosoftCOCO} benchmark, which include 118K training images (\textit{train2017}) and 5K validation images (\textit{val2017}). We adopt RetinaNet~\citep{iccv2017retinanet}, Mask R-CNN~\citep{2017iccvmaskrcnn}, and Cascade Mask R-CNN~\citep{tpami2019cascade} as the standard detectors and use ImageNet-1K pre-trained weights as the initialization of the backbones. As for RetinaNet and Mask R-CNN, we employ AdamW~\citep{iclr2019AdamW} optimizer for training $1\times$ scheduler (12 epochs) with a basic learning rate of $1\times 10^{-4}$ and a batch size of 16. As for Cascade Mask R-CNN, the $3\times$ training scheduler and multi-scale training resolutions (MS) are adopted. The pre-trained weights on ImageNet-1K and ImageNet-21K are used accordingly to initialize backbones. The shorter side of training images is resized to 800 pixels, and the longer side is resized to not more than 1333 pixels. We calculate the FLOPs of compared models at $800\times 1280$ resolutions. Experiments of COCO detection are implemented on \texttt{MMDetection}~\citep{mmdetection} codebase and run on 8 NVIDIA A100 GPUs. View detailed results in Appendix \ref{app:exp_det_coco}.

\subsection{Semantic Segmentation on ADE20K}
\label{app:ade20k_seg_settings}
We evaluate semantic segmentation on ADE20K~\citep{Zhou2018ADE20k} benchmark, which contains 20K training images and 2K validation images, covering 150 fine-grained semantic categories. 
We first adopt Semantic FPN~\citep{cvpr2019semanticFPN} following PoolFormer~\citep{yu2022metaformer} and Uniformer~\citep{iclr2022uniformer}, which train models for 80K iterations by AdamW~\citep{iclr2019AdamW} optimizer with a basic learning rate of $2\times 10^{-4}$, a batch size of 16, and a poly learning rate scheduler. Then, we utilize UperNet~\citep{eccv2018upernet} following Swin~\citep{liu2021swin}, which employs AdamW optimizer using a basic learning rate of $6\times 10^{-5}$, a weight decay of 0.01, a poly scheduler with a linear warmup of 1,500 iterations. We use ImageNet-1K and ImageNet-21K pre-trained weights to initialize the backbones accordingly. The training images are resized to $512^2$ resolutions, and the shorter side of testing images is resized to 512 pixels. We calculate the FLOPs of models at $800\times 2048$ resolutions. Experiments of ADE20K segmentation are implemented on \texttt{MMSegmentation}~\citep{mmseg2020} codebase and run on 8 NVIDIA A100 GPUs. View full comparison results in Appendix \ref{app:exp_seg_ade20k}.

\subsection{2D Human Pose Estimation on COCO}
\label{app:coco_pose_settings}
We evaluate 2D human keypoints estimation tasks on COCO~\citep{2014MicrosoftCOCO} benchmark based on Top-Down SimpleBaseline~\citep{eccv2018simple} (adding a Top-Down estimation head after the backbone) following PVT~\citep{iccv2021PVT} and UniFormer~\citep{iclr2022uniformer}. 
We fine-tune all models for 210 epochs with Adam optimizer \citep{iclr2014adam} using a basic learning rate selected in \{$1\times 10^{-3}, 5\times 10^{-4}$\}, a multi-step learning rate scheduler decay at 170 and 200 epochs. ImageNet-1K pre-trained weights are used as the initialization of the backbones. The training and testing images are resized to $256\times 192$ or $384\times 288$ resolutions, and the FLOPs of models are calculated at both resolutions. COCO pose estimation experiments are implemented on \texttt{MMPose}~\citep{mmpose2020} codebase and run on 8 NVIDIA A100 GPUs. View full experiment results in Appendix \ref{app:exp_2d_pose}.

\subsection{3D Human Pose Estimation}
\label{app:coco_3d_settings}
We evaluate MogaNet and popular architectures with 3D human pose estimation tasks with a single monocular image based on ExPose~\citep{eccv2020ExPose}. We first benchmark widely-used ConvNets with the 3D face mesh surface estimation task based on ExPose. All models are trained for 100 epochs on Flickr-Faces-HQ Dataset (FFHQ)~\citep{cvpr2019ffhq} and tested on Stirling/ESRC 3D dataset~\citep{feng2018evaluation}, which consists of facial RGB images with ground-truth 3D face scans. 3D Root Mean Square Error (3DRMSE) measures errors between the predicted and ground-truth face scans. Following ExPose, the Adam optimizer is employed with a batch size of 256, a basic learning rate selected in \{$2\times 10^{-4}, 1\times 10^{-4}$\}, a multi-step learning rate scheduler decay at 60 and 100 epochs. ImageNet-1K pre-trained weights are adopted as the backbone initialization. The training and testing images are resized to $256\times 256$ resolutions.
Then, we evaluate ConvNets with the hand 3D pose estimation tasks. FreiHAND dataset \citep{iccv2019freihand}, which contains multi-view RGB hand images, 3D MANO hand pose, and shape annotations, is adopted for training and testing. Mean Per-Joint Position Error (PA-MPJPE) is used to evaluate 3D skeletons. Notice that a ``PA'' prefix denotes that the metric measures error after solving rotation, scaling, and translation transforms using Procrustes Alignment. Refer to ExPose for more implementation details. All models use the same training settings as the 3D face task, and the training and testing resolutions are $224\times 224$. Experiments of 3D pose estimation are implemented on \texttt{MMHuman3D}~\citep{mmhuman3d} codebase and run on 4 NVIDIA A100 GPUs. View full results in Appendix \ref{app:exp_3d_pose}.

\subsection{Video Prediction on Moving MNIST}
\label{app:mmnist_vp_settings}
We evaluate various Metaformer architectures~\citep{yu2022metaformer} and MogaNet with video prediction tasks on Moving MNIST (MMNIST)~\citep{2014MicrosoftCOCO} based on SimVP~\citep{cvpr2022simvp}. Notice that the hidden translator of SimVP is a 2D network module to learn spatio-temporal representation, which any 2D architecture can replace. Therefore, we can benchmark various architectures based on the SimVP framework. In MMNIST~\citep{icml2015mmnist}, each video is randomly generated with 20 frames containing two digits in $64\times 64$ resolutions, and the model takes 10 frames as the input to predict the next 10 frames. Video predictions are evaluated by Mean Square Error (MSE), Mean Absolute Error (MAE), and Structural Similarity Index (SSIM). All models are trained on MMNIST from scratch for 200 or 2000 epochs with Adam optimizer, a batch size of 16, a OneCycle learning rate scheduler, an initial learning rate selected in \{$1\times 10^{-2}, 5\times 10^{-3}, 1\times 10^{-3}, 5\times 10^{-4}$\}. Experiments of video prediction are implemented on \texttt{OpenSTL}{\footnote{\url{https://github.com/chengtan9907/OpenSTL}}} codebase~\citep{tan2023openstl} and run on a single NVIDIA Tesla V100 GPU. View full benchmark results in Appendix \ref{app:exp_vp_mmnist}.

\section{Empirical Experiment Results}
\label{app:empirical}
\subsection{Representation Bottleneck of DNNs from the View of Multi-order Interaction}
\label{app:interaction}
\paragraph{Multi-order game-theoretic interaction.}
In Sec.~\ref{sec:rep_bottleneck}, we interpret the learned representation of DNNs through the lens of multi-order game-theoretic interaction~\citep{zhang2020interpreting, deng2021discovering}, which disentangles inter-variable communication effects in a DNN into diverse game-theoretic
components of different interaction orders. The order here denotes the \textit{scale of context} involved in the whole computation process of game-theoretic interaction. 

For computer vision, the $m$-th order interaction $I^{(m)}(i,j)$ measures the average game-theoretic interaction effects between image patches $i$ and $j$ on all $m$ image patch contexts. 
Take face recognition as an example, we can consider patches $i$ and $j$ as \textit{two eyes} on this face. Besides, we regard other $m$ visible image patches included on the face. The interaction effect and contribution between the eye's patches $i$ and $j$ toward the task depend on such $m$ visible patches as the context, which is measured as the aforementioned $I^{(m)}(i,j)$.
If $I^{(m)}(i,j) > 0$ , patches $i$ and $j$ show a positive effect under $m$ context. Accordingly, if $I^{(m)}(i,j) < 0$, we consider$i$ and $j$ have a negative effect under $m$ context.
More importantly, interactions of low-order mainly reflect \textbf{widely-shared local texture} and \textbf{common} visual concepts. The middle-order interactions are primarily responsible for encoding \textbf{discriminative high-level} representations. 
However, the high-order ones are inclined to let DNNs memorize the pattern of \textbf{rare outliers} and large-scale shape with \textbf{intensive global interactions}, which can presumably over-fit our deep models~\citep{deng2021discovering, cheng2021game}. Consequently, the occurrence of \textit{excessively low- or high-order} game-theoretic interaction in a deep architecture may therefore be undesirable.

\begin{figure*}[t]
    \vspace{-2.0em}
    \centering
    \subfloat[]{\label{fig:interaction_cnns}
    \hspace{-0.75em}
    \includegraphics[width=0.335\linewidth,trim= 0 10 0 0,clip]{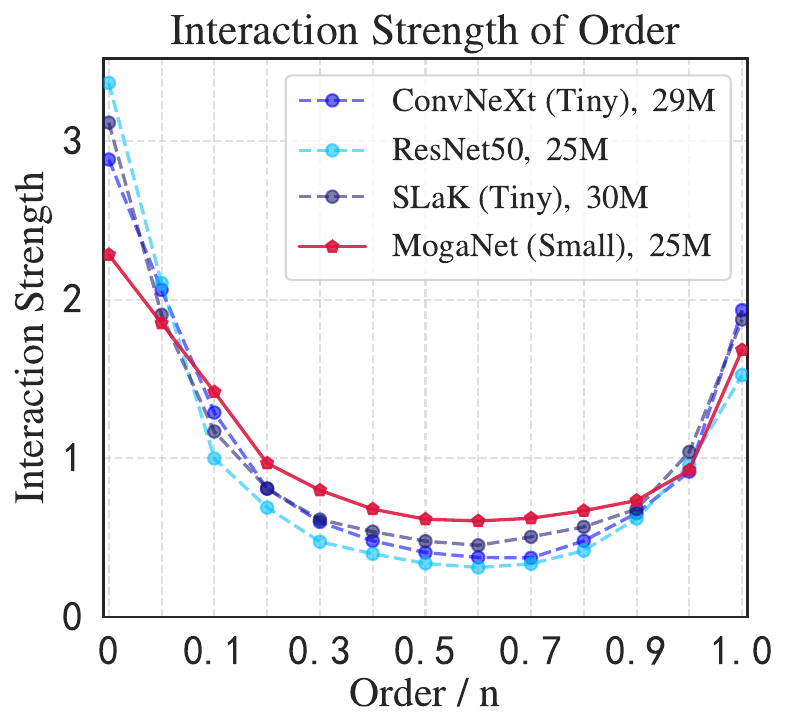}
    }
    \subfloat[]{\label{fig:interaction_cnns_gating}
    \hspace{-0.75em}
    \includegraphics[width=0.335\linewidth,trim= 0 10 0 0,clip]{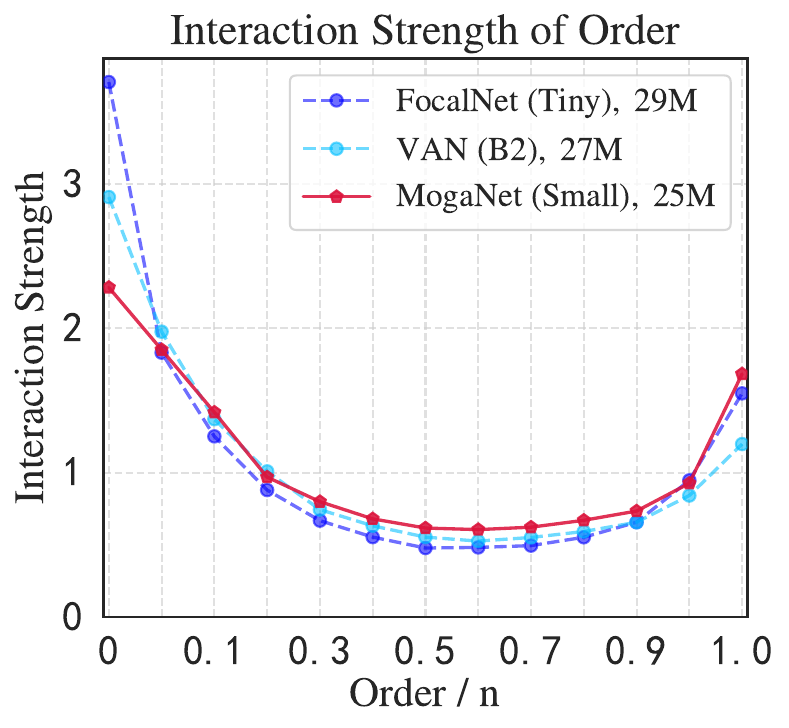}
    }
    \subfloat[]{\label{fig:interaction_vits}
    \hspace{-0.75em}
    \includegraphics[width=0.335\linewidth,trim= 0 10 0 0,clip]{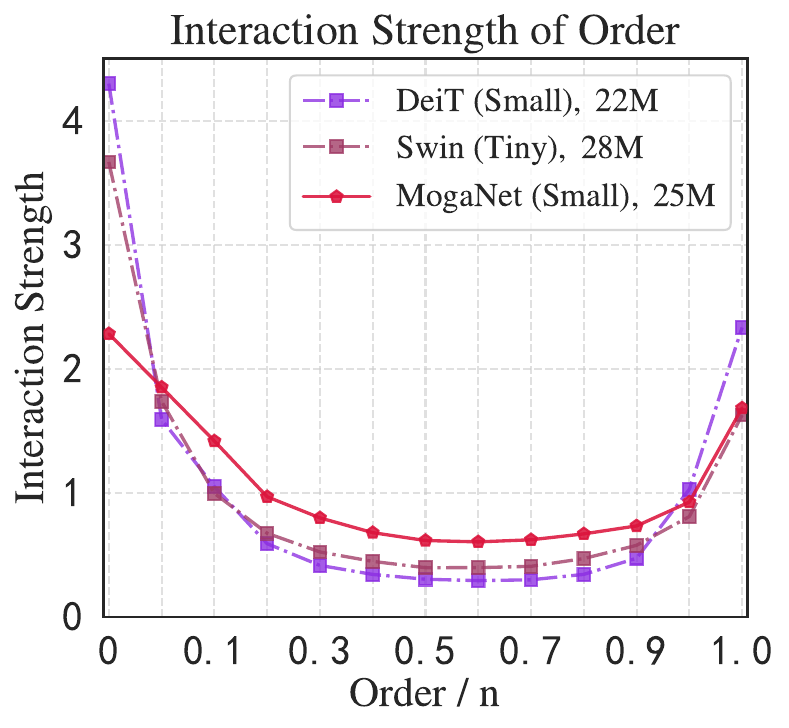}
    }
    \vspace{-1.0em}
    \caption{
    \textbf{Distributions of the interaction strength $J^{(m)}$} for \pl{(a) ConvNets with different convolution kernel sizes, (b) ConvNets with gating aggregations,} and (c) Transformers on ImageNet-1K with $224^2$ resolutions. Middle-order strengths mean the middle-complex interaction, where a medium number of patches (\textit{e.g.,} 0.2$\sim$0.8n) participate.
    }
    \label{fig:spatial_interaction_app}
    \vspace{-1.0em}
\end{figure*}

Formally, given an input image $x$ with a set of $n$ patches $N = \{1,\dots,n\}$ (\textit{e.g.}, an image with $n$ pixels in total), the multi-order interaction $I^{(m)}(i,j)$ can be calculated as:
\begin{equation}
\begin{aligned}
    I^{(m)}(i,j) = \mathbb{E}_{S \subseteq N \setminus \{i,j\}, |S|=m}[\Delta f(i,j,S)],
\end{aligned}
    \label{eq:interaction}
\end{equation}
where $\Delta f(i,j,S) = f(S \cup \{i,j\}) - f(S \cup \{i\}) - f(S \cup \{j\}) + f(S)$. $f(S)$ indicates the score of output with patches in $N \setminus S$ kept unchanged but replaced with the baseline value~\citep{ancona2019explaining}, For example, a low-order interaction (\textit{e.g.,} $m=0.05n$) means the relatively simple collaboration between variables $i,j$ under a small range of context, while a high-order interaction (\textit{e.g.,} $m=0.95n$) corresponds to the complex collaboration under a large range of context. Then, we can measure the overall interaction complexity of deep neural networks (DNNs) by the relative interaction strength $J^{(m)}$ of the encoded $m$-th order interaction:
\begin{equation}
\begin{aligned}
    J^{(m)} = \frac{\mathbb{E}_{x \in \Omega}\mathbb{E}_{i,j}|I^{(m)}(i,j|x)|}{\mathbb{E}_{m^{'}}\mathbb{E}_{x \in \Omega}\mathbb{E}_{i,j}|I^{(m^{'})}(i,j|x)|},
\end{aligned}
    \label{eq:strength}
\end{equation}
where $\Omega$ is the set of all samples and $0\le m \ge n-2$. Note that $J^{(m)}$ is the average interaction strength over all possible patch pairs of the input samples and indicates the distribution (area under curve sums up to one) of the order of interactions of DNNs.
In Fig.~\ref{fig:spatial_interaction_app}, we calculate the interaction strength $J^{(m)}$ with Eq.~\ref{eq:strength} for the models trained on ImageNet-1K using the official implementation{\footnote{\url{https://github.com/Nebularaid2000/bottleneck}}} provided by~\citep{deng2021discovering}. Specially, we use the image of $224\times 224$ resolution as the input and calculate $J^{(m)}$ on $14\times 14$ grids, \textit{i.e.,} $n=14\times 14$. And we set the model output as $f(x_S) = \log \frac{P(\hat y = y|x_S)}{1-P(\hat y = y|x_S)}$ given the masked sample $x_S$, where $y$ denotes the ground-truth label and $P(\hat y = y|x_S)$ denotes the probability of classifying the masked sample $x_S$ to the true category.
\pl{Fig.~\ref{fig:interaction_cnns} and Fig.~\ref{fig:interaction_cnns_gating} compare existing ConvNets with large kernels or gating designs and demonstrate that MogaNet can model middle-order interactions better to learn more informative representations.}

\paragraph{Relationship of explaining works of ViTs.}
Since the thriving of ViTs in a wide range of computer vision tasks, recent studies mainly investigate the \textit{why} ViTs work from two directions: (a) Evaluation of robustness against noises finds that self-attentions~\citep{naseer2021intriguing, iclr2022how, zhou2021ibot, li2022A2MIM} or gating mechanisms~\citep{icml2022FAN} in ViTs are more robust than classical convolutional operations~\citep{simonyan2014very, cvpr2016inceptionv3}. For example, ViTs can still recognize the target object with large occlusion ratios (\textit{e.g.,} only 10$\sim$20\% visible patches) or corruption noises. This phenomenon might stem from the inherent redundancy of images and the competition property of self-attention mechanisms~\citep{nips2020linformer, icml2022Flowformer}. Several recently proposed works~\citep{cvpr2022AViT, iccv2021levit} show that ViTs can work with some essential tokens (\textit{e.g.,} 5$\sim$50\%) that are selected according to the complexity of input images by dynamic sampling strategies, which also utilize the feature selection properties of self-attentions. From the perspective of multi-order interactions, convolutions with local inductive bias (using small kernel sizes) prefer low-order interactions, while self-attentions without any inductive bias tend to learn low-order and high-order interactions.
(b) Evaluation of out-of-distribution samples reveals that both self-attention mechanisms and depth-wise convolution (DWConv) with large kernel designs share similar shape-bias tendency as human vision~\citep{2021shapebias, nips2021partial, cvpr2022replknet}, while canonical ConvNets (using convolutions with small kernel sizes) exhibit strong bias on local texture~\citep{iclr2019shapebias, hermann2020origins}. Current works \citep{cvpr2022replknet} attribute shape or texture-bias tendency to the receptive field of self-attention or convolution operations, \textit{i.e.,} an operation with the larger receptive field or more long-range dependency is more likely to be shape-bias. However, there are still gaps between shape-bias operations and human vision. Human brains~\citep{treisman1980feature, deng2021discovering} attain visual patterns and clues and conduct middle-complexity interactions to recognize objects, while a self-attention or convolution operation can only encode global or local features to conduct high or low-complexity interactions. As the existing design of DNNs only stacks regionality perception or context aggregation operations in a cascaded way, it is inevitable to encounter the representation bottleneck.

\subsection{Visualization of CAM}
\label{app:gradcam}
\begin{figure*}[t]
    \vspace{-1.0em}
    \centering
    \includegraphics[width=0.99\linewidth]{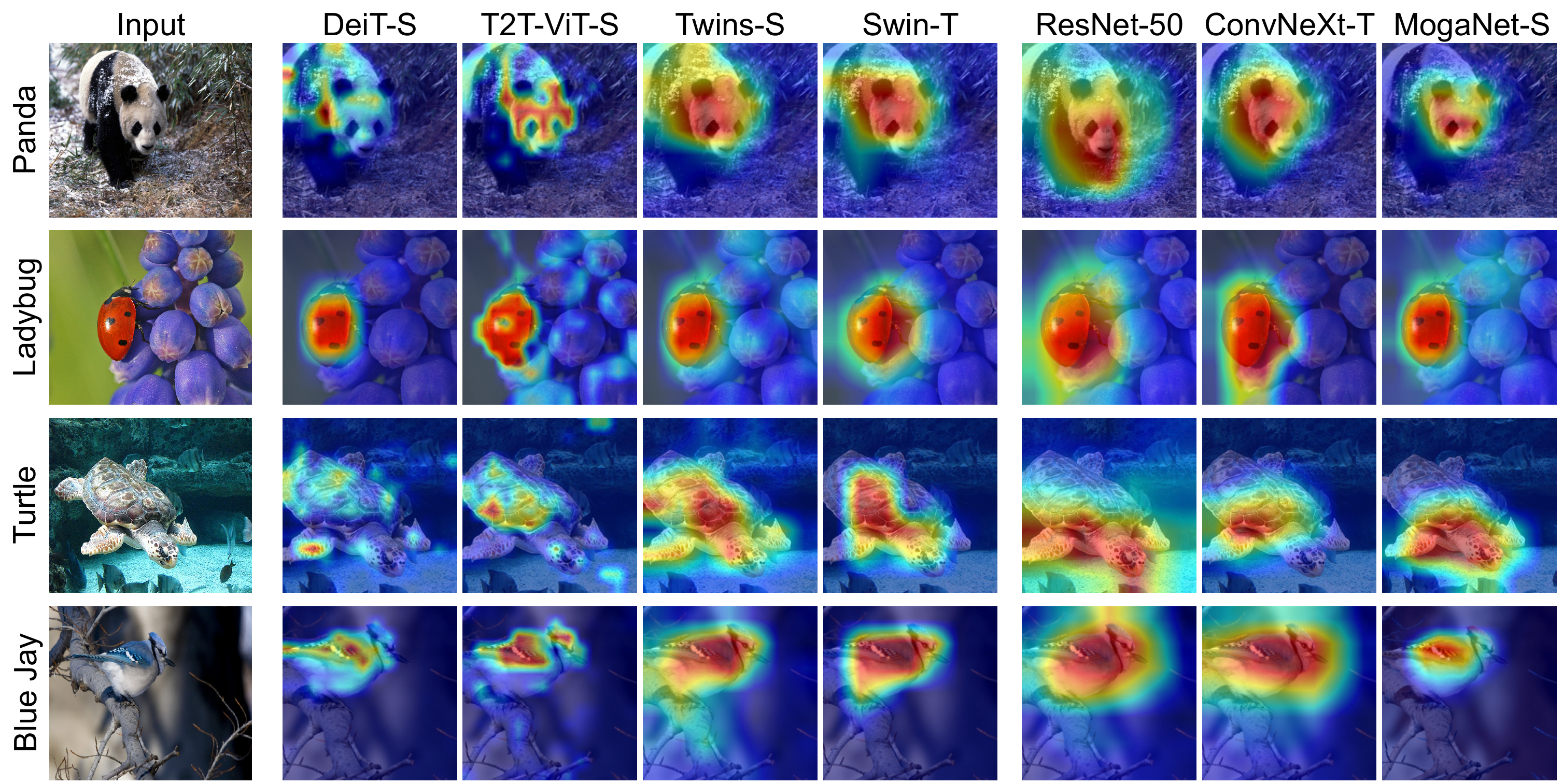}
    \vspace{-0.5em}
    \caption{
    Visualization of Grad-CAM activation maps of the models trained on ImageNet-1K.}
    \label{fig:app_gradcam}
    \vspace{-0.5em}
\end{figure*}

We further visualize more examples of Grad-CAM~\citep{cvpr2017grad} activation maps of MogaNet-S in comparison to Transformers, including DeiT-S~\citep{icml2021deit}, T2T-ViT-S~\citep{iccv2021t2t}, Twins-S~\citep{nips2021Twins}, and Swin~\citep{liu2021swin}, and ConvNets, including ResNet-50~\citep{he2016deep} and ConvNeXt-T~\citep{cvpr2022convnext}, on ImageNet-1K in Fig.~\ref{fig:app_gradcam}. Due to the self-attention mechanism, the pure Transformers architectures (DeiT-S and T2T-ViT-S) show more refined activation maps than ConvNets, but they also activate some irrelevant parts. Combined with the design of local windows, local attention architectures (Twins-S and Swin-T) can locate the full semantic objects. Results of previous ConvNets can roughly localize the semantic target but might contain some background regions.
The activation parts of our proposed MogaNet-S are more similar to local attention architectures than previous ConvNets, which are more gathered on the semantic objects.

\section{More Ablation and Analysis Results}
\label{app:ablation}
In addition to Sec.~\ref{sec:exp_ablation}, we further conduct more ablation and analysis of our proposed MogaNet on ImageNet-1K. We adopt the same experimental settings as Sec.~\ref{tab:ablation}.

\subsection{Ablation of Activation Functions}
\label{app:ablation_gating}
We conduct the ablation of activation functions used in the proposed multi-order gated aggregation module on ImageNet-1K. Table~\ref{tab:ablation_gating} shows that using SiLU~\citep{elfwing2018sigmoid} activation for both branches achieves the best performance. Similar results were also found in Transformers, \textit{e.g.,} GLU variants with SiLU or GELU~\citep{hendrycks2016bridging} yield better performances than using Sigmoid or Tanh activation functions~\citep{Shazeer2020GLU, icml2022FLASH}. We assume that SiLU is the most suitable activation because it owns both the property of Sigmoid (gating effects) and GELU (training friendly), which is defined as $x\cdot \mathrm{Sigmoid}(x)$.

\begin{figure*}[ht]
\vspace{-1.0em}
\begin{minipage}{0.35\linewidth}
\centering
    \input{Tabs/tab_ablation_gate.tex}
\end{minipage}
\begin{minipage}{0.65\linewidth}
\centering
    \input{Tabs/tab_ablation_conv.tex}
\end{minipage}
\vspace{-1.0em}
\end{figure*}

\subsection{Ablation of Multi-order DWConv Layers}
\label{app:ablation_multiorder}
In addition to Sec.~\ref{sec:moga} and Sec.~\ref{sec:exp_ablation}, we also analyze the multi-order depth-wise convolution (DWConv) layers as the static regionality perception in the multi-order aggregation module $\mathrm{Moga}(\cdot)$ on ImageNet-1K. As shown in Table~\ref{tab:ablation_conv}, we analyze the channel configuration of three parallel dilated DWConv layers: $\mathrm{DW}_{5\times 5, d=1}$, $\mathrm{DW}_{5\times 5, d=2}$, and $\mathrm{DW}_{7\times 7, d=3}$ with the channels of $C_l$, $C_m$, $C_h$.
we first compare the performance of serial DWConv layers (\textit{e.g.,} $\mathrm{DW}_{5\times 5, d=1}$+$\mathrm{DW}_{7\times 7, d=3}$) and parallel DWConv layers. We find that the parallel design can achieve the same performance with fewer computational overloads because the DWConv kernel is equally applied to all channels. When we adopt three DWConv layers, the proposed parallel design reduces $C_l+C_h$ and $C_l+C_m$ times computations of $\mathrm{DW}_{5\times 5, d=2}$ and $\mathrm{DW}_{5\times 5, d=2}$ in comparison to the serial stack of these DWConv layers. Then, we empirically explore the optimal configuration of the three channels. We find that $C_l:$ $C_m:$ $C_h$ = 1: 3: 4 yields the best performance, which well balances the small, medium, and large DWConv kernels to learn low, middle, and high-order contextual representations. We calculate and discuss the FLOPs of the proposed three DWConv layers in the next subsection to verify the efficiency. Similar conclusions are also found in relevant designs~\citep{nips2022hilo, nips2022iformer, nips2022hornet}, where global context aggregations take the majority (\textit{e.g.}, $\frac{1}{2} \sim \frac{3}{4}$ channels or context components). We also verify the parallel design with the optimal configuration based on MogaNet-S/B. Therefore, we can conclude that our proposed multi-order DWConv layers can efficiently learn multi-order contextual information for the context branch of $\mathrm{Moga}(\cdot)$.

\subsection{FLOPs and Throughputs of MogaNet}
\label{app:flops_throughput}

\paragraph{FLOPs of Multi-order Gated Aggregation Module}
We divide the computation of the proposed multi-order gated aggregation module into two parts of convolution operations and calculate the FLOPs for each part.
\begin{itemize}
    \item \textbf{Conv1$\times$1.} The FLOPs of 1$\times$1 convolution operation $\phi_{\rm gate}$ , $\phi_{\rm context}$ and $\phi_{\rm out}$ can be derived as:
    \begin{equation}
    \begin{aligned}
        \mathrm{FLOPs}(\phi_{\rm gate}) &= 2HWC^{2}, \\
        \mathrm{FLOPs}(\phi_{\rm context}) &= 2HWC^{2}, \\
        \mathrm{FLOPs}(\phi_{\rm out}) &= 2HWC^{2}.
    \vspace{-0.50em}
    \end{aligned}
    \end{equation}
    \item \textbf{Depth-wise convolution.} We consider the depth-wise convolution ($\mathrm{DW}$) with dilation ratio $d$. The $\mathrm{DW}$Conv is performed for the input $X$, where $X\in \mathbb{R}^{HW\times C_{in}}$. Therefore, the FLOPs for all $\mathrm{DW}$ in Moga module are:
    \begin{equation}
    \begin{aligned}
        \mathrm{FLOPs}(\mathrm{DW}_{5\times 5, d=1}) &= 2HWC_{in}K_{5\times 5}^{2}, \\
        \mathrm{FLOPs}(\mathrm{DW}_{5\times 5, d=2}) &= \frac{3}{4}HWC_{in}K_{5\times 5}^{2}, \\
        \mathrm{FLOPs}(\mathrm{DW}_{7\times 7, d=3}) &= HWC_{in}K_{7\times 7}^{2}.
    \vspace{-0.50em}
    \end{aligned}
    \end{equation}
\end{itemize}
Overall, the total FLOPs of our Moga module can be derived as follows:
\vspace{-0.5em}
\begin{equation}
\begin{aligned}
    \mathrm{FLOPs}(\mathrm{Moga}) &= 2HWC_{in}\left[\frac{11}{8}K_{5\times 5}^{2} + \frac{1}{2}K_{7\times 7}^{2}
    + 3C_{in}\right]\\
    &= HWC_{in}\left[\frac{471}{4} + 6C_{in}\right].
    \vspace{-0.5em}
\end{aligned}
\end{equation}

% figure: throughput
\begin{wrapfigure}{r}{0.55\linewidth}
\centering
    \vspace{-1.75em}
    \includegraphics[width=1.0\linewidth]{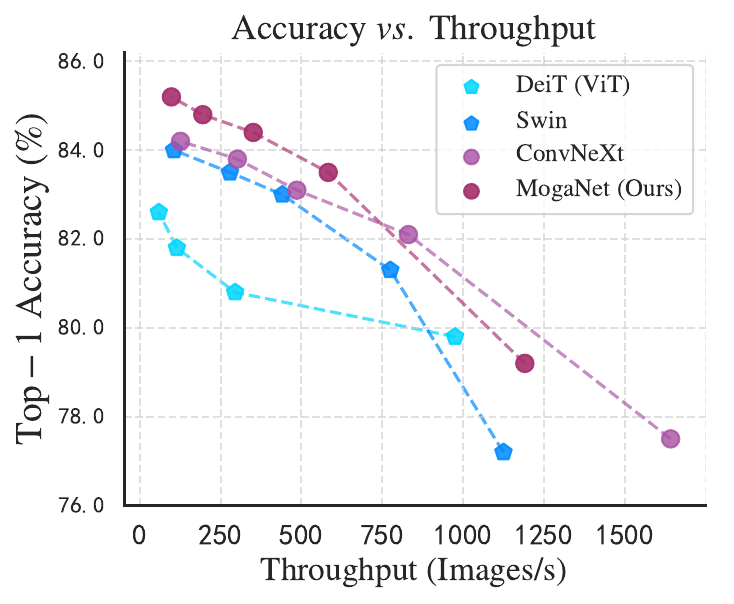}
    \vspace{-2.25em}
    \caption{Accuracy-throughput diagram of models on ImageNet-1K measured on an NVIDIA V100 GPU.}
    \label{fig:app_throughput}
    \vspace{-3.75em}
\end{wrapfigure}
% \begin{figure}[ht]
%     \centering
%     \vspace{-0.5em}
%     \includegraphics[width=0.825\linewidth]{Figs/fig_acc_throughput.pdf}
%     \vspace{-1.0em}
%     \caption{Accuracy-throughput diagram of models on ImageNet-1K measured on an NVIDIA Tesla V100 GPU.}
%     \label{fig:app_throughput}
%     \vspace{-1.5em}
% \end{figure}

\vspace{-1.0em}
\paragraph{Throughput of MogaNet}
We further analyze throughputs of MogaNet variants on ImageNet-1K. As shown in Fig.~\ref{fig:app_throughput}, MogaNet has similar throughputs as Swin Transformer while producing better performances than Swin and ConvNet. Since we add channel splitting and GAP operations in MogaNet, the throughput of ConvNeXt exceeds MogaNet to some extent.

% figure: gradcam

\subsection{Ablation of Normalization Layers}
\label{app:ablation_norm}
For most ConvNets, BatchNorm~\citep{nips2015batchnorm} (BN) is considered an essential component to improve the convergence speed and prevent overfitting. However, BN might cause some instability~\citep{Wu2021PreciseBN} or harm the final performance of models~\citep{iclr2021characterizing, Brock2021NFNet}. Some recently proposed ConvNets~\citep{cvpr2022convnext, guo2022van} replace BN by LayerNorm~\citep{2016layernorm} (LN), which has been widely used in Transformers~\citep{iclr2021vit} and Metaformer architectures~\citep{yu2022metaformer}, achieving relatively good performances in various scenarios. Here, we conduct an ablation of normalization (Norm) layers in MogaNet on ImageNet-1K, as shown in Table~\ref{tab:ablation_norm}. As discussed in ConvNeXt~\citep{cvpr2022convnext}, the Norm layers used in each block (\textbf{within}) and after each stage (\textbf{after}) have different effects. Thus we study them separately. Table~\ref{tab:ablation_norm} shows that using BN in both places yields better performance than using LN (after) and BN (within), except MogaNet-T with $224^2$ resolutions, while using LN in both places performs the worst.
Consequently, we use BN as the default Norm layers in our proposed MogaNet for two reasons: (\romannumeral1) With pure convolution operators, the rule of combining convolution operations with BN within each stage is still useful for modern ConvNets. (\romannumeral2) Although using LN after each stage might help stabilize the training process of Transformers and hybrid models and might sometimes bring good performance for ConvNets, adopting BN after each stage in pure convolution models still yields better performance.
Moreover, we replace BN with precise BN~\citep{Wu2021PreciseBN} (pBN), which is an optimal alternative normalization strategy to BN. We find slight performance improvements (around 0.1\%), especially when MogaNet-S/B adopts the EMA strategy (by default), indicating that we can further improve MogaNet with advanced BN. As discussed in ConvNeXt, EMA might severely hurt the performances of models with BN. This phenomenon might be caused by the unstable and inaccurate BN statistics estimated by EMA in the vanilla BN with large models, which will deteriorate when using another EMA of model parameters. We solve this dilemma by exponentially increasing the EMA decay from 0.9 to 0.9999 during training as momentum-based contrastive learning methods~\citep{iccv2021dino, bao2021beit}, \textit{e.g.,} BYOL \citep{nips2020byol}. It can also be tackled by advanced BN variants~\citep{NIPS2017GhostBN, Wu2021PreciseBN}.

\input{Tabs/tab_ablation_norm.tex}

\subsection{Refined Training Settings for Lightweight Models}
\label{app:advanced_tiny}
To explore the full power of lightweight models of our MogaNet, we refined the basic training settings for MogaNet-XT/T according to RSB A2~\citep{wightman2021rsb} and DeiT-III~\citep{eccv2022deit3}. Compared to the default setting as provided in Table~\ref{tab:in1k_config}, we only adjust the learning rate and the augmentation strategies for faster convergence while keeping other settings unchanged. As shown in Table~\ref{tab:advanced_tiny}, MogaNet-XT/T gain +0.4$\sim$0.6\% when use the large learning rate of $2\times 10^{-3}$ and 3-Augment~\citep{eccv2022deit3} without complex designs. Based on the advanced setting, MogaNet with $224^2$ input resolutions yields significant performance improvements against previous methods, \textit{e.g.,} MogaNet-T gains +3.5\% over DeiT-T~\citep{icml2021deit} and +1.2\% over Parc-Net-S~\citep{eccv2022edgeformer}.
Especially, MogaNet-T with $256^2$ resolutions achieves top-1 accuracy of 80.0\%, outperforming DeiT-S of 79.8\% reported in the original paper, while MogaNet-XT with $224^2$ resolutions outperforms DeiT-T under the refined training scheme by 1.2\% with only 3M parameters.

\input{Tabs/tab_advanced_tiny.tex}

\input{Tabs/tab_coco_r_1x_app}

\section{More Comparison Experiments}
\label{app:comparison}
\subsection{Fast Training on ImageNet-1K}
\label{app:exp_in1k}
In addition to Sec.~\ref{sec:exp_in1k}, we further provide comparison results for 100 and 300 epochs training on ImageNet-1K. As for 100-epoch training, we adopt the original RSB A3~\citep{wightman2021rsb} setting for all methods, which adopts LAMB \citep{iclr2020lamb} optimizer and a small training resolution of $160^2$. We search the basic learning in \{$0.006, 0.008$\} for all architectures and adopt the gradient clipping for Transformer-based networks. As for 300-epoch training, we report results of RSB A2 \citep{wightman2021rsb} for classical CNN or the original setting for Transformers or modern ConvNets. In Table~\ref{tab:in1k_app_rsb}, when compared with models of similar parameter size, our proposed MogaNet-XT/T/S/B achieves the best performance in both 100 and 300 epochs training. Results of 100-epoch training show that MogaNet has a faster convergence speed than previous architectures of various types. For example, MogaNet-T outperforms EfficientNet-B0 and DeiT-T by 2.4\% and 8.7\%, MogaNet-S outperforms Swin-T by 3.4\%, and MogaNet-B outperforms Swin-S by 2.0\%. Notice that ConvNeXt variants have a great convergence speed, \textit{e.g.,} ConvNeXt-S achieves 81.7\% surpassing Swin-S by 1.5 and recently proposed ConvNet HorNet-S$_{7\times 7}$ by 0.5 with similar parameters. But our proposed MogaNet convergences faster than ConvNet, \textit{e.g.,} MogaNet-S outperforms ConvNeXt-T by 2.3\% with similar parameters while MogaNet-B/L reaching competitive performances as ConvNeXt-B/L with only 44$\sim$50\% parameters.

\input{Tabs/tab_coco_m_1x_app}
\input{Tabs/tab_coco_c_3x_app}

\subsection{Detection and Segmentation Results on COCO}
\label{app:exp_det_coco}
In addition to Sec.~\ref{sec:exp_det_seg}, we provide full results of object detection and instance segmentation tasks with RetinaNet, Mask R-CNN, and Cascade Mask R-CNN on COCO.
As shown in Table~\ref{tab:coco_r_1x_app} and Table~\ref{tab:coco_m_1x_app}, RetinaNet or Mask R-CNN with MogaNet variants outperforms existing models when training $1\times$ schedule. For example, RetinaNet with MogaNet-T/S/B/L achieve 45.8/47.7/48.7 AP$^b$, outperforming PVT-T/S/M and PVTV2-B1/B2/B3/B5 by 4.7/4.6/5.8 and 0.3/1.2/1.7/2.6 AP$^b$; Nask R-CNN with MogaNet-S/B/L achieve 46.7/47.9/49.4 AP$^b$, exceeding Swin-T/S/B and ConvNeXt-T/S/B by 4.5/3.1/2.5 and 2.5/2.5/2.4 with similar parameters and computational overloads. Noticeably, MogaNet-XT/T can achieve better detection results with fewer parameters and lower FLOPs than lightweight architectures, while MogaNet-T even surpasses some Transformers like Swin-S and PVT-S. For example, Mask R-CNN with MogaNet-T improves Swin-T by 0.4 AP$^b$ and outperforms PVT-S by 1.3 AP$^m$ using only around 2/3 parameters.
As shown in Table~\ref{tab:coco_c_3x_app}, Cascade Mask R-CNN with MogaNet variants still achieves the state-of-the-art detection and segmentation results when training $3\times$ schedule with multi-scaling (MS) and advanced augmentations. For example, MogaNet-L/XL yield 53.3/56.2 AP$^b$ and 46.1/48.8 AP$^m$, which improves Swin-B/L and ConvNeXt-B/L by 1.4/2.3 and 0.6/1.4 AP$^b$ with similar parameters and FLOPS.

\subsection{Sementic Segmentation Results on ADE20K}
\label{app:exp_seg_ade20k}
In addition to Sec.~\ref{sec:exp_det_seg}, we provide comprehensive comparison results of semantic segmentation based on UperNet on ADE20K. As shown in Table~\ref{tab:ade20k_upernet_app}, UperNet with MogaNet produces state-of-the-art performances in a wide range of parameter scales compared to famous Transformer, hybrid, and convolution models. As for the lightweight models, MogaNet-XT/T significantly improves ResNet-18/50 with fewer parameters and FLOPs budgets. As for medium-scaling models, MogaNet-S/B achieves 49.2/50.1 mIoU$^{ss}$, which outperforms the recently proposed ConvNets, \textit{e.g.,} +1.1 over HorNet-T using similar parameters and +0.7 over SLaK-S using 17M fewer parameters. As for large models, MogaNet-L/XL surpass Swin-B/L and ConvNeXt-B/L by 1.2/1.9 and 1.8/0.3 mIoU$^{ss}$ while using fewer parameters.

\input{Tabs/tab_ade20k_uper_app}
\input{Tabs/tab_3d_app}

\subsection{2D Human Pose Estimation Results on COCO}
\label{app:exp_2d_pose}
In addition to Sec.~\ref{sec:exp_det_seg}, we provide comprehensive experiment results of 2D human key points estimation based on Top-Down SimpleBaseline on COCO.
As shown in Table~\ref{tab:coco_pose_app}, MogaNet variants achieve competitive or state-of-the-art performances compared to popular architectures with two types of resolutions. As for lightweight models, MogaNet-XT/T significantly improves the performances of existing models while using similar parameters and FLOPs. Meanwhile, MogaNet-S/B also produces 74.9/75.3 and 76.4/77.3 AP using $256\times 192$ and $384\times 288$ resolutions, outperforming Swin-B/L by 2.0/1.0 and 1.5/1.0 AP with nearly half of the parameters and computation budgets.

\input{Tabs/tab_coco_pose_app}

\subsection{3D Human Pose Estimation Results}
\label{app:exp_3d_pose}
In addition to Sec.~\ref{sec:exp_det_seg}, we evaluate popular ConvNets and MogaNet for 3D human pose estimation tasks based on ExPose~\citep{eccv2020ExPose}. As shown in Table~\ref{tab:3d_app}, MogaNet achieves lower regression errors with efficient usage of parameters and computational overheads. Compared to lightweight architectures, MogaNet-T achieves 6.82 MPJPE and 2.36 3DRMSE on hand and face reconstruction tasks, improving ResNet-18 and MobileNetV2 $1\times$ by 1.29/0.04 and 1.51/0.28. Compared to models around 25$\sim$50M parameters, MogaNet-S surpasses ResNet-101 and ConvNeXt-T, achieving competitive results as ConvNeXt-S with relatively smaller parameters and FLOPs (\textit{e.g.,} 27M/6.5G \textit{vs} 52M/11.4G on FFHP). Notice that some backbones with more parameters produce worse results than their lightweight variants on the face estimation tasks (\textit{e.g.,} ResNet-50 and Swin-S), while MogaNet-S still yields the better performance of 2.24 3DRMSE.

\clearpage

\subsection{Video Prediction Results on Moving MNIST}
\label{app:exp_vp_mmnist}
In addition to Sec.~\ref{sec:exp_det_seg}, We verify video prediction performances of various architectures by replacing the hidden translator in SimVP with the architecture blocks. All models use the same number of network blocks and have similar parameters and FLOPs. As shown in Table~\ref{tab:vp_app}, Compared to Transformer-based and Metaformer-based architectures, pure ConvNets usually achieve lower prediction errors. When training 200 epochs, it is worth noticing that using MogaNet blocks in SimVP significantly improves the SimVP baseline by 6.58/13.86 MSE/MAE and outperforms ConvNeXt and HorNet by 1.37 and 4.07 MSE. MogaNet also holds the best performances in the extended 2000-epoch training setting.

\input{Tabs/tab_vp_app}

% Table: in1k RSB A3
\input{Tabs/tab_in1k_app_rsb.tex}

\section{\pl{Extensive Related Work}}
\label{sec:related_work}
\paragraph{Convolutional Neural Networks}
% History of Pre-ViT ConvNets and modern ConvNets
ConvNets~\citep{lecun1998gradient, Krizhevsky2012ImageNetCW, he2016deep} have dominated a wide range of computer vision (CV) tasks for decades. 
VGG~\citep{simonyan2014very} proposes a modular network design strategy, stacking the same type of blocks repeatedly, which simplifies both the design workflow and transfer learning for downstream tasks.
ResNet~\citep{he2016deep} introduces identity skip connections and bottleneck modules that alleviate training difficulties (\textit{e.g.,} vanishing gradient). With the desired properties, ResNet and its variants~\citep{bmvc2016wrn, xie2017aggregated, hu2018squeeze, cvpr2022resnest} have become the most widely adopted ConvNet architectures in numerous CV applications. 
For practical usage, efficient models~\citep{eccv2018shufflenet, 2017MobileNet, cvpr2018mobilenetv2, iccv2019mobilenetv3, icml2019efficientnet, cvpr2020regnet} are designed for a complexity-accuracy trade-off and hardware devices.
Since the limited reception fields, spatial and temporal convolutions struggle to capture global dependency \citep{Luo2016ERF}. 
Various spatial-wise or channel-wise attention strategies~\citep{iccv2017Deformable, hu2018squeeze, wang2018non, eccv2018CBAM, iccv2019GCNet} are introduced.
Recently, taking the merits of Transformer-like macro design~\citep{iclr2021vit}, modern ConvNets~\citep{2022convmixer, cvpr2022replknet, Liu2022SLak, nips2022hornet, iccv2023Oriented1D} show thrilling performance with large depth-wise convolutions~\citep{han2021demystifying} for global contextual features.
% Among them, HorNet~\citep{nips2022hornet} and FocalNet~\citep{nips2022focalnet} exploit . However, these methods memorize and contextualize features in the regions with pairwise operations repeatedly in parallel, which leads to high computational complexity.
\pl{
Among them, VAN~\citep{guo2022van}, FocalNet~\citep{nips2022focalnet}, HorNet~\citep{nips2022hornet}, and Conv2Former~\citep{Hou2022Conv2Former} exploit multi-scale convolutional kernels with gating operations. However, these methods fail to ensure the networks learn the inherently overlooked features \citep{deng2021discovering} and achieve ideal contextual aggregation. Unlike the previous works, we first design three groups of multi-order depth-wise convolutions in parallel followed by a double-branch activated gating operation, and then propose a channel aggregation module to enforce the network to learn informative features of various interaction scales.
% However, these methods memorize and contextualize features in the regions with pairwise operations repeatedly in parallel, which leads to high computational complexity.
}

\vspace{-1.0em}
\paragraph{\pl{Vision Transformers}}
\pl{
Transformer~\citep{vaswani2017attention} with self-attention mechanism has become the mainstream choice in natural language processing (NLP) community~\citep{devlin2018bert, brown2020language}.
Considering that global information is also essential for CV tasks, Vision Transformer (ViT)~\citep{iclr2021vit} is proposed and has achieved promising results on ImageNet~\citep{cvpr2009imagenet}. In particular, ViT splits raw images into non-overlapping fixed-size patches as visual tokens to capture long-range feature interactions among these tokens by self-attention. By introducing regional inductive bias, ViT and its variants have been extended to various vision tasks \cite{carion2020end, zhu2020deformable, chen2021pre, parmar2018image, jiang2021transgan, arnab2021vivit}. Equipped with advanced training strategies~\citep{icml2021deit, eccv2022deit3} or extra knowledge~\citep{nips2021TL, Lin2022SuperViT, eccv2022tinyvit}, pure ViTs can achieve competitive performance as ConvNets in CV tasks. 
In the literature of \cite{yu2022metaformer}, the MetaFormer architecture substantially influenced the design of vision backbones, and all Transformer-like models~\citep{icml2021deit, 2022convmixer, aaai2022shiftvit} are classified by how they treat the token-mixing approaches, such as relative position encoding~\citep{wu2021rethinking}, local window shifting~\citep{liu2021swin} and MLP layer~\citep{nips2021MLPMixer}, \textit{etc.} 
Beyond the aspect of macro design, \cite{touvron2021training, yuan2021tokens} introduced knowledge distillation and progressive tokenization to boost training data efficiency. 
% Hybird ViTs: Swin, Uniformer, Next-ViT
Compared to ConvNets banking on the inherent inductive biases (\textit{e.g.,} locality and translation equivariance), the pure ViTs are more over-parameterized and rely on large-scale pre-training~\citep{iclr2021vit, Li2023MIMSurvey} by contrastive learning~\citep{cvpr2020moco, zang2022dlme, tnnls2023genurl} or masked image modeling~\citep{bao2021beit, cvpr2022mae, li2022A2MIM, Woo2023ConvNeXtV2} to a great extent. Targeting this problem, one branch of researchers proposes lightweight ViTs~\citep{nips2021vitc, iclr2022mobilevit, nips2022EfficientFormer, chen2022CFViT} with more efficient self-attentions variants~\citep{nips2020linformer}.
Meanwhile, the incorporation of self-attention and convolution as a hybrid backbone has been vigorously studied~\citep{guo2021cmt, wu2021cvt, nips2021coatnet, d2021convit, iclr2022uniformer, aaai2022LIT, nips2022iformer} for imparting regional priors to ViTs.
}

%% file: Tabs/tab_architecture.tex
\begin{table}[ht]
    \vspace{-0.5em}
    \setlength{\tabcolsep}{0.3mm}
    \centering
\resizebox{0.65\linewidth}{!}{
\begin{tabular}{ccc|cccccc}
\toprule
\multicolumn{1}{c|}{Stage}               & \multicolumn{1}{c|}{Output}                                            & Layer         & \multicolumn{6}{c}{MogaNet}                                                                                                                          \\ \cline{4-9}
\multicolumn{1}{c|}{}                    & \multicolumn{1}{c|}{Size}                                              & Settings      & \multicolumn{1}{c|}{XTiny}               & \multicolumn{1}{c|}{Tiny}                & \multicolumn{1}{c|}{Small}               & \multicolumn{1}{c|}{Base}                & \multicolumn{1}{c|}{Large} & XLarge   \\ \hline
\multicolumn{1}{c|}{\multirow{4}{*}{S1}} & \multicolumn{1}{c|}{\multirow{4}{*}{$\frac{H\times W}{4\times 4}$}}    & Stem          & \multicolumn{6}{c}{\begin{tabular}[c]{@{}c@{}}$\rm{Conv}_{3\times 3},~\rm{stride}~2, C/2$ \\ $\rm{Conv}_{3\times 3},~\rm{stride}~2, C$\end{tabular}} \\ \cline{3-9}
\multicolumn{1}{c|}{}                    & \multicolumn{1}{c|}{}                                                  & Embed. Dim.   & \multicolumn{1}{c|}{32}                  & \multicolumn{1}{c|}{32}                  & \multicolumn{1}{c|}{64}                  & \multicolumn{1}{c|}{64}                  & \multicolumn{1}{c|}{64}    & 96       \\ \cline{3-9}
\multicolumn{1}{c|}{}                    & \multicolumn{1}{c|}{}                                                  & \# Moga Block & \multicolumn{1}{c|}{3}                   & \multicolumn{1}{c|}{3}                   & \multicolumn{1}{c|}{2}                   & \multicolumn{1}{c|}{4}                   & \multicolumn{1}{c|}{4}     & 6        \\ \cline{3-9}
\multicolumn{1}{c|}{}                    & \multicolumn{1}{c|}{}                                                  & MLP Ratio     & \multicolumn{6}{c}{8}                                                                                                                                \\ \hline
\multicolumn{1}{c|}{\multirow{4}{*}{S2}} & \multicolumn{1}{c|}{\multirow{4}{*}{$\frac{H\times W}{8\times 8}$}}    & Stem          & \multicolumn{6}{c}{$\rm{Conv}_{3\times 3}, \rm{stride}~2$}                                                                                           \\ \cline{3-9}
\multicolumn{1}{c|}{}                    & \multicolumn{1}{c|}{}                                                  & Embed. Dim.   & \multicolumn{1}{c|}{64}                  & \multicolumn{1}{c|}{64}                  & \multicolumn{1}{c|}{128}                 & \multicolumn{1}{c|}{160}                 & \multicolumn{1}{c|}{160}   & 192      \\ \cline{3-9}
\multicolumn{1}{c|}{}                    & \multicolumn{1}{c|}{}                                                  & \# Moga Block & \multicolumn{1}{c|}{3}                   & \multicolumn{1}{c|}{3}                   & \multicolumn{1}{c|}{3}                   & \multicolumn{1}{c|}{6}                   & \multicolumn{1}{c|}{6}     & 6        \\ \cline{3-9}
\multicolumn{1}{c|}{}                    & \multicolumn{1}{c|}{}                                                  & MLP Ratio     & \multicolumn{6}{c}{8}                                                                                                                                \\ \hline
\multicolumn{1}{c|}{\multirow{4}{*}{S3}} & \multicolumn{1}{c|}{\multirow{4}{*}{$\frac{H\times W}{16\times 16}$}}  & Stem          & \multicolumn{6}{c}{$\rm{Conv}_{3\times 3},~\rm{stride}~2$}                                                                                           \\ \cline{3-9}
\multicolumn{1}{c|}{}                    & \multicolumn{1}{c|}{}                                                  & Embed. Dim.   & \multicolumn{1}{c|}{96}                  & \multicolumn{1}{c|}{128}                 & \multicolumn{1}{c|}{320}                 & \multicolumn{1}{c|}{320}                 & \multicolumn{1}{c|}{320}   & 480      \\ \cline{3-9}
\multicolumn{1}{c|}{}                    & \multicolumn{1}{c|}{}                                                  & \# Moga Block & \multicolumn{1}{c|}{10}                  & \multicolumn{1}{c|}{12}                  & \multicolumn{1}{c|}{12}                  & \multicolumn{1}{c|}{22}                  & \multicolumn{1}{c|}{44}    & 44       \\ \cline{3-9}
\multicolumn{1}{c|}{}                    & \multicolumn{1}{c|}{}                                                  & MLP Ratio     & \multicolumn{6}{c}{4}                                                                                                                                \\ \hline
\multicolumn{1}{c|}{\multirow{4}{*}{S4}} & \multicolumn{1}{c|}{\multirow{4}{*}{$\frac{H\times W}{32\times 32}$}}  & Stem          & \multicolumn{6}{c}{$\rm{Conv}_{3\times 3},~\rm{stride}~2$}                                                                                           \\ \cline{3-9}
\multicolumn{1}{c|}{}                    & \multicolumn{1}{c|}{}                                                  & Embed. Dim.   & \multicolumn{1}{c|}{192}                 & \multicolumn{1}{c|}{256}                 & \multicolumn{1}{c|}{512}                 & \multicolumn{1}{c|}{512}                 & \multicolumn{1}{c|}{640}   & 960      \\ \cline{3-9}
\multicolumn{1}{c|}{}                    & \multicolumn{1}{c|}{}                                                  & \# Moga Block & \multicolumn{1}{c|}{2}                   & \multicolumn{1}{c|}{2}                   & \multicolumn{1}{c|}{2}                   & \multicolumn{1}{c|}{3}                   & \multicolumn{1}{c|}{4}     & 4        \\ \cline{3-9}
\multicolumn{1}{c|}{}                    & \multicolumn{1}{c|}{}                                                  & MLP Ratio     & \multicolumn{6}{c}{4}                                                                                                                                \\ \hline
\multicolumn{3}{c|}{Classifier}                                                                                                   & \multicolumn{6}{c}{Global Average Pooling, Linear}                                                                                                   \\ \hline
\multicolumn{3}{c|}{Parameters (M)}                                                                                               & \multicolumn{1}{c|}{2.97}                & \multicolumn{1}{c|}{5.20}                & \multicolumn{1}{c|}{25.3}                & \multicolumn{1}{c|}{43.8}                & \multicolumn{1}{c|}{82.5}  & 180.8    \\ \hline
\multicolumn{3}{c|}{FLOPs (G)}                                                                                                    & \multicolumn{1}{c|}{0.80}                & \multicolumn{1}{c|}{1.10}                & \multicolumn{1}{c|}{4.97}                & \multicolumn{1}{c|}{9.93}                & \multicolumn{1}{c|}{15.9}  & 34.5     \\
\bottomrule
\end{tabular}
    }
    \vspace{-0.5em}
    \caption{Architecture configurations of MogaNet variants.}
    \label{tab:app_architecture}
    \vspace{-1.0em}
\end{table}

%% file: Tabs/tab_train_conf.tex
\begin{table}[H]
    % \vspace{-0.5em}
    \setlength{\tabcolsep}{0.3mm}
    \centering
\resizebox{1.0\linewidth}{!}{
\begin{tabular}{l|c|c|cccccc}
    \toprule
    Configuration              & DeiT              & RSB               & \multicolumn{6}{c}{MogaNet}                                        \\ \cline{4-9}
                               &                   & A2                & XT              & T       & S       & B       & L       & XL       \\ \hline
    Input resolution           & 224$^2$           & 224$^2$           & \multicolumn{6}{c}{224$^2$}                                        \\
    Epochs                     & 300               & 300               & \multicolumn{6}{c}{300}                                            \\
    Batch size                 & 1024              & 2048              & \multicolumn{6}{c}{1024}                                           \\
    Optimizer                  & AdamW             & LAMB              & \multicolumn{6}{c}{AdamW}                                          \\
    AdamW $(\beta_1, \beta_2)$ & \small{$0.9, 0.999$} & -              & \multicolumn{6}{c}{$0.9, 0.999$}                                   \\
    Learning rate              & 0.001             & 0.005             & \multicolumn{6}{c}{0.001}                                          \\
    Learning rate decay        & Cosine            & Cosine            & \multicolumn{6}{c}{Cosine}                                         \\
    Weight decay               & 0.05              & 0.02              & 0.03            & 0.04    & 0.05    & 0.05    & 0.05    & 0.05     \\
    Warmup epochs              & 5                 & 5                 & \multicolumn{6}{c}{5}                                              \\
    Label smoothing $\epsilon$ & 0.1               & 0.1               & \multicolumn{6}{c}{0.1}                                            \\
    Stochastic Depth           & \cmarkg           & \cmarkg           & 0.05            & 0.1     & 0.1     & 0.2     & 0.3     & 0.4      \\
    Rand Augment               & 9/0.5             & 7/0.5             & 7/0.5           & 7/0.5   & 9/0.5   & 9/0.5   & 9/0.5   & 9/0.5    \\
    Repeated Augment           & \cmarkg           & \cmarkg           & \multicolumn{6}{c}{\xmarkg}                                        \\
    Mixup $\alpha$             & 0.8               & 0.1               & 0.1             & 0.1     & 0.8     & 0.8     & 0.8     & 0.8      \\
    CutMix $\alpha$            & 1.0               & 1.0               & \multicolumn{6}{c}{1.0}                                            \\
    Erasing prob.              & 0.25              & \xmarkg           & \multicolumn{6}{c}{0.25}                                           \\
    ColorJitter                & \xmarkg           & \xmarkg           & \xmarkg         & \xmarkg & 0.4     & 0.4     & 0.4     & 0.4      \\
    Gradient Clipping          & \cmarkg           & \xmarkg           & \multicolumn{6}{c}{\xmarkg}                                        \\
    EMA decay                  & \cmarkg           & \xmarkg           & \xmarkg         & \xmarkg & \cmarkg & \cmarkg & \cmarkg & \cmarkg  \\
    Test crop ratio            & 0.875             & 0.95              & \multicolumn{6}{c}{0.90}                                           \\
    \bottomrule
    \end{tabular}
    }
    \vspace{-0.5em}
    \caption{
    % Hyper-parameters for ImageNet-1K training of DeiT, RSB A2, and MogaNet.
    Hyper-parameters for ImageNet-1K training of DeiT, RSB A2, and MogaNet. We use a similar setting as RSB for the XL and T versions of MogaNet and DeiT for the other versions.
    }
    \label{tab:in1k_config}
    \vspace{-1.5em}
\end{table}

%% file: Tabs/tab_train_conf_in21k.tex
\begin{table}[H]
    % \vspace{-0.5em}
    \setlength{\tabcolsep}{1.3mm}
    \centering
\resizebox{1.0\linewidth}{!}{
\begin{tabular}{l|cccc|cccc}
    \toprule
    Configuration              & \multicolumn{4}{c|}{IN-21K PT}              & \multicolumn{4}{c}{IN-1K FT}               \\ \cline{2-9}
                               & S         & B       & L         & XL        & S         & B         & L        & XL      \\ \hline
    Input resolution           & \multicolumn{4}{c|}{224$^2$}                & \multicolumn{4}{c}{384$^2$}                \\
    Epochs                     & \multicolumn{4}{c|}{90}                     & \multicolumn{4}{c}{30}                     \\
    Batch size                 & \multicolumn{4}{c|}{1024}                   & \multicolumn{4}{c}{512}                    \\
    Optimizer                  & \multicolumn{4}{c|}{AdamW}                  & \multicolumn{4}{c}{AdamW}                  \\
    AdamW $(\beta_1, \beta_2)$ & \multicolumn{4}{c|}{$0.9, 0.999$}           & \multicolumn{4}{c}{$0.9, 0.999$}           \\
    Learning rate              & \multicolumn{4}{c|}{$1\times 10^{-3}$}      & \multicolumn{4}{c}{$5\times 10^{-5}$}      \\
    Learning rate decay        & \multicolumn{4}{c|}{Cosine}                 & \multicolumn{4}{c}{Cosine}                 \\
    Weight decay               & \multicolumn{4}{c|}{0.05}                   & \multicolumn{4}{c}{0.05}                   \\
    Warmup epochs              & \multicolumn{4}{c|}{5}                      & \multicolumn{4}{c}{0}                      \\
    Label smoothing $\epsilon$ & \multicolumn{4}{c|}{0.2}                    & 0.1       & 0.1       & 0.2      & 0.2     \\
    Stochastic Depth           & 0         & 0.1      & 0.1       & 0.1       & 0.4       & 0.6       & 0.7      & 0.8     \\
    Rand Augment               & \multicolumn{4}{c|}{9/0.5}                  & \multicolumn{4}{c}{9/0.5}                  \\
    Repeated Augment           & \multicolumn{4}{c|}{\xmarkg}                & \multicolumn{4}{c}{\xmarkg}                \\
    Mixup $\alpha$             & \multicolumn{4}{c|}{0.8}                    & \multicolumn{4}{c}{\xmarkg}                \\
    CutMix $\alpha$            & \multicolumn{4}{c|}{1.0}                    & \multicolumn{4}{c}{\xmarkg}                \\
    Erasing prob.              & \multicolumn{4}{c|}{0.25}                   & \multicolumn{4}{c}{0.25}                   \\
    ColorJitter                & \multicolumn{4}{c|}{0.4}                    & \multicolumn{4}{c}{0.4}                    \\
    Gradient Clipping          & \multicolumn{4}{c|}{\xmarkg}                & \multicolumn{4}{c}{\xmarkg}                \\
    EMA decay                  & \multicolumn{4}{c|}{\xmarkg}                & \multicolumn{4}{c}{\cmarkg}                \\
    Test crop ratio            & \multicolumn{4}{c|}{0.90}                   & \multicolumn{4}{c}{1.0}                    \\
    \bottomrule
    \end{tabular}
    }
    \vspace{-0.5em}
    \caption{Detailed training recipe for ImageNet-21K pre-training (IN-21K PT) and ImageNet-1K fine-tuning (IN-1K FT) in high resolutions for MogaNet.}
    \label{tab:in21k_config}
    \vspace{-1.5em}
\end{table}

%% file: Tabs/tab_ablation_gate.tex
\begin{table}[H]
    \vspace{-2.25em}
    \setlength{\tabcolsep}{1.1mm}
    \centering
\resizebox{1.0\linewidth}{!}{
\begin{tabular}{cl|ccc}
    \toprule
    Top-1  &                        & \multicolumn{3}{c}{Context branch}        \\
           & Acc (\%)               & None & GELU & \cellcolor{gray94}SiLU      \\ \hline
           & None                   & 76.3 & 76.7 & 76.7                        \\
    Gating & Sigmoid                & 76.8 & 77.0 & 76.9                        \\
    branch & GELU                   & 76.7 & 76.8 & 77.0                        \\
           & \cellcolor{gray94}SiLU & 76.9 & 77.1 & \cellcolor{gray94}\bf{77.2} \\
    \bottomrule
    \end{tabular}
    }
    \vspace{-0.5em}
    \caption{
    % Ablation of activation functions for the gating and context branches in the $\mathrm{Moga}(\cdot)$ module.
    Ablation of various activation functions for the gating and context branches in the proposed $\mathrm{Moga}(\cdot)$ module, which SiLU achieves the best performance in two branches.
    }
    \label{tab:ablation_gating}
    % \vspace{-1.0em}
\end{table}

%% file: Tabs/tab_ablation_conv.tex
\begin{table}[H]
    \setlength{\tabcolsep}{0.9mm}
    \centering
\resizebox{1.0\linewidth}{!}{
\begin{tabular}{l|ccc}
    \toprule
    Modules                                                                                  & Top-1     & Params. & FLOPs \\
                                                                                             & Acc (\%)  & (M)     & (G)   \\ \hline
    Baseline (+Gating branch)                                                                & 77.2      & 5.09    & 1.070 \\
    $\mathrm{DW}_{7\times 7}$                                                                & 77.4      & 5.14    & 1.094 \\
    $\mathrm{DW}_{5\times 5, d=1}+\mathrm{DW}_{7\times 7, d=3}$                              & 77.5      & 5.15    & 1.112 \\
    $\mathrm{DW}_{5\times 5, d=1}+\mathrm{DW}_{5\times 5, d=2}+\mathrm{DW}_{7\times 7, d=3}$ & 77.5      & 5.17    & 1.185 \\ \hline
    +Multi-order, $C_l: C_m: C_h=1: 0: 3$                                                    & 77.5      & 5.17    & 1.099 \\
    +Multi-order, $C_l: C_m: C_h=0: 1: 1$                                                    & 77.6      & 5.17    & 1.103 \\
    +Multi-order, $C_l: C_m: C_h=1: 6: 9$                                                    & 77.7      & 5.17    & 1.104 \\
    \rowcolor{gray94}+Multi-order, $C_l: C_m: C_h=1: 3: 4$                                   & \bf{77.8} & 5.17    & 1.102 \\
    \bottomrule
    \end{tabular}
    }
    \vspace{-0.5em}
    \caption{Ablation of multi-order DWConv layers in the proposed $\mathrm{Moga}(\cdot)$. The baseline adopts the MogaNet framework using the non-linear projection, $\mathrm{DW}_{5\times 5}$, and the SiLU gating branch as $\mathrm{SMixer}(\cdot)$ and using the vanilla MLP as $\mathrm{CMixer}(\cdot)$.}
    \label{tab:ablation_conv}
    % \vspace{-1.0em}
\end{table}

%% file: Tabs/tab_ablation_norm.tex
\begin{table}[ht]
    \vspace{-0.5em}
    \setlength{\tabcolsep}{1.5mm}
    \centering
\resizebox{0.575\linewidth}{!}{
\begin{tabular}{l|ccccc}
    \toprule
    Norm (after)    & Input   & LN   & LN        & BN                          & pBN                         \\
    Norm (within)   & size    & LN   & BN        & BN                          & pBN                         \\ \hline
    MogaNet-T       & 224$^2$ & 78.4 & \bf{79.1} & \cellcolor{gray94}79.0      & \bf{79.1}                   \\
    MogaNet-T       & 256$^2$ & 78.8 & 79.4      & \cellcolor{gray94}\bf{79.6} & \bf{79.6}                   \\
    MogaNet-S       & 224$^2$ & 82.5 & 83.2      & \cellcolor{gray94}\bf{83.3} & \bf{83.3}                   \\
    MogaNet-S (EMA) & 224$^2$ & 82.7 & 83.2      & 83.3                        & \cellcolor{gray94}\bf{83.4} \\
    MogaNet-B       & 224$^2$ & 83.4 & 83.9      & \cellcolor{gray94}84.1      & \bf{84.2}                   \\
    MogaNet-B (EMA) & 224$^2$ & 83.7 & 83.8      & 84.3                        & \cellcolor{gray94}\bf{84.4} \\
    \bottomrule
    \end{tabular}
    }
    \vspace{-0.5em}
    \caption{Ablation of normalization layers in MogaNet.}
    \label{tab:ablation_norm}
    \vspace{-0.5em}
\end{table}

%% file: Tabs/tab_advanced_tiny.tex
\begin{table}[t]
    \vspace{-1.0em}
    \setlength{\tabcolsep}{1.0mm}
    \centering
\resizebox{0.82\linewidth}{!}{
\begin{tabular}{lccccccc}
    \toprule
Architecture                                          & Input   & Learning          & Warmup & Rand    & 3-Augment            & EMA     & Top-1     \\
                                                      % & size    & rate              & epochs & Augment & \cite{eccv2022deit3} &         & Acc (\%)  \\ \hline
                                                      & size    & rate              & epochs & Augment &                      &         & Acc (\%)  \\ \hline
DeiT-T                                                & $224^2$ & $1\times 10^{-3}$ & 5      & 9/0.5   & \xmarkg              & \cmarkg & 72.2      \\
\rowcolor{gray94}DeiT-T                               & $224^2$ & $2\times 10^{-3}$ & 20     & \xmarkg & \cmarkg              & \xmarkg & 75.9      \\
ParC-Net-S                                            & $256^2$ & $1\times 10^{-3}$ & 5      & 9/0.5   & \xmarkg              & \cmarkg & 78.6      \\
\rowcolor{gray94}ParC-Net-S                           & $256^2$ & $2\times 10^{-3}$ & 20     & \xmarkg & \cmarkg              & \xmarkg & 78.8      \\ \hline
MogaNet-XT                                            & $224^2$ & $1\times 10^{-3}$ & 5      & 7/0.5   & \xmarkg              & \xmarkg & 76.5      \\
\rowcolor{gray94}MogaNet-XT                           & $224^2$ & $2\times 10^{-3}$ & 20     & \xmarkg & \cmarkg              & \xmarkg & 77.1      \\
MogaNet-XT                                            & $256^2$ & $1\times 10^{-3}$ & 5      & 7/0.5   & \xmarkg              & \xmarkg & 77.2      \\
\rowcolor{gray94}MogaNet-XT                           & $256^2$ & $2\times 10^{-3}$ & 20     & \xmarkg & \cmarkg              & \xmarkg & 77.6      \\
MogaNet-T                                             & $224^2$ & $1\times 10^{-3}$ & 5      & 7/0.5   & \xmarkg              & \xmarkg & 79.0      \\
\rowcolor{gray94}MogaNet-T                            & $224^2$ & $2\times 10^{-3}$ & 20     & \xmarkg & \cmarkg              & \xmarkg & 79.4      \\
MogaNet-T                                             & $256^2$ & $1\times 10^{-3}$ & 5      & 7/0.5   & \xmarkg              & \xmarkg & 79.6      \\
\rowcolor{gray94}MogaNet-T                            & $256^2$ & $2\times 10^{-3}$ & 20     & \xmarkg & \cmarkg              & \xmarkg & \bf{80.0} \\
    \bottomrule
    \end{tabular}
    }
    \vspace{-0.5em}
    \caption{Advanced training recipes for Lightweight models of MogaNet on ImageNet-1K.}
    \label{tab:advanced_tiny}
    \vspace{-0.5em}
\end{table}

%% file: Tabs/tab_coco_r_1x_app.tex
\begin{table}[t]
    % \vspace{-0.25em}
    \setlength{\tabcolsep}{1.5mm}
    \centering
\resizebox{0.77\linewidth}{!}{
\begin{tabular}{lccccccccc}
    \toprule
Architecture                     & Type & \#P. & FLOPs & \multicolumn{6}{c}{RetinaNet $1\times$}                               \\
                                 &      & (M)  & (G)   & AP        & AP$_{50}$ & AP$_{75}$ & AP$^{S}$  & AP$_{M}$  & AP$_{L}$  \\ \hline
RegNet-800M                      & C    & 17   & 168   & 35.6      & 54.7      & 37.7      & 19.7      & 390       & 47.8      \\
PVTV2-B0                         & T    & 13   & 160   & 37.1      & 57.2      & 39.2      & 23.4      & 40.4      & 49.2      \\
\rowcolor{gray94}\bf{MogaNet-XT} & C    & 12   & 167   & \bf{39.7} & \bf{60.0} & \bf{42.4} & \bf{23.8} & \bf{43.6} & \bf{51.7} \\ \hline
ResNet-18                        & C    & 21   & 189   & 31.8      & 49.6      & 33.6      & 16.3      & 34.3      & 43.2      \\
RegNet-1.6G                      & C    & 20   & 185   & 37.4      & 56.8      & 39.8      & 22.4      & 41.1      & 49.2      \\
RegNet-3.2G                      & C    & 26   & 218   & 39.0      & 58.4      & 41.9      & 22.6      & 43.5      & 50.8      \\
PVT-T                            & T    & 23   & 183   & 36.7      & 56.9      & 38.9      & 22.6      & 38.8      & 50.0      \\
PoolFormer-S12                   & T    & 22   & 207   & 36.2      & 56.2      & 38.2      & 20.8      & 39.1      & 48.0      \\
PVTV2-B1                         & T    & 24   & 187   & 41.1      & 61.4      & 43.8      & 26.0      & 44.6      & 54.6      \\
\rowcolor{gray94}\bf{MogaNet-T}  & C    & 14   & 173   & \bf{41.4} & \bf{61.5} & \bf{44.4} & \bf{25.1} & \bf{45.7} & \bf{53.6} \\ \hline
ResNet-50                        & C    & 37   & 239   & 36.3      & 55.3      & 38.6      & 19.3      & 40.0      & 48.8      \\
Swin-T                           & T    & 38   & 245   & 41.8      & 62.6      & 44.7      & 25.2      & 45.8      & 54.7      \\
PVT-S                            & T    & 34   & 226   & 40.4      & 61.3      & 43.0      & 25.0      & 42.9      & 55.7      \\
Twins-SVT-S                      & T    & 34   & 209   & 42.3      & 63.4      & 45.2      & 26.0      & 45.5      & 56.5      \\
Focal-T                          & T    & 39   & 265   & 43.7      & -         & -         & -         & -         & -         \\
PoolFormer-S36                   & T    & 41   & 272   & 39.5      & 60.5      & 41.8      & 22.5      & 42.9      & 52.4      \\
PVTV2-B2                         & T    & 35   & 281   & 44.6      & 65.7      & 47.6      & 28.6      & 48.5      & 59.2      \\
CMT-S                            & H    & 45   & 231   & 44.3      & 65.5      & 47.5      & 27.1      & 48.3      & 59.1      \\
\rowcolor{gray94}\bf{MogaNet-S}  & C    & 35   & 253   & \bf{45.8} & \bf{66.6} & \bf{49.0} & \bf{29.1} & \bf{50.1} & \bf{59.8} \\ \hline
ResNet-101                       & C    & 57   & 315   & 38.5      & 57.8      & 41.2      & 21.4      & 42.6      & 51.1      \\
PVT-M                            & T    & 54   & 258   & 41.9      & 63.1      & 44.3      & 25.0      & 44.9      & 57.6      \\
Focal-S                          & T    & 62   & 367   & 45.6      & -         & -         & -         & -         & -         \\
PVTV2-B3                         & T    & 55   & 263   & 46.0      & 67.0      & 49.5      & 28.2      & 50.0      & 61.3      \\
PVTV2-B4                         & T    & 73   & 315   & 46.3      & 67.0      & 49.6      & 29.0      & 50.1      & 62.7      \\
\rowcolor{gray94}\bf{MogaNet-B}  & C    & 54   & 355   & \bf{47.7} & \bf{68.9} & \bf{51.0} & \bf{30.5} & \bf{52.2} & \bf{61.7} \\ \hline
ResNeXt-101-64                   & C    & 95   & 473   & 41.0      & 60.9      & 44.0      & 23.9      & 45.2      & 54.0      \\
PVTV2-B5                         & T    & 92   & 335   & 46.1      & 66.6      & 49.5      & 27.8      & 50.2      & 62.0      \\
\rowcolor{gray94}\bf{MogaNet-L}  & C    & 92   & 477   & \bf{48.7} & \bf{69.5} & \bf{52.6} & \bf{31.5} & \bf{53.4} & \bf{62.7} \\
    \bottomrule
    \end{tabular}
    }
    \vspace{-0.5em}
    \caption{\textbf{Object detection} with RetinaNet ($1\times$ training schedule) on COCO \textit{val2017}. The FLOPs are measured at resolution $800\times 1280$.}
    \vspace{-1.5em}
    \label{tab:coco_r_1x_app}
\end{table}

%% file: Tabs/tab_coco_m_1x_app.tex
\begin{table}[t]
    \vspace{-1.5em}
    \setlength{\tabcolsep}{1.3mm}
    \centering
\resizebox{0.76\linewidth}{!}{
\begin{tabular}{lccccccccc}
    \toprule
Architecture                     & Type & \#P. & FLOPs & \multicolumn{6}{c}{Mask R-CNN $1\times$}                                              \\
                                 &      & (M)  & (G)   & AP$^{b}$  & AP$_{50}^{b}$ & AP$_{75}^{b}$ & AP$^{m}$  & AP$_{50}^{m}$ & AP$_{75}^{m}$ \\ \hline
RegNet-800M                      & C    & 27   & 187   & 37.5      & 57.9          & 41.1          & 34.3      & 56.0          & 36.8          \\
\rowcolor{gray94}\bf{MogaNet-XT} & C    & 23   & 185   & \bf{40.7} & \bf{62.3}     & \bf{44.4}     & \bf{37.6} & \bf{59.6}     & \bf{40.2}     \\ \hline
ResNet-18                        & C    & 31   & 207   & 34.0      & 54.0          & 36.7          & 31.2      & 51.0          & 32.7          \\
RegNet-1.6G                      & C    & 29   & 204   & 38.9      & 60.5          & 43.1          & 35.7      & 57.4          & 38.9          \\
PVT-T                            & T    & 33   & 208   & 36.7      & 59.2          & 39.3          & 35.1      & 56.7          & 37.3          \\
PoolFormer-S12                   & T    & 32   & 207   & 37.3      & 59.0          & 40.1          & 34.6      & 55.8          & 36.9          \\
\rowcolor{gray94}\bf{MogaNet-T}  & C    & 25   & 192   & \bf{42.6} & \bf{64.0}     & \bf{46.4}     & \bf{39.1} & \bf{61.3}     & \bf{42.0}     \\ \hline
ResNet-50                        & C    & 44   & 260   & 38.0      & 58.6          & 41.4          & 34.4      & 55.1          & 36.7          \\
RegNet-6.4G                      & C    & 45   & 307   & 41.1      & 62.3          & 45.2          & 37.1      & 59.2          & 39.6          \\
PVT-S                            & T    & 44   & 245   & 40.4      & 62.9          & 43.8          & 37.8      & 60.1          & 40.3          \\
Swin-T                           & T    & 48   & 264   & 42.2      & 64.6          & 46.2          & 39.1      & 61.6          & 42.0          \\
MViT-T                           & T    & 46   & 326   & 45.9      & \bf{68.7}     & 50.5          & 42.1      & \bf{66.0}     & 45.4          \\
PoolFormer-S36                   & T    & 32   & 207   & 41.0      & 63.1          & 44.8          & 37.7      & 60.1          & 40.0          \\
Focal-T                          & T    & 49   & 291   & 44.8      & 67.7          & 49.2          & 41.0      & 64.7          & 44.2          \\
PVTV2-B2                         & T    & 45   & 309   & 45.3      & 67.1          & 49.6          & 41.2      & 64.2          & 44.4          \\
LITV2-S                          & T    & 47   & 261   & 44.9      & 67.0          & 49.5          & 40.8      & 63.8          & 44.2          \\
CMT-S                            & H    & 45   & 249   & 44.6      & 66.8          & 48.9          & 40.7      & 63.9          & 43.4          \\
Conformer-S/16                   & H    & 58   & 341   & 43.6      & 65.6          & 47.7          & 39.7      & 62.6          & 42.5          \\
Uniformer-S                      & H    & 41   & 269   & 45.6      & 68.1          & 49.7          & 41.6      & 64.8          & 45.0          \\
ConvNeXt-T                       & C    & 48   & 262   & 44.2      & 66.6          & 48.3          & 40.1      & 63.3          & 42.8          \\
FocalNet-T (SRF)                 & C    & 49   & 267   & 45.9      & 68.3          & 50.1          & 41.3      & 65.0          & 44.3          \\
FocalNet-T (LRF)                 & C    & 49   & 268   & 46.1      & 68.2          & 50.6          & 41.5      & 65.1          & 44.5          \\
\rowcolor{gray94}\bf{MogaNet-S}  & C    & 45   & 272   & \bf{46.7} & 68.0          & \bf{51.3}     & \bf{42.2} & 65.4          & \bf{45.5}     \\ \hline
ResNet-101                       & C    & 63   & 336   & 40.4      & 61.1          & 44.2          & 36.4      & 57.7          & 38.8          \\
RegNet-12G                       & C    & 64   & 423   & 42.2      & 63.7          & 46.1          & 38.0      & 60.5          & 40.5          \\
PVT-M                            & T    & 64   & 302   & 42.0      & 64.4          & 45.6          & 39.0      & 61.6          & 42.1          \\
Swin-S                           & T    & 69   & 354   & 44.8      & 66.6          & 48.9          & 40.9      & 63.4          & 44.2          \\
Focal-S                          & T    & 71   & 401   & 47.4      & 69.8          & 51.9          & 42.8      & 66.6          & 46.1          \\
PVTV2-B3                         & T    & 65   & 397   & 47.0      & 68.1          & 51.7          & 42.5      & 65.7          & 45.7          \\
LITV2-M                          & T    & 68   & 315   & 46.5      & 68.0          & 50.9          & 42.0      & 65.1          & 45.0          \\
UniFormer-B                      & H    & 69   & 399   & 47.4      & 69.7          & 52.1          & 43.1      & 66.0          & 46.5          \\
ConvNeXt-S                       & C    & 70   & 348   & 45.4      & 67.9          & 50.0          & 41.8      & 65.2          & 45.1          \\
\rowcolor{gray94}\bf{MogaNet-B}  & C    & 63   & 373   & \bf{47.9} & \bf{70.0}     & \bf{52.7}     & \bf{43.2} & \bf{67.0}     & \bf{46.6}     \\ \hline
Swin-B                           & T    & 107  & 496   & 46.9      & 69.6          & 51.2          & 42.3      & 65.9          & 45.6          \\
PVTV2-B5                         & T    & 102  & 557   & 47.4      & 68.6          & 51.9          & 42.5      & 65.7          & 46.0          \\
ConvNeXt-B                       & C    & 108  & 486   & 47.0      & 69.4          & 51.7          & 42.7      & 66.3          & 46.0          \\
FocalNet-B (SRF)                 & C    & 109  & 496   & 48.8      & 70.7          & 53.5          & 43.3      & 67.5          & 46.5          \\
\rowcolor{gray94}\bf{MogaNet-L}  & C    & 102  & 495   & \bf{49.4} & \bf{70.7}     & \bf{54.1}     & \bf{44.1} & \bf{68.1}     & \bf{47.6}     \\
    \bottomrule
    \end{tabular}
    }
    \vspace{-0.5em}
    \caption{\textbf{Object detection and instance segmentation} with Mask R-CNN ($1\times$ training schedule) on COCO \textit{val2017}. The FLOPs are measured at resolution $800\times 1280$.}
    \vspace{-1.0em}
    \label{tab:coco_m_1x_app}
\end{table}

%% file: Tabs/tab_coco_c_3x_app.tex
\begin{table}[t!]
    \vspace{-1.0em}
    \setlength{\tabcolsep}{1.3mm}
    \centering
\resizebox{0.76\linewidth}{!}{
\begin{tabular}{lccccccccc}
    \toprule
Architecture                             & Type & \#P. & FLOPs & \multicolumn{6}{c}{Cascade Mask R-CNN +MS $3\times$}                                  \\
                                         &      & (M)  & (G)   & AP$^{bb}$ & AP$_{50}^{b}$ & AP$_{75}^{b}$ & AP$^{m}$  & AP$_{50}^{m}$ & AP$_{75}^{m}$ \\ \hline
ResNet-50                                & C    & 77   & 739   & 46.3      & 64.3          & 50.5          & 40.1      & 61.7          & 43.4          \\
Swin-T                                   & T    & 86   & 745   & 50.4      & 69.2          & 54.7          & 43.7      & 66.6          & 47.3          \\
Focal-T                                  & T    & 87   & 770   & 51.5      & 70.6          & 55.9          & -         & -             & -             \\
ConvNeXt-T                               & C    & 86   & 741   & 50.4      & 69.1          & 54.8          & 43.7      & 66.5          & 47.3          \\
FocalNet-T (SRF)                         & C    & 86   & 746   & 51.5      & 70.1          & 55.8          & 44.6      & 67.7          & 48.4          \\
\rowcolor{gray94}\bf{MogaNet-S}          & C    & 83   & 750   & \bf{51.6} & \bf{70.8}     & \bf{56.3}     & \bf{45.1} & \bf{68.7}     & \bf{48.8}     \\ \hline
ResNet-101-32                            & C    & 96   & 819   & 48.1      & 66.5          & 52.4          & 41.6      & 63.9          & 45.2          \\
Swin-S                                   & T    & 107  & 838   & 51.9      & 70.7          & 56.3          & 45.0      & 68.2          & 48.8          \\
ConvNeXt-S                               & C    & 108  & 827   & 51.9      & 70.8          & 56.5          & 45.0      & 68.4          & 49.1          \\
\rowcolor{gray94}\bf{MogaNet-B}          & C    & 101  & 851   & \bf{52.6} & \bf{72.0}     & \bf{57.3}     & \bf{46.0} & \bf{69.6}     & \bf{49.7}     \\ \hline
Swin-B                                   & T    & 145  & 982   & 51.9      & 70.5          & 56.4          & 45.0      & 68.1          & 48.9          \\
ConvNeXt-B                               & C    & 146  & 964   & 52.7      & 71.3          & 57.2          & 45.6      & 68.9          & 49.5          \\
\rowcolor{gray94}\bf{MogaNet-L}          & C    & 140  & 974   & \bf{53.3} & \bf{71.8}     & \bf{57.8}     & \bf{46.1} & \bf{69.2}     & \bf{49.8}     \\ \hline
Swin-L$^\ddag$                           & T    & 253  & 1382  & 53.9      & 72.4          & 58.8          & 46.7      & 70.1          & 50.8          \\
ConvNeXt-L$^\ddag$                       & C    & 255  & 1354  & 54.8      & 73.8          & 59.8          & 47.6      & 71.3          & 51.7          \\
ConvNeXt-XL$^\ddag$                      & C    & 407  & 1898  & 55.2      & 74.2          & 59.9          & 47.7      & 71.6          & 52.2          \\
RepLKNet-31L$^\ddag$                     & C    & 229  & 1321  & 53.9      & 72.5          & 58.6          & 46.5      & 70.0          & 50.6          \\
HorNet-L$^\ddag$                         & C    & 259  & 1399  & 56.0      & -             & -             & 48.6      & -             & -             \\
\rowcolor{gray94}\bf{MogaNet-XL}$^\ddag$ & C    & 238  & 1355  & \bf{56.2} & \bf{75.0}     & \bf{61.2}     & \bf{48.8} & \bf{72.6}     & \bf{53.3}     \\
    \bottomrule
    \end{tabular}
    }
    \vspace{-0.5em}
    \caption{\textbf{Object detection and instance segmentation} with Cascade Mask R-CNN ($3\times$ training schedule) with multi-scaling training (MS) on COCO \textit{val2017}. $^\ddag$ denotes the model is pre-trained on ImageNet-21K. The FLOPs are measured at resolution $800\times 1280$.}
    \vspace{-1.0em}
    \label{tab:coco_c_3x_app}
\end{table}

%% file: Tabs/tab_ade20k_uper_app.tex
\begin{table}[t]
    \vspace{-1.0em}
    \setlength{\tabcolsep}{1.5mm}
    \centering
\resizebox{0.74\linewidth}{!}{
\begin{tabular}{llccccc}
    \toprule
    Architecture                                 & Date      & Type & Crop    & Param. & FLOPs & mIoU$^{ss}$ \\
                                                 &           &      & size    & (M)    & (G)   & (\%)        \\ \hline
    ResNet-18                                    & CVPR'2016 & C    & 512$^2$ & 41     & 885   & 39.2        \\
    \rowcolor{gray94}\bf{MogaNet-XT}             & Ours      & C    & 512$^2$ & 30     & 856   & \bf{42.2}   \\ \hline
    ResNet-50                                    & CVPR'2016 & C    & 512$^2$ & 67     & 952   & 42.1        \\
    \rowcolor{gray94}\bf{MogaNet-T}              & Ours      & C    & 512$^2$ & 33     & 862   & \bf{43.7}   \\ \hline
    DeiT-S                                       & ICML'2021 & T    & 512$^2$ & 52     & 1099  & 44.0        \\
    Swin-T                                       & ICCV'2021 & T    & 512$^2$ & 60     & 945   & 46.1        \\
    TwinsP-S                                     & NIPS'2021 & T    & 512$^2$ & 55     & 919   & 46.2        \\
    Twins-S                                      & NIPS'2021 & T    & 512$^2$ & 54     & 901   & 46.2        \\
    Focal-T                                      & NIPS'2021 & T    & 512$^2$ & 62     & 998   & 45.8        \\
    Uniformer-S$_{h32}$                          & ICLR'2022 & H    & 512$^2$ & 52     & 955   & 47.0        \\
    UniFormer-S                                  & ICLR'2022 & H    & 512$^2$ & 52     & 1008  & 47.6        \\
    ConvNeXt-T                                   & CVPR'2022 & C    & 512$^2$ & 60     & 939   & 46.7        \\
    FocalNet-T (SRF)                             & NIPS'2022 & C    & 512$^2$ & 61     & 944   & 46.5        \\
    HorNet-T$_{7\times 7}$                       & NIPS'2022 & C    & 512$^2$ & 52     & 926   & 48.1        \\
    \rowcolor{gray94}\bf{MogaNet-S}              & Ours      & C    & 512$^2$ & 55     & 946   & \bf{49.2}   \\ \hline
    Swin-S                                       & ICCV'2021 & T    & 512$^2$ & 81     & 1038  & 48.1        \\
    Twins-B                                      & NIPS'2021 & T    & 512$^2$ & 89     & 1020  & 47.7        \\
    Focal-S                                      & NIPS'2021 & T    & 512$^2$ & 85     & 1130  & 48.0        \\
    Uniformer-B$_{h32}$                          & ICLR'2022 & H    & 512$^2$ & 80     & 1106  & 49.5        \\
    ConvNeXt-S                                   & CVPR'2022 & C    & 512$^2$ & 82     & 1027  & 48.7        \\
    FocalNet-S (SRF)                             & NIPS'2022 & C    & 512$^2$ & 83     & 1035  & 49.3        \\
    SLaK-S                                       & ICLR'2023 & C    & 512$^2$ & 91     & 1028  & 49.4        \\
    \rowcolor{gray94}\bf{MogaNet-B}              & Ours      & C    & 512$^2$ & 74     & 1050  & \bf{50.1}   \\ \hline
    Swin-B                                       & ICCV'2021 & T    & 512$^2$ & 121    & 1188  & 49.7        \\
    Focal-B                                      & NIPS'2021 & T    & 512$^2$ & 126    & 1354  & 49.0        \\
    ConvNeXt-B                                   & CVPR'2022 & C    & 512$^2$ & 122    & 1170  & 49.1        \\
    RepLKNet-31B                                 & CVPR'2022 & C    & 512$^2$ & 112    & 1170  & 49.9        \\
    FocalNet-B (SRF)                             & NIPS'2022 & C    & 512$^2$ & 124    & 1180  & 50.2        \\
    SLaK-B                                       & ICLR'2023 & C    & 512$^2$ & 135    & 1185  & 50.2        \\
    \rowcolor{gray94}\bf{MogaNet-L}              & Ours      & C    & 512$^2$ & 113    & 1176  & \bf{50.9}   \\ \hline
    Swin-L$^\ddag$                               & ICCV'2021 & T    & 640$^2$ & 234    & 2468  & 52.1        \\
    ConvNeXt-L$^\ddag$                           & CVPR'2022 & C    & 640$^2$ & 245    & 2458  & 53.7        \\
    RepLKNet-31L$^\ddag$                         & CVPR'2022 & C    & 640$^2$ & 207    & 2404  & 52.4        \\
    \rowcolor{gray94}\bf{MogaNet-XL}$^\ddag$     & Ours      & C    & 640$^2$ & 214    & 2451  & \bf{54.0}   \\
    \bottomrule
    \end{tabular}
    }
    \vspace{-0.5em}
    \caption{\textbf{Semantic segmentation} with UperNet (160K) on ADE20K validation set. $^\ddag$ indicates using IN-21K pre-trained models. The FLOPs are measured at $512\times 2048$ or $640\times 2560$ resolutions.}
    \vspace{-0.5em}
    \label{tab:ade20k_upernet_app}
\end{table}

%% file: Tabs/tab_3d_app.tex
\begin{table}[t!]
    % \vspace{-0.25em}
    \setlength{\tabcolsep}{1.1mm}
    \centering
\resizebox{0.74\linewidth}{!}{
\begin{tabular}{l|c|ccc|ccc}
    \toprule
    Architecture                           &      & \multicolumn{3}{c|}{Hand}       & \multicolumn{3}{c}{Face}    \\
                                           & Type & \#P. & FLOPs & PA-MPJPE         & \#P. & FLOPs & 3DRMSE       \\
                                           &      & (M)  & (G)   & (mm)$\downarrow$ & (M)  & (G)   & $\downarrow$ \\ \hline
    MobileNetV2                            & C    & 4.8  & 0.3   & 8.33             & 4.9  & 0.4   & 2.64         \\
    ResNet-18                              & C    & 13.0 & 1.8   & 7.51             & 13.1 & 2.4   & 2.40         \\
    \rowcolor{gray94}\bf{MogaNet-T}        & C    & 6.5  & 1.1   & \bf{6.82}        & 6.6  & 1.5   & \bf{2.36}    \\ \hline
    ResNet-50                              & C    & 26.9 & 4.1   & 6.85             & 27.0 & 5.4   & 2.48         \\
    ResNet-101                             & C    & 45.9 & 7.9   & 6.44             & 46.0 & 10.3  & 2.47         \\
    DeiT-S                                 & T    & 23.4 & 4.3   & 7.86             & 23.5 & 5.5   & 2.52         \\
    Swin-T                                 & T    & 30.2 & 4.6   & 6.97             & 30.3 & 6.1   & 2.45         \\
    Swin-S                                 & T    & 51.0 & 13.8  & 6.50             & 50.9 & 8.5   & 2.48         \\
    ConvNeXt-T                             & C    & 29.9 & 4.5   & 6.18             & 30.0 & 5.8   & 2.34         \\
    ConvNeXt-S                             & C    & 51.5 & 8.7   & 6.04             & 51.6 & 11.4  & 2.27         \\
    HorNet-T                               & C    & 23.7 & 4.3   & 6.46             & 23.8 & 5.6   & 2.39         \\
    \rowcolor{gray94}\bf{MogaNet-S}        & C    & 26.6 & 5.0   & \bf{6.08}        & 26.7 & 6.5   & \bf{2.24}    \\
    \bottomrule
    \end{tabular}
    }
    \vspace{-0.5em}
    \caption{\textbf{3D human pose estimation} with ExPose on FFHQ and FreiHAND datasets. The face and hand tasks use pre-vertex and pre-joint errors as the metric. The FLOPs of the face and hand tasks are measured with input images at $256^2$ and $224^2$ resolutions.}
    \vspace{-1.5em}
    \label{tab:3d_app}
\end{table}

%% file: Tabs/tab_coco_pose_app.tex
\begin{table}[t]
    \vspace{-1.0em}
    \setlength{\tabcolsep}{1.1mm}
    \centering
\resizebox{0.78\linewidth}{!}{
\begin{tabular}{lcccccccc}
    \toprule
Architecture                     & Type & Crop            & \#P. & FLOPs & AP        & AP$^{50}$ & AP$^{75}$ & AR        \\
                                 &      & size            & (M)  & (G)   & (\%)      & (\%)      & (\%)      & (\%)      \\ \hline
MobileNetV2                      & C    & $256\times 192$ & 10   & 1.6   & 64.6      & 87.4      & 72.3      & 70.7      \\
ShuffleNetV2~$2\times$           & C    & $256\times 192$ & 8    & 1.4   & 59.9      & 85.4      & 66.3      & 66.4      \\
\rowcolor{gray94}\bf{MogaNet-XT} & C    & $256\times 192$ & 6    & 1.8   & \bf{72.1} & \bf{89.7} & \bf{80.1} & \bf{77.7} \\ \hline
RSN-18                           & C    & $256\times 192$ & 9    & 2.3   & 70.4      & 88.7      & 77.9      & 77.1      \\
\rowcolor{gray94}\bf{MogaNet-T}  & C    & $256\times 192$ & 8    & 2.2   & \bf{73.2} & \bf{90.1} & \bf{81.0} & \bf{78.8} \\ \hline
ResNet-50                        & C    & $256\times 192$ & 34   & 5.5   & 72.1      & 89.9      & 80.2      & 77.6      \\
HRNet-W32                        & C    & $256\times 192$ & 29   & 7.1   & 74.4      & 90.5      & 81.9      & 78.9      \\
Swin-T                           & T    & $256\times 192$ & 33   & 6.1   & 72.4      & 90.1      & 80.6      & 78.2      \\
PVT-S                            & T    & $256\times 192$ & 28   & 4.1   & 71.4      & 89.6      & 79.4      & 77.3      \\
PVTV2-B2                         & T    & $256\times 192$ & 29   & 4.3   & 73.7      & 90.5      & 81.2      & 79.1      \\
Uniformer-S                      & H    & $256\times 192$ & 25   & 4.7   & 74.0      & 90.3      & 82.2      & 79.5      \\
ConvNeXt-T                       & C    & $256\times 192$ & 33   & 5.5   & 73.2      & 90.0      & 80.9      & 78.8      \\
\rowcolor{gray94}\bf{MogaNet-S}  & C    & $256\times 192$ & 29   & 6.0   & \bf{74.9} & \bf{90.7} & \bf{82.8} & \bf{80.1} \\ \hline
ResNet-101                       & C    & $256\times 192$ & 53   & 12.4  & 71.4      & 89.3      & 79.3      & 77.1      \\
ResNet-152                       & C    & $256\times 192$ & 69   & 15.7  & 72.0      & 89.3      & 79.8      & 77.8      \\
HRNet-W48                        & C    & $256\times 192$ & 64   & 14.6  & 75.1      & 90.6      & 82.2      & 80.4      \\
Swin-B                           & T    & $256\times 192$ & 93   & 18.6  & 72.9      & 89.9      & 80.8      & 78.6      \\
Swin-L                           & T    & $256\times 192$ & 203  & 40.3  & 74.3      & 90.6      & 82.1      & 79.8      \\
Uniformer-B                      & H    & $256\times 192$ & 54   & 9.2   & 75.0      & 90.6      & 83.0      & 80.4      \\
ConvNeXt-S                       & C    & $256\times 192$ & 55   & 9.7   & 73.7      & 90.3      & 81.9      & 79.3      \\
ConvNeXt-B                       & C    & $256\times 192$ & 94   & 16.4  & 74.0      & 90.7      & 82.1      & 79.5      \\
\rowcolor{gray94}\bf{MogaNet-B}  & C    & $256\times 192$ & 47   & 10.9  & \bf{75.3} & \bf{90.9} & \bf{83.3} & \bf{80.7} \\ \hline
MobileNetV2                      & C    & $384\times 288$ & 10   & 3.6   & 67.3      & 87.9      & 74.3      & 72.9      \\
ShuffleNetV2~$2\times$           & C    & $384\times 288$ & 8    & 3.1   & 63.6      & 86.5      & 70.5      & 69.7      \\
\rowcolor{gray94}\bf{MogaNet-XT} & C    & $384\times 288$ & 6    & 4.2   & \bf{74.7} & \bf{90.1} & \bf{81.3} & \bf{79.9} \\ \hline
RSN-18                           & C    & $384\times 288$ & 9    & 5.1   & 72.1      & 89.5      & 79.8      & 78.6      \\
\rowcolor{gray94}\bf{MogaNet-T}  & C    & $384\times 288$ & 8    & 4.9   & \bf{75.7} & \bf{90.6} & \bf{82.6} & \bf{80.9} \\ \hline
HRNet-W32                        & C    & $384\times 288$ & 29   & 16.0  & 75.8      & 90.6      & 82.7      & 81.0      \\
Uniformer-S                      & H    & $384\times 288$ & 25   & 11.1  & 75.9      & 90.6      & 83.4      & 81.4      \\
ConvNeXt-T                       & C    & $384\times 288$ & 33   & 33.1  & 75.3      & 90.4      & 82.1      & 80.5      \\
\rowcolor{gray94}\bf{MogaNet-S}  & C    & $384\times 288$ & 29   & 13.5  & \bf{76.4} & \bf{91.0} & \bf{83.3} & \bf{81.4} \\ \hline
ResNet-152                       & C    & $384\times 288$ & 69   & 35.6  & 74.3      & 89.6      & 81.1      & 79.7      \\
HRNet-W48                        & C    & $384\times 288$ & 64   & 32.9  & 76.3      & 90.8      & 82.0      & 81.2      \\
Swin-B                           & T    & $384\times 288$ & 93   & 39.2  & 74.9      & 90.5      & 81.8      & 80.3      \\
Swin-L                           & T    & $384\times 288$ & 203  & 86.9  & 76.3      & 91.2      & 83.0      & 814       \\
HRFormer-B                       & T    & $384\times 288$ & 54   & 30.7  & 77.2      & 91.0      & 83.6      & 82.0      \\
ConvNeXt-S                       & C    & $384\times 288$ & 55   & 21.8  & 75.8      & 90.7      & 83.1      & 81.0      \\
ConvNeXt-B                       & C    & $384\times 288$ & 94   & 36.6  & 75.9      & 90.6      & 83.1      & 81.1      \\
Uniformer-B                      & C    & $384\times 288$ & 54   & 14.8  & 76.7      & 90.8      & 84.0      & 81.4      \\
\rowcolor{gray94}\bf{MogaNet-B}  & C    & $384\times 288$ & 47   & 24.4  & \bf{77.3} & \bf{91.4} & \bf{84.0} & \bf{82.2} \\
    \bottomrule
    \end{tabular}
    }
    \vspace{-0.5em}
    \caption{\textbf{2D human pose estimation} with Top-Down SimpleBaseline on COCO \textit{val2017}. The FLOPs are measured at $256\times 192$ or $384\times 288$ resolutions.}
    \vspace{-1.0em}
    \label{tab:coco_pose_app}
\end{table}

%% file: Tabs/tab_vp_app.tex
\begin{table}[ht]
    % \vspace{-0.5em}
    \setlength{\tabcolsep}{1.3mm}
    \centering
\resizebox{0.82\linewidth}{!}{
\begin{tabular}{l|ccc|ccc|ccc}
    \toprule
Architecture                  & \#P. & FLOPs & FPS & \multicolumn{3}{c|}{200 epochs}                    & \multicolumn{3}{c}{2000 epochs}                    \\
                              & (M)  & (G)   & (s) & MSE$\downarrow$ & MAE$\downarrow$ & SSIM$\uparrow$ & MSE$\downarrow$ & MAE$\downarrow$ & SSIM$\uparrow$ \\ \hline
ViT                           & 46.1 & 16.9  & 290 & 35.15           & 95.87           & 0.9139         & 19.74           & 61.65           & 0.9539         \\
Swin                          & 46.1 & 16.4  & 294 & 29.70           & 84.05           & 0.9331         & 19.11           & 59.84           & 0.9584         \\
Uniformer                     & 44.8 & 16.5  & 296 & 30.38           & 85.87           & 0.9308         & 18.01           & 57.52           & 0.9609         \\ \hline
MLP-Mixer                     & 38.2 & 14.7  & 334 & 29.52           & 83.36           & 0.9338         & 18.85           & 59.86           & 0.9589         \\
ConvMixer                     & 3.9  & 5.5   & 658 & 32.09           & 88.93           & 0.9259         & 22.30           & 67.37           & 0.9507         \\
Poolformer                    & 37.1 & 14.1  & 341 & 31.79           & 88.48           & 0.9271         & 20.96           & 64.31           & 0.9539         \\ \hline
SimVP                         & 58.0 & 19.4  & 209 & 32.15           & 89.05           & 0.9268         & 21.15           & 64.15           & 0.9536         \\
ConvNeXt                      & 37.3 & 14.1  & 344 & 26.94           & 77.23           & 0.9397         & 17.58           & 55.76           & 0.9617         \\
VAN                           & 44.5 & 16.0  & 288 & 26.10           & 76.11           & 0.9417         & 16.21           & 53.57           & 0.9646         \\
HorNet                        & 45.7 & 16.3  & 287 & 29.64           & 83.26           & 0.9331         & 17.40           & 55.70           & 0.9624         \\
\rowcolor{gray94}\bf{MogaNet} & 46.8 & 16.5  & 255 & \bf{25.57}      & \bf{75.19}      & \bf{0.9429}    & \bf{15.67}      & \bf{51.84}      & \bf{0.9661}    \\
    \bottomrule
    \end{tabular}
    }
    \vspace{-0.5em}
    \caption{\textbf{Video prediction} with SimVP on Moving MNIST. The FLOPs and FPS are measured at the input tensor of $10\times 1\times 64\times 64$ on an NVIDIA Tesla V100 GPU.}
    \vspace{-0.5em}
    \label{tab:vp_app}
\end{table}

%% file: Tabs/tab_in1k_app_rsb.tex
\begin{table*}[t!]
    \centering
    \vspace{-1.0em}
    \setlength{\tabcolsep}{0.7mm}
\resizebox{1.0\linewidth}{!}{
\begin{tabular}{llcccccccc}
    \toprule
    Architecture                                       & Date       & Type & Param. & \multicolumn{3}{c}{100-epoch} & \multicolumn{3}{c}{300-epoch} \\
                                                       &            &      & (M)    & Train   & Test    & Acc (\%)   & Train   & Test    & Acc (\%)   \\ \hline
    ResNet-18~\citep{he2016deep}                       & CVPR'2016  & C    & 12     & 160$^2$ & 224$^2$ & 68.2       & 224$^2$ & 224$^2$ & 70.6       \\
    ResNet-34~\citep{he2016deep}                       & CVPR'2016  & C    & 22     & 160$^2$ & 224$^2$ & 73.0       & 224$^2$ & 224$^2$ & 75.5       \\
    ResNet-50~\citep{he2016deep}                       & CVPR'2016  & C    & 26     & 160$^2$ & 224$^2$ & 78.1       & 224$^2$ & 224$^2$ & 79.8       \\
    ResNet-101~\citep{he2016deep}                      & CVPR'2016  & C    & 45     & 160$^2$ & 224$^2$ & 79.9       & 224$^2$ & 224$^2$ & 81.3       \\
    ResNet-152~\citep{he2016deep}                      & CVPR'2016  & C    & 60     & 160$^2$ & 224$^2$ & 80.7       & 224$^2$ & 224$^2$ & 82.0       \\
    ResNet-200~\citep{he2016deep}                      & CVPR'2016  & C    & 65     & 160$^2$ & 224$^2$ & 80.9       & 224$^2$ & 224$^2$ & 82.1       \\
    ResNeXt-50~\citep{xie2017aggregated}               & CVPR'2017  & C    & 25     & 160$^2$ & 224$^2$ & 79.2       & 224$^2$ & 224$^2$ & 80.4       \\
    SE-ResNet-50~\citep{hu2018squeeze}                 & CVPR'2018  & C    & 28     & 160$^2$ & 224$^2$ & 77.0       & 224$^2$ & 224$^2$ & 80.1       \\
    EfficientNet-B0~\citep{icml2019efficientnet}       & ICML'2019  & C    & 5      & 160$^2$ & 224$^2$ & 73.0       & 224$^2$ & 224$^2$ & 77.1       \\
    EfficientNet-B1~\citep{icml2019efficientnet}       & ICML'2019  & C    & 8      & 160$^2$ & 224$^2$ & 74.9       & 240$^2$ & 240$^2$ & 79.4       \\
    EfficientNet-B2~\citep{icml2019efficientnet}       & ICML'2019  & C    & 9      & 192$^2$ & 256$^2$ & 77.5       & 260$^2$ & 260$^2$ & 80.1       \\
    EfficientNet-B3~\citep{icml2019efficientnet}       & ICML'2019  & C    & 12     & 224$^2$ & 288$^2$ & 79.2       & 300$^2$ & 300$^2$ & 81.4       \\
    EfficientNet-B4~\citep{icml2019efficientnet}       & ICML'2019  & C    & 19     & 320$^2$ & 380$^2$ & 81.2       & 380$^2$ & 380$^2$ & 82.4       \\
    RegNetY-800MF~\citep{cvpr2020regnet}               & CVPR'2020  & C    & 6      & 160$^2$ & 224$^2$ & 73.8       & 224$^2$ & 224$^2$ & 76.3       \\
    RegNetY-4GF~\citep{cvpr2020regnet}                 & CVPR'2020  & C    & 21     & 160$^2$ & 224$^2$ & 79.0       & 224$^2$ & 224$^2$ & 79.4       \\
    RegNetY-8GF~\citep{cvpr2020regnet}                 & CVPR'2020  & C    & 39     & 160$^2$ & 224$^2$ & 81.1       & 224$^2$ & 224$^2$ & 79.9       \\
    RegNetY-16GF~\citep{cvpr2020regnet}                & CVPR'2020  & C    & 84     & 160$^2$ & 224$^2$ & 81.7       & 224$^2$ & 224$^2$ & 80.4       \\
    EfficientNetV2-rw-S~\citep{icml2021EfficientNetV2} & ICML'2021  & C    & 24     & 224$^2$ & 288$^2$ & 80.9       & 288$^2$ & 384$^2$ & 82.9       \\
    EfficientNetV2-rw-M~\citep{icml2021EfficientNetV2} & ICML'2021  & C    & 53     & 256$^2$ & 384$^2$ & 82.3       & 320$^2$ & 384$^2$ & 81.9       \\ \hline
    ViT-T~\citep{iclr2021vit}                          & ICLR'2021  & T    & 6      & 160$^2$ & 224$^2$ & 66.7       & 224$^2$ & 224$^2$ & 72.2       \\
    ViT-S~\citep{iclr2021vit}                          & ICLR'2021  & T    & 22     & 160$^2$ & 224$^2$ & 73.8       & 224$^2$ & 224$^2$ & 79.8       \\
    ViT-B~\citep{iclr2021vit}                          & ICLR'2021  & T    & 86     & 160$^2$ & 224$^2$ & 76.0       & 224$^2$ & 224$^2$ & 81.8       \\
    PVT-T~\citep{iccv2021PVT}                          & ICCV'2021  & T    & 13     & 160$^2$ & 224$^2$ & 71.5       & 224$^2$ & 224$^2$ & 75.1       \\
    PVT-S~\citep{iccv2021PVT}                          & ICCV'2021  & T    & 25     & 160$^2$ & 224$^2$ & 72.1       & 224$^2$ & 224$^2$ & 79.8       \\
    Swin-T~\citep{liu2021swin}                         & ICCV'2021  & T    & 28     & 160$^2$ & 224$^2$ & 77.7       & 224$^2$ & 224$^2$ & 81.3       \\
    Swin-S~\citep{liu2021swin}                         & ICCV'2021  & T    & 50     & 160$^2$ & 224$^2$ & 80.2       & 224$^2$ & 224$^2$ & 83.0       \\
    Swin-S~\citep{liu2021swin}                         & ICCV'2021  & T    & 50     & 160$^2$ & 224$^2$ & 80.5       & 224$^2$ & 224$^2$ & 83.5       \\
    LITV2-T~\citep{nips2022hilo}                       & NIPS'2022  & T    & 28     & 160$^2$ & 224$^2$ & 79.7       & 224$^2$ & 224$^2$ & 82.0       \\
    LITV2-M~\citep{nips2022hilo}                       & NIPS'2022  & T    & 49     & 160$^2$ & 224$^2$ & 80.5       & 224$^2$ & 224$^2$ & 83.3       \\
    LITV2-B~\citep{nips2022hilo}                       & NIPS'2022  & T    & 87     & 160$^2$ & 224$^2$ & 81.3       & 224$^2$ & 224$^2$ & 83.6       \\
    ConvMixer-768-d32~\citep{2022convmixer}            & arXiv'2022 & T    & 21     & 160$^2$ & 224$^2$ & 77.6       & 224$^2$ & 224$^2$ & 80.2       \\
    PoolFormer-S12~\citep{yu2022metaformer}            & CVPR'2022  & T    & 12     & 160$^2$ & 224$^2$ & 69.3       & 224$^2$ & 224$^2$ & 77.2       \\
    PoolFormer-S24~\citep{yu2022metaformer}            & CVPR'2022  & T    & 21     & 160$^2$ & 224$^2$ & 74.1       & 224$^2$ & 224$^2$ & 80.3       \\
    PoolFormer-S36~\citep{yu2022metaformer}            & CVPR'2022  & T    & 31     & 160$^2$ & 224$^2$ & 74.6       & 224$^2$ & 224$^2$ & 81.4       \\
    PoolFormer-M36~\citep{yu2022metaformer}            & CVPR'2022  & T    & 56     & 160$^2$ & 224$^2$ & 80.7       & 224$^2$ & 224$^2$ & 82.1       \\
    PoolFormer-M48~\citep{yu2022metaformer}            & CVPR'2022  & T    & 73     & 160$^2$ & 224$^2$ & 81.2       & 224$^2$ & 224$^2$ & 82.5       \\ \hline
    ConvNeXt-T~\citep{cvpr2022convnext}                & CVPR'2022  & C    & 29     & 160$^2$ & 224$^2$ & 78.8       & 224$^2$ & 224$^2$ & 82.1       \\
    ConvNeXt-S~\citep{cvpr2022convnext}                & CVPR'2022  & C    & 50     & 160$^2$ & 224$^2$ & 81.7       & 224$^2$ & 224$^2$ & 83.1       \\
    ConvNeXt-B~\citep{cvpr2022convnext}                & CVPR'2022  & C    & 89     & 160$^2$ & 224$^2$ & 82.1       & 224$^2$ & 224$^2$ & 83.8       \\
    ConvNeXt-L~\citep{cvpr2022convnext}                & CVPR'2022  & C    & 189    & 160$^2$ & 224$^2$ & 82.8       & 224$^2$ & 224$^2$ & 84.3       \\
    ConvNeXt-XL~\citep{cvpr2022convnext}               & CVPR'2022  & C    & 350    & 160$^2$ & 224$^2$ & 82.9       & 224$^2$ & 224$^2$ & 84.5       \\
    HorNet-T$_{7\times 7}$~\citep{nips2022hornet}      & NIPS'2022  & C    & 22     & 160$^2$ & 224$^2$ & 80.1       & 224$^2$ & 224$^2$ & 82.8       \\
    HorNet-S$_{7\times 7}$~\citep{nips2022hornet}      & NIPS'2022  & C    & 50     & 160$^2$ & 224$^2$ & 81.2       & 224$^2$ & 224$^2$ & 84.0       \\
    VAN-B0~\citep{guo2022van}                          & CVMJ'2023  & C    & 4      & 160$^2$ & 224$^2$ & 72.6       & 224$^2$ & 224$^2$ & 75.8       \\
    VAN-B2~\citep{guo2022van}                          & CVMJ'2023  & C    & 27     & 160$^2$ & 224$^2$ & 81.0       & 224$^2$ & 224$^2$ & 82.8       \\
    VAN-B3~\citep{guo2022van}                          & CVMJ'2023  & C    & 45     & 160$^2$ & 224$^2$ & 81.9       & 224$^2$ & 224$^2$ & 83.9       \\
    \rowcolor{gray94}\bf{MogaNet-XT}                   & Ours       & C    & 3      & 160$^2$ & 224$^2$ & 72.8       & 224$^2$ & 224$^2$ & 76.5       \\
    \rowcolor{gray94}\bf{MogaNet-T}                    & Ours       & C    & 5      & 160$^2$ & 224$^2$ & 75.4       & 224$^2$ & 224$^2$ & 79.0       \\
    \rowcolor{gray94}\bf{MogaNet-S}                    & Ours       & C    & 25     & 160$^2$ & 224$^2$ & 81.1       & 224$^2$ & 224$^2$ & 83.4       \\
    \rowcolor{gray94}\bf{MogaNet-B}                    & Ours       & C    & 44     & 160$^2$ & 224$^2$ & 82.2       & 224$^2$ & 224$^2$ & 84.3       \\
    \rowcolor{gray94}\bf{MogaNet-L}                    & Ours       & C    & 83     & 160$^2$ & 224$^2$ & 83.2       & 224$^2$ & 224$^2$ & 84.7       \\
    \bottomrule
    \end{tabular}
    }
    \vspace{-0.5em}
    \caption{ImageNet-1K classification performance of tiny to medium size models (5$\sim$50M) training 100 and 300 epochs. RSB A3~\citep{wightman2021rsb} setting is used for 100-epoch training of all methods. As for 300-epoch results, the RSB A2~\citep{wightman2021rsb} setting is used for ResNet, ResNeXt, SE-ResNet, EfficientNet, and EfficientNetV2 as reproduced in timm~\citep{wightman2021rsb}, while other methods adopt settings in their original paper.}
    \label{tab:in1k_app_rsb}
    \vspace{-1.0em}
\end{table*}